\documentclass[mlabstract,submit]{jmlr}

\usepackage{longtable}
\usepackage{booktabs}
\usepackage[load-configurations=version-1]{siunitx} 

\theorembodyfont{\upshape}
\theoremheaderfont{\scshape}
\theorempostheader{:}
\theoremsep{\newline}

\usepackage{enumitem}
\usepackage{caption}
\title[Bispectral Optimal Transport]{Bispectral OT: Dataset Comparison using Symmetry-Aware Optimal Transport}

   \author{\Name{Annabel Ma} \Email{annabelma@college.harvard.edu}\\
   \Name{Kaiying Hou} \Email{kaiyinghou@gmail.com}\\
   \Name{David Alvarez-Melis} \Email{dam@seas.harvard.edu}\\
   \Name{Melanie Weber} \Email{mweber@seas.harvard.edu}\\
   \addr Harvard University, Cambridge, MA 02138, USA}

\begin{document}

\maketitle
\vspace{-\baselineskip}
\thispagestyle{plain}

\begin{abstract}

Optimal transport (OT) is a widely used technique in machine learning, graphics, and vision that aligns two distributions or datasets using their relative geometry. In symmetry-rich settings, however, OT alignments based solely on pairwise geometric distances between raw features can ignore the intrinsic coherence structure of the data. We introduce Bispectral Optimal Transport, a symmetry-aware extension of discrete OT that compares elements using their representation using the bispectrum, a group Fourier invariant that preserves all signal structure while removing only the variation due to group actions. Empirically, we demonstrate that the transport plans computed with Bispectral OT achieve greater class preservation accuracy than naive feature OT on benchmark datasets transformed with visual symmetries, improving the quality of meaningful correspondences that capture the underlying semantic label structure in the dataset while removing nuisance variation not affecting class or content.  

\end{abstract}


\section{Introduction}
\label{sec:intro}

Optimal Transport is a widely used technique for distribution alignment, providing a rigorous framework to learn a transport plan that moves mass from one distribution to match another. It is playing an increasing role in machine learning, due to applications in transfer learning \citep{alvarez2020geometric}, domain adaptation \citep{courty2017joint}, generative modeling \citep{arjovsky2017wasserstein, bousquet2017optimal}, generalization bounds \citep{chuang2021measuring}, and imitation learning \citep{luo2023optimal}. In symmetry-rich settings, however, OT alignments based solely on pairwise geometric distances can ignore the intrinsic coherence structure of the data (e.g., labels in supervised settings). For example, in an image dataset with rotational symmetries, naive OT on the raw features can match images based upon orientation, rather than the shape of the object depicted. Thus, we seek a transport plan that is \emph{symmetry-aware}: one that compares distributions in a way that is invariant to natural transformations of the data while retaining selectivity to informative structure.

We introduce \emph{Bispectral Optimal Transport (BOT)}, a symmetry-aware extension of discrete OT that compares elements through their representation under the \emph{bispectrum}---a complete invariant from group Fourier analysis that simultaneously encodes signal structure and invariance to group actions \citep{kakarala1993group}, first introduced to machine learning by \citet{kondor2007novel}. By computing couplings in this bispectral embedding, BOT produces correspondences that are invariant to transformations induced by symmetry groups acting on the data, without discarding discriminative information. To the authors knowledge, this is the first work that encodes symmetry awareness into OT, with a more extended discussion of related work included in Appendix \ref{apd:related-works}. We demonstrate that BOT better preserves label information than standard OT on datasets augmented via symmetry transformations, effectively encoding relevant symmetries in the learned transport plans. This is critical in settings where the symmetries acting on the two distributions do not align: for example, matching images captured by cameras at different poses or orientations. In such cases, we want the comparison to reflect the distribution of semantic labels rather than the distribution of camera angles; BOT achieves this and substantially outperforms standard OT on datasets perturbed by synthetic transformations. 

\section{Background}

\paragraph{Optimal Transport} Optimal transport provides an elegant mathematical framework for aligning probability distributions \citep{villani2008optimal}. At a high level, OT seeks to transfer probability mass between distributions while minimizing a cost function of transportation, yielding a notion of distance from the coupling. We are interested in the problem's discrete formulation, which considers two finite collections of points $\{\mathbf{x}^{(i)}\}_{i = 1}^n \in \mathcal{X}^n$ and $\{\mathbf{y}^{(j)}\}_{j = 1}^m \in \mathcal{Y}^n$ represented as empirical distributions: $\mu = \sum_{i = 1}^n \mathbf{p}_i \delta_{\mathbf{x}^{(i)}}, \nu = \sum_{j = 1}^m \mathbf{q}_j \delta_{\mathbf{y}^{(j)}}$ where $\mathbf{p}$ and $\mathbf{q}$ are probability vectors (non-negative and sum to one). Given a transportation cost $C \in \mathbb{R}^{n \times m}$ (also known as the \textit{ground metric}) between pairs of points (e.g., $C_{ij} = \|x_i - y_j\|$), OT finds a correspondence between $\mu$ and $\nu$ that minimizes this cost. Formally, Kantorovich's formulation of optimal transport \citep{kantorovich1942translocation} finds a coupling  $\Gamma \in \mathbb{R}^{n \times m}$ that solves 
\begin{equation}
\mathrm{OT}_c(\mu,\nu)=\min_{\Gamma\in\mathbb{R}_+^{n\times m}} \langle \Gamma, C\rangle
\quad\text{s.t.}\quad
\Gamma\mathbf{1}=\mathbf{p},\ \Gamma^\top\mathbf{1}=\mathbf{q}.
\end{equation}
This coupling $\Gamma$ can be interpreted as a soft matching between elements of $\mu$ and $\nu,$ in the sense that $\Gamma_{ij}$ is high if $\mathbf{x}^{(i)}$ and $\mathbf{y}^{(j)}$ are in correspondence, and low otherwise. In practice, the entropic-regularized Sinkhorn distance \citep{cuturi2013sinkhorn} is widely used, as it can be solved more efficiently via the Sinkhorn-Knopp algorithm. Specifically, the Sinkhorn distance solves $\min_{\Gamma \in \Pi(\mathbf{p}, \mathbf{q})} \langle \Gamma, C\rangle - \epsilon H(\Gamma),$ where $H(\Gamma) = - \sum_{ij} \Gamma_{ij} \log \Gamma_{ij}$ is an entropic regularizer that smooths the transportation plan. 

\paragraph{Bispectrum and Group-Invariant Fourier Embeddings} We encode symmetries via the bispectrum, a Fourier invariant that is \emph{complete}, removing specified group actions while preserving relative phase structure. To do this, we describe the group Fourier Transform \citep{rudin2017fourier}, with additional background deferred to Appendix \ref{apd:groups}. Let $f: G \rightarrow \mathbb{C}$ be a signal on a group $G$ with set of irreducible representations $Irr(G).$ The \emph{Generalized Fourier Transform (GFT)} is the linear map $f \mapsto \hat{f},$ where the Fourier frequencies are indexed by $\rho \in Irr(G),$ defined in the discrete case as  
\begin{equation}
    \hat{f}_{\rho} = \sum_{g \in G} f(g) \rho(g).
\end{equation}
In particular, for $G = \mathbb{Z}/n\mathbb{Z},$ the $\rho(g)$ are the $n$ discrete Fourier frequencies: $\rho_k(g) = e^{-\mathbf{i} 2 \pi  k g / n}$, for $k=0,\ldots,n-1$ and $g \in \mathbb Z/n \mathbb Z.$ For translation by $t\in G$, defined as $f^t(x):=f(t^{-1}x)$, the GFT, like the classical FT, obeys the \emph{Fourier shift property}: $\hat{f}^t_{\rho} = \rho(t) \hat{f}_{\rho}.$ This equivariance is exploited in the well-known \emph{power spectrum} $q_{\rho} = \hat{f}_{\rho}^{\dagger} \hat{f}_{\rho}$, which is invariant to translation but discards relative phase, losing structural information.

The \emph{bispectrum} is a lesser known Fourier invariant restoring this structure. For 1D signals $f \in \mathbb{R}^n$ with Fourier coefficients $\hat{f} = (\hat{f}_0, \dots, \hat{f}_{n-1}),$ the translation-invariant \emph{bispectrum} is the complex matrix $B \in \mathbb{C}^{n \times n}$ with entries 
\begin{equation}
    B_{i, j} = \hat{f}_i \hat{f}_j \hat{f}_{i+j \pmod n}^{\dagger},
\end{equation}
which is invariant to phase shifts due to translation, but does so while preserving the signal's relative phase structure. For compact commutative groups, the bispectrum is defined as 
\begin{equation}
    B_{\rho_i, \rho_j} = \hat{f}_{\rho_i} \hat{f}_{\rho_j} \hat{f}_{\rho_i \rho_j}.
\end{equation}
The non-commmutative bispectrum (Appendix \ref{apd:nc-bis}) is defined analogously, but accounts for matrix-valued irreps. Foundational work \citep{kakarala1993group} shows that the bispectrum is the lowest-degree spectral invariant that is \emph{complete}: it factors out specified group actions without losing signal structure. In labeled image datasets, this yields invariance to natural transformations (e.g., rotations) while preserving shape information---exactly the behavior required for symmetry-aware OT.

\section{Bispectral OT} \label{sec: bispectralot}

The Bispectral OT framework combines the symmetry-invariant properties of the bispectrum with the alignment capabilities of OT. The key idea is to compute an OT plan on symmetry-aware bispectral features, ensuring that the resulting correspondence respects the symmetries of the underlying data. Concretely, we propose a framework to construct the bispectral representation of grids (e.g., images) acted on by $SO(2),$ the group of planar rotations. For many non-commutative groups like $SO(3)$, bispectral feature embeddings as described in \citet{kondor2007novel} become computationally prohibitive for OT settings, which require us to compute and store a pairwise cost matrix between all bispectral features. The selective $G$-bispectrum \citep{mataigne2024selective} reduces this complexity, but it is unclear whether distances between its compressed features are expressive enough of global geometry for OT. Balancing efficiency with faithful geometry remains the central challenge. The rotation-invariant features are constructed as follows: 
\begin{enumerate}[itemsep=0.2em, topsep=4pt, parsep=0pt, partopsep=0pt]
    \item Convert each $M \times N$ pixel image into a discretized polar representation of size $R \times K$ (where $R$ is the number of radial bins and $K$ the number of angular bins).
    \item For each fixed radius $r$, extract the $1 \times K$ angular slice and compute its 1D discrete Fourier transform (DFT) along the angular dimension $(f_r(\theta_1), \dots, f_r(\theta_K)).$ 
    \item Compute the bispectrum of each slice. Since cyclic shifts of the angular dimension correspond to actions by $\mathbb{Z}/K\mathbb{Z}$ (i.e. discrete rotations), the bispectrum provides invariance to such transformations. Concretely, this maps our $R \times K$ polar image to $\mathbb{C}^{R \times K \times K}$ by mapping each DFT 
    \[(f_r(\theta_1), \dots, f_r(\theta_K)) \mapsto (f_r(\theta_i) f_r(\theta_j) f_r(\theta_i + \theta_j)^{\dagger})_{i, j \in 1, \ldots , K}.\]
    \item Concatenate the bispectral features across radii to obtain a global $SO(2)$-invariant representation of the image.
\end{enumerate}
These bispectral representations are used as inputs to the OT problem, using pairwise distance as cost. By aligning data in this invariant feature space, BOT computes transport plans that respect rotational symmetries, removing nuisance variation not affecting class or content while preserving the structural relationships needed for meaningful correspondences. 

\section{Experiments}

\begin{figure}[htbp]
\floatconts
  {fig:mds}
  {\caption{MDS visualization of raw and $\mathbb{Z}/40\mathbb{Z}$-bispectral features for rotated \textsc{mnist} digits}}
  {%
    \subfigure[1 Digit per Class, rotated by $\mathbb{Z}/40\mathbb{Z}$]{\label{fig:1digmds}%
      \includegraphics[height=1.3in]{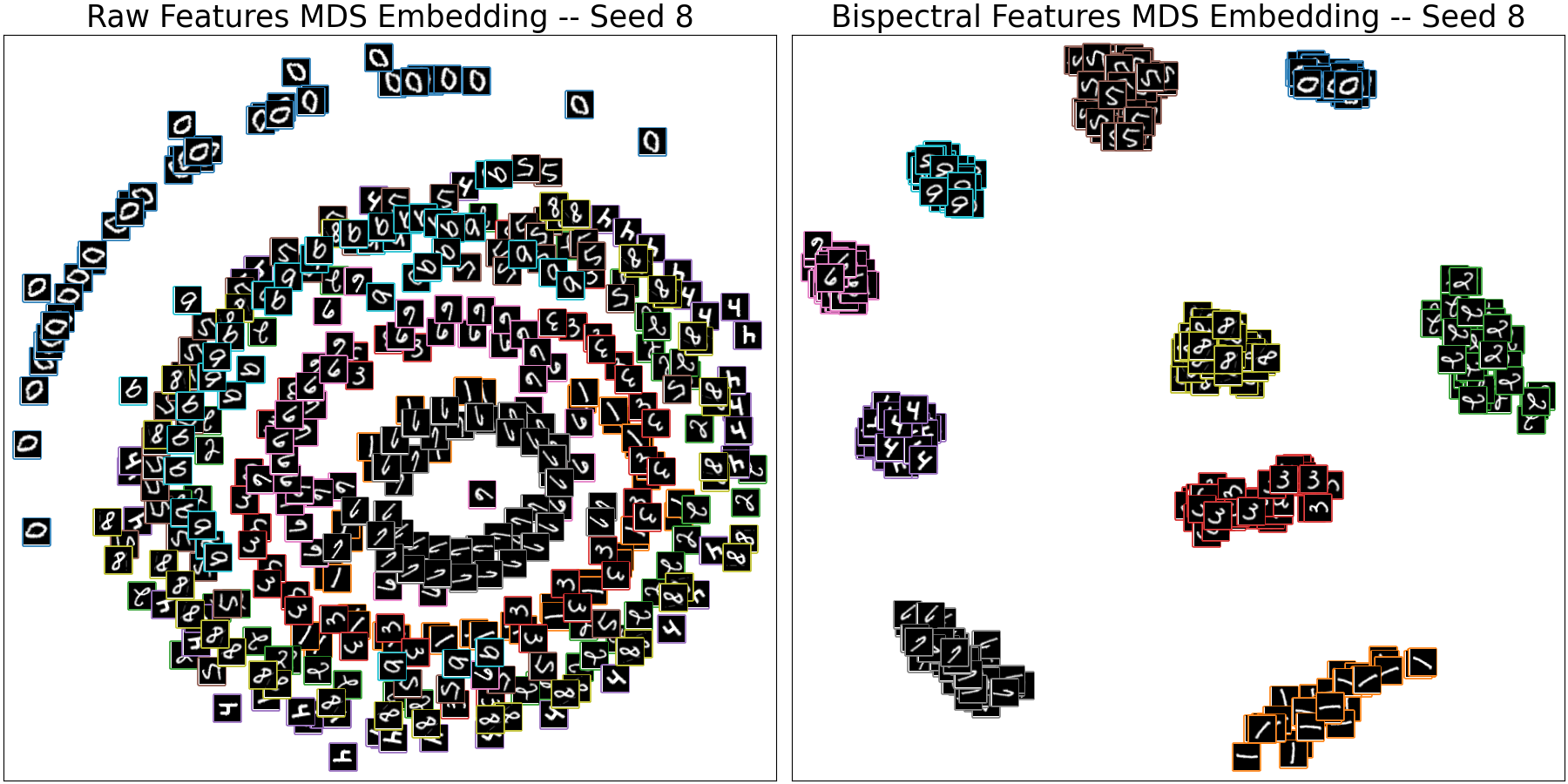}}%
    \hspace{0.5em}
    \subfigure[10 Digits per Class, rotated by $\mathbb{Z}/40\mathbb{Z}$]{\label{fig:10digmds}%
      \includegraphics[height=1.3in]{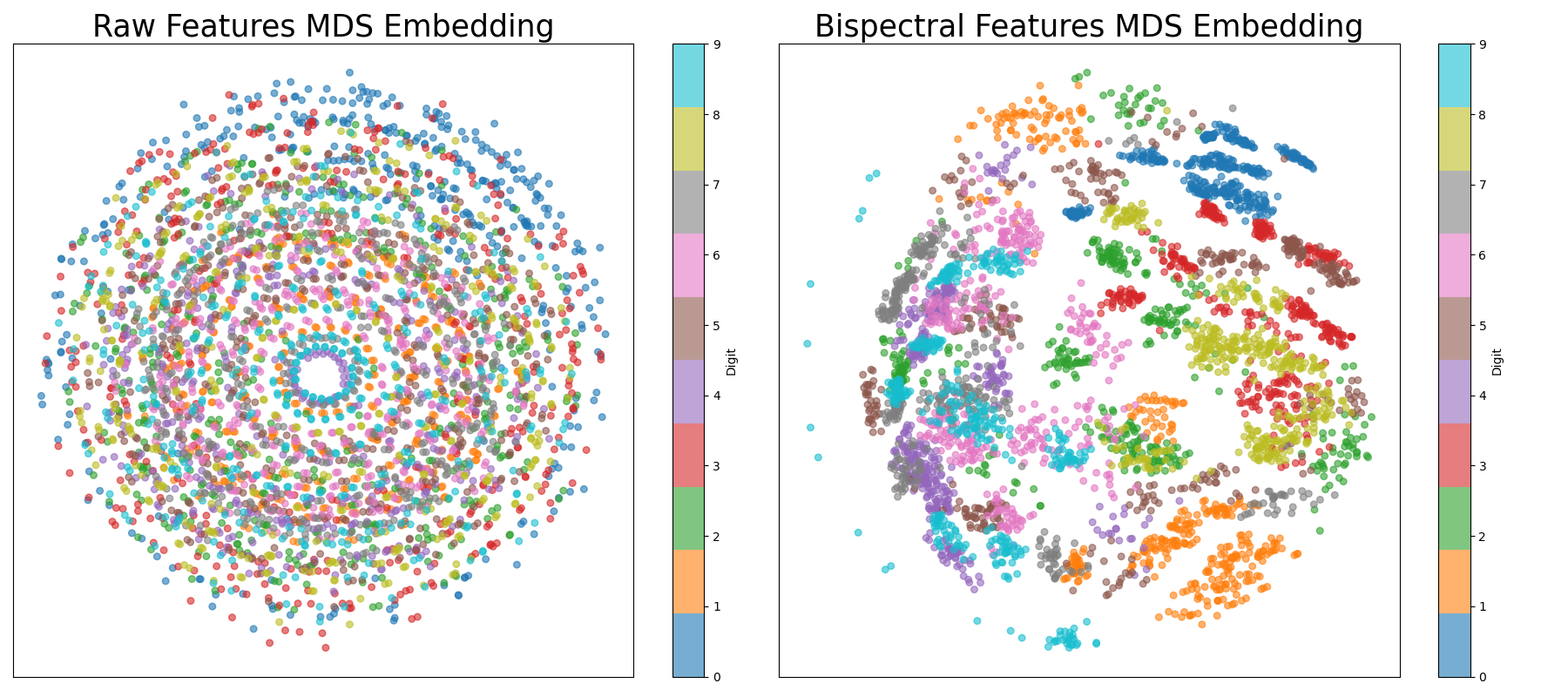}}
  }
\end{figure}

\paragraph{Understanding the Geometry of Bispectral Feature Space}\label{sec:prelims} We conduct preliminary experiments on rotated \textsc{mnist} \citep{mnist} to visualize the geometry of bispectral features. Figure~\ref{fig:mds} shows 2D MDS embeddings of raw pixel and bispectral representations. While pixel space scatters rotated digits uniformly, the rotation-invariant bispectral space clusters them by label. Moreover, rotationally symmetric digits (e.g., 6 and 9) are brought closer in bispectral space, and the digit with the most rotational symmetries (0) yields the most tightly-packed cluster, while others with fewer symmetries (e.g., 2) are more spread out (Subfigure~\ref{fig:1digmds}). Appendix~\ref{apd:rotmnist} details configurations and additional distance statistics for raw and bispectral features. 

\paragraph{Evaluating Bispectral OT}\label{sec:main-exp} We evaluate the performance of Bispectral OT using four benchmarks for classification: the \textsc{mnist}, \textsc{kuzushiji-mnist} (\textsc{kmnist}), \textsc{fashion-mnist} (\textsc{fmnist}), and \textsc{emnist} datasets \citep{mnist, kmnist, fashionmnist, emnist}. Details about the datasets and experimental setup are included in Appendix \ref{apd:dataset_details} and \ref{apd:main-results}, respectively. We split each dataset in half, applying rotations sampled uniformly at random to the first half of images and leaving the second half unrotated. We then compute the optimal transport plan using naive OT and BOT from the rotated images to the unrotated images using various pairwise distances in the latent embedding space as the ground metric for OT. To convert the OT plan into a mapping, we assign each source image to the target class receiving the largest transported mass under its row of the coupling $\Gamma$: $\hat y_i=\arg\max_{k\in\{1,\ldots,K\}}(\Gamma H)_{i,k}$, where $H$ is the one-hot encoding of the labels. Table \ref{tab:main_results} measures the fraction of images that are mapped to a image of the same class in each dataset under each coupling, showing the greatly improved ability of BOT to preserve semantic label structure on datasets transformed with rotation. Statistics for raw OT using all metrics and more detailed confusion matrices with per-class matching statistics are in Appendix \ref{apd:main-results}.

\begin{table}[htbp]
\floatconts
  {tab:main_results}
  {\caption{Class preservation accuracies for raw pixel OT vs. Bispectral OT}}
  {{\resizebox{\linewidth}{!}{
    \begin{tabular}{|l||c|c|c|c|c||c|c|c|c|c|}
    \hline
    & \multicolumn{5}{|c||}{\bfseries Matching Unrotated to Rotated} & \multicolumn{5}{c|}{\bfseries Baseline (Unrotated to Unrotated)} \\
    \cline{2-11}
    \bfseries Dataset & \bfseries OT & \multicolumn{4}{c||}{\bfseries Bispectral OT} & \bfseries OT & \multicolumn{4}{c|}{\bfseries Bispectral OT} \\
    \cline{3-6}\cline{8-11}
    & ($L_1$) & \bfseries $L_1$ &  \bfseries $L_2$ & \bfseries $L_2^2$ & \bfseries $\cos$ & ($L_1$) & \bfseries $L_1$ &  \bfseries $L_2$ & \bfseries $L_2^2$ & \bfseries $\cos$ \\
    \hline
    \textsc{mnist}        & 0.3297 & 0.8405 & 0.8020 & 0.8155
 & 0.8155 & 0.9725 & 0.8603 & 0.8242 & 0.8329 & 0.8329 \\
    \hline
    \textsc{kmnist}        & 0.2420 & 0.7815
 & 0.7225 & 0.7354
 & 0.7354 & 0.9724 & 0.8143
 & 0.7468
 & 0.7597 & 0.7597 \\
    \hline
    \textsc{fmnist}  & 0.3003 & 0.7617 & 0.7662 & 0.7319 & 0.7319 & 0.8726 & 0.7982 & 0.7913 & 0.7576 & 0.7576 \\
    \hline
    \textsc{emnist}        & 0.1969 & 0.5983 & 0.5693 & 0.5693 & 0.5716 & 0.8754 & 0.6416 & 0.6032 & 0.6032 & 0.6054 \\
    \hline
    \end{tabular}}}
  }
\end{table}

\section{Discussion}

In this work, we propose Bispectral OT, a symmetry-aware extension of optimal transport that computes distribution-wise distances and correspondences from bispectral representations of data invariant under group actions. Across benchmarks, we show BOT preserves semantic structure while discarding variability from transformations such as rotations. While encouraging, open challenges remain about scaling to complex non-gridded structures and richer groups or measuring distances between bispectral representations that respect their complex algebraic structure beyond common norms ($L_1$, $L_2$, $\cos$) without losing tractability. Overall, Bispectral OT provides a promising direction for symmetry-aware distribution alignment through OT, with potential applications in transfer learning and dataset comparison in symmetry-rich domains, opening the door to a family of transport methods leveraging algebraic structure to improve robustness and interpretability of dataset comparisons.

\section*{Reproducibility Statement}

The code for all experiments can be found at \url{https://github.com/annabel-ma/bispectral-ot}, with the specific configurations (random seeds and hyperparameters) detailed in the appendices. 

\acks{AM acknowledges support from KURE (Kempner Undergraduate Research Experience). DAM acknowledges support from the Kempner Institute, the Aramont Fellowship Fund, and the FAS Dean’s Competitive Fund for Promising Scholarship. MW was partially supported by NSF awards DMS-2406905 and CBET-2112085 and a Sloan Research Fellowship in Mathematics.}

\bibliography{bispectralOT}

\appendix

\section{Extended Related Works}\label{apd:related-works}

Representing collections of objects as empirical measures and comparing them via OT is an active area of research, with costs typically defined directly from features or via latent embeddings. For example, \citet{muzellec2018generalizing} models objects as elliptical distributions, and \citet{frogner_learning_2019} represents the embeddings as discrete measures. In supervised settings, label information can be injected into the cost \citep{courty_domain_2014, alvarez2018structured}, and hierarchical formulations address discrete labels \citep{alvarez2020geometric}. We seek to build upon this line of work by using a bispectral representation of an object to encode symmetries within a dataset. 

In another line of work, the theory of the group-invariant bispectrum was primarily developed in ~\citet{kakarala1993group, kakarala2009completeness, kakarala2012bispectrum} for signal processing contexts. The invariant bispectrum was first introduced to machine learning in \citet{kondor2007novel, kondor2008group}, utilizing embeddings of the non-commutative bispectrum in vision tasks. More recently, \citet{sanborn2022bispectral} described a neural network architecture using the bispectrum to learn groups from the data, and \citet{mataigne2024selective} introduced an algorithm to reduce the computational cost of the group bispectrum and other similar invariants. To the authors knowledge, this is the first work that encodes symmetry awareness into OT, bridging these two lines of work.

\section{Background on Group Representation Theory}\label{apd:groups}

We introduce the fundamentals of group representation theory, which serves as the foundation of the theory of the group bispectrum.

\begin{definition}[Group]
A \emph{group} $(G, \cdot)$ is a set $G$ with a binary operation referred to as the group product that satisfies the following axioms:
\begin{enumerate}[itemsep=0.2em, topsep=4pt, parsep=0pt, partopsep=0pt]
    \item \emph{Closure:} For all $a, b \in G,$ we have $ab \in G.$
    \item \emph{Associativity:} For any $a, b, c \in G,$ we have that $(ab)c = a(bc).$ 
    \item \emph{Identity:} There exists some \emph{identity} element $e$ such that for all $g \in G,$ we have $eg = ge = g.$ 
    \item \emph{Inverse:} For every element $g,$ there exists an \emph{inverse} element $g^{-1}$ such that $gg^{-1} = g^{-1} g = e.$ 
\end{enumerate}
\end{definition}

Concretely, a group $G$ can define a class of transformations like rotations or translations in the plane, with each element of the group defining a particular transformation. Groups that are important to us in this paper include the planar rotation group $SO(2),$ also known as the special orthogonal group, and its discrete analog, the cyclic group $\mathbb{Z}/n\mathbb{Z} = \{0, 1, \dots, n-1\}$ with group product addition modulo $n,$ which is the group of all rotational symmetries of a regular $n$-gon. These groups are \emph{commutative} or \emph{abelian,} which means that the order of operations does not matter: for all $a, b \in G$ for $G$ commutative, we have $ab = ba.$ This is in contrast to \emph{non-commutative} groups $G,$ where there exists some $a, b \in G$ such that $ab \neq ba.$ Examples of non-commutative symmetry group include $O(2),$ the group of all planar rotations and reflections, $SO(3),$ the group of all 3D rotations, and $D_n,$ the group of symmetries of an $n$-gon, which can be viewed as the discrete version of $O(2).$ 

\begin{definition}[Group Homomorphism]
    A \emph{group homomorphism} between two groups $G$ and $H$ with operations $\cdot$ and $*,$ respectively, is a map $\rho: G \rightarrow H$ that respects the underlying group structure of $G$ and $H,$ i.e. $\rho(u \cdot v) = \rho(u) * \rho(v).$ If such a map exists, then $G$ and $H$ are called \emph{homomorphic.} If such a map is a bijection, it is then called an \emph{isomorphism}. 
\end{definition}

Note that isomorphic groups are essentially the same group, but arising in different contexts. This motivates \emph{representation theory}, which studies groups via linear actions on vector spaces, using linear algebra to make abstract structures concrete. 
\begin{definition}[Group Representation]
A \emph{representation} of a group $G$ is a group homomorphism $\rho : G \rightarrow GL(V)$ assiging elements of $G$ to elements of the group of linear transformations over a vector space $V.$ In most contexts, $V$ is $\mathbb{R}^n$ or $\mathbb{C}^n.$
\end{definition}
A representation is \emph{reducible} if there exists a change of basis that decomposes the representation into a direct sum of other representations. An \emph{irreducible representation} cannot be decomposed in this way, and the set of them $Irr(G)$ are often called the \emph{irreps} of $G.$ For all finite groups and compact groups, which is the only classes of groups we will consider, the irreps consist only of unitary transformations, so throughout this paper, we assume $\rho(g^{-1}) = \rho(g)^{-1} = \rho(g)^{\dagger}$ for $\dagger$ the conjugate transpose. For commutative groups, the irreducible representations are scalars and in bijection with the group elements. This nice characterization does not extend to non-commutative groups, where the irreducible representations are matrix valued of variable dimension. 

The discussion of using symmetry groups to explain natural transformations in datasets motivates the definition of a \emph{group action,} which we provide now: 

\begin{definition}[Group Action] For $G$ a group and $X$ a set, a \emph{group action} is a map $T: G \times X \rightarrow X$ with the following properties:
\begin{enumerate}[itemsep=0.2em, topsep=4pt, parsep=0pt, partopsep=0pt]
\item The identity $e$ maps an element $x \in X$ to itself: for all $x \in X,$, we have $T(e, x) = x.$ 
\item For all $g_1, g_2 \in G,$ we have $T(g_1, T(g_2, x)) = T(g_1g_2, x).$
\end{enumerate}
\end{definition}
For simplicity, given a group action $T,$ we often say that a point $x$ maps to $gx$ ($= T(g, x)$). Then, if $X$ is a space on which a group $G$ acts, we define the \emph{orbit} of a point $x \in X$ to be the set $\{gx : g \in G\}.$ In the context of image transformations, the orbit is the set of all transformed versions of an image. For example, if $G = SO(2),$ the orbit contains all rotated versions of that image. Using group actions, we can also define the concepts of \emph{invariance} and \emph{equivariance,} which are critical to literature in geometric machine learning. 

\begin{definition}[Invariance]
For sets $X, Y$, a function $f : X \mapsto Y$ is \emph{$G$-invariant} if $f(x) = f(gx)$ for all $g \in G$ and $x \in X.$ In other words, group actions on the input space have no effect on the output. 
\end{definition}

\begin{definition}[Equivariance]
For sets $X, Y$, a function $f : X \mapsto Y$ is \emph{$G$-equivariant} if $f(gx) = g'f(x)$ for all $g \in G$ and $x \in X,$ where $g'\in G',$ a group homomorphic to $G.$ In other words, group actions on the input space results in a corresponding group action on the output space. 
\end{definition}

The bispectrum of the group $G$ is an example of a $G$-invariant function, which we exploit in the representations of images used in the paper. 

\section{The Non-Commutative Bispectrum}\label{apd:nc-bis}

The bispectrum has an analogous form in the setting of non-commutative groups, but is adjusted to account for the fact that the irreducible representations of a non-commutative group are not necessarily one dimensional, unlike in the compact commutative case. This more general form of the bispectrum is defined to be
\begin{equation}
    \beta_{\rho_i, \rho_j} = [\hat{f}_{\rho_i} \otimes \hat{f}_{\rho_j}] C_{\rho_i, \rho_j} \left[\bigoplus_{\rho \in \rho_i \otimes \rho_j} \hat{f}_{\rho}^{\dagger}\right] C_{\rho_i, \rho_j}^{\dagger},
\end{equation}
where $\oplus$ is a direct sum over irreducible representations, $\otimes$ is a tensor product, and $C_{\rho_i, \rho_j}$ is a unitary matrix defining a Clebsch-Gordan decomposition on the tensor product of a pair of irreducible representations \citep{kondor2007novel}.

\section{Dataset Details}\label{apd:dataset_details}

Information about the datasets used, including references, are provided in Table \ref{tab:dataset_details}. \textsc{mnist} has one class for each digit, \textsc{kuzushiji-mnist} has classes corresponding to distinct cursive Japanese calligraphy characters, \textsc{fashion-mnist} has classes corresponding to different articles of clothing, and the letters split of \textsc{emnist} has one class for each letter of the alphabet.

\begin{table}[htbp]
\floatconts
  {tab:dataset_details}
  {\caption{Summary of all the datasets used in this work. For all experiments, we normalize the dataset to have mean $0$ and standard deviation $1.$}}
  {\resizebox{\linewidth}{!}{%
		\begin{tabular}{r c c c c c} 
			\toprule
			 Dataset & Input Dimension & Number of Classes & Train Examples & Test Examples & Source \\
			\midrule
            \textsc{mnist} & $28\times 28$ & $10$ & $60$K & $10$K & \citep{mnist}\\
            \textsc{kuzushiji-mnist} & $28\times 28$ & $10$ & $60$K & $10$K & \citep{kmnist}\\
            \textsc{fashion-mnist} & $28\times 28$ & 10 &$60$K & $10$K & \citep{fashionmnist} \\
            \textsc{emnist} (letters) & $28\times 28$ & 26 & $145$K & $10$K & \citep{emnist}\\
			\bottomrule
		\end{tabular}%
    }}
\end{table}

\section{$O(2)$-\textsc{mnist} Preliminary Experiments}\label{apd:rotmnist}

\subsection{Experimental Setup}

Throughout this paper, we use $R = \frac{\min(M, N)}{2}$ radial bins and $K = 40$ angular bins in the discretized polar representation of our images of size $M \times N$ for the sake of computational feasibility. Thus, the bispectrum we compute is the $\mathbb{Z}/40\mathbb{Z}$ bispectrum, which is invariant to group actions that define rotations by $\frac{360^{\circ}}{40} = 9^{\circ}.$ Due to the discrete grid structure of features in pixel space, increasing $K$ does not necessarily give us better symmetry-invariant representations -- our discrete polar representation is of size $R \times K,$ and reflects at most $M \times N$ pixels of information. Since $R \sim \min(M, N),$ the number of angular bins must be on the order of $K \sim \max(M, N),$ which in the case of \textsc{mnist}, is $M = N = 28.$ 

To construct the 2D MDS embeddings of the pixel and bispectral representations, we randomly sampled representative images from each class of the train split of \textsc{mnist}, and rotated each by all multiples of $9^{\circ}$ from $0^\circ$ to $360^{\circ}.$ Specifically, Figure \ref{fig:1digmds} depicts MDS embeddings of the pixel and bispectral representations of rotations of one representative image for clearer visualization of cluster structure (for a total of $10\cdot 40 = 400$ embedded features), and Figure \ref{fig:10digmds} depicts MDS embeddings of the pixel and bispectral representations of rotations of ten representative images per class for a more global visualization of the embeddings with intra-class variation (for a total of $10\cdot 10 \cdot 40 = 4000$ embedded features). The code to generate these plots is included in the linked repository (seed 8).

\subsection{Additional Inter-Class Distance Statistics}

We also include the following confusion matrices in Figure \ref{fig:inter-class-distance-stats} depicting the average inter-class and intra-class distance statistics for the raw and bispectral representations of ten randomly sampled images of each class of \textsc{mnist} rotated by multiples of $9^{\circ}$ using different geometric distances. Euclidean distance denotes the $L_2$ norm $\sqrt{\sum_i (x_i - y_i)^2}$, cityblock distance denotes the $L_1$ norm $\sum_i|x_i - y_i|$, sqeuclidean denotes the squared $L_2$ norm $\sum_i (x_i - y_i)^2$, and cosine denotes the cosine distance, defined as $1 - \frac{u \cdot v}{\|u\|_2 \|v\|_2}$ between two vectors where $\cdot$ denotes the dot product and $\|* \|_2$ is the $L_2$ norm.

\begin{figure}[htbp]
\floatconts
  {fig:inter-class-distance-stats}
  {\caption{Average inter-class and intra-class distances for pixel and bispectral representations of MNIST digits using different metrics.}}
  {%
    \subfigure[Pixels, $L_2$]{
      \includegraphics[width=0.22\linewidth]{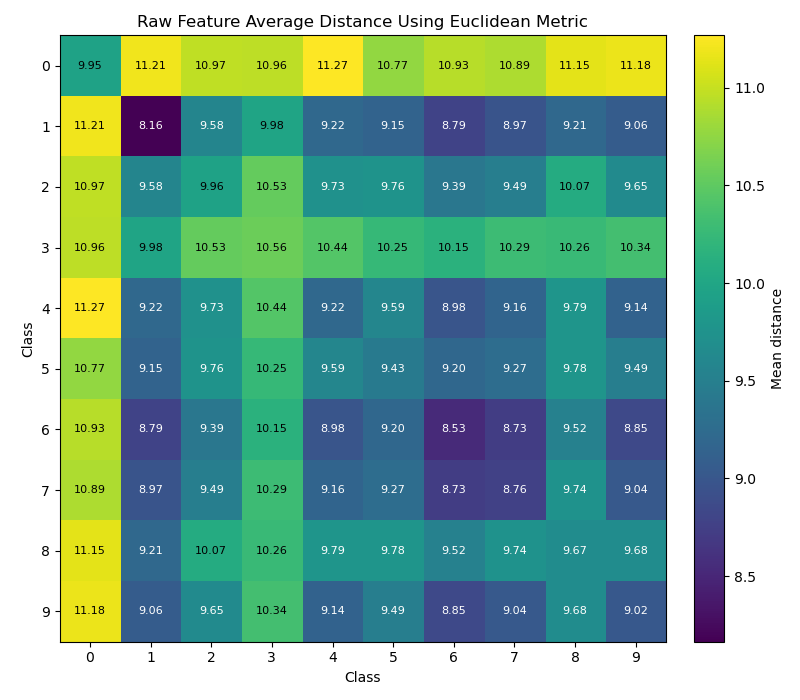}}%
    \quad
    \subfigure[Bispectral, $L_2$]{
      \includegraphics[width=0.22\linewidth]{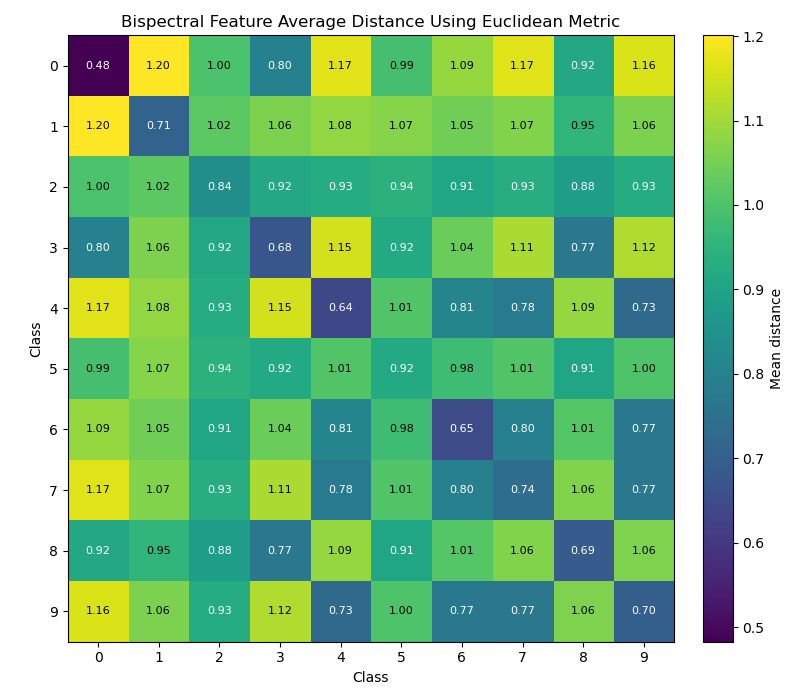}}
    \quad 
      \subfigure[Pixels, $L_1$]{
      \includegraphics[width=0.22\linewidth]{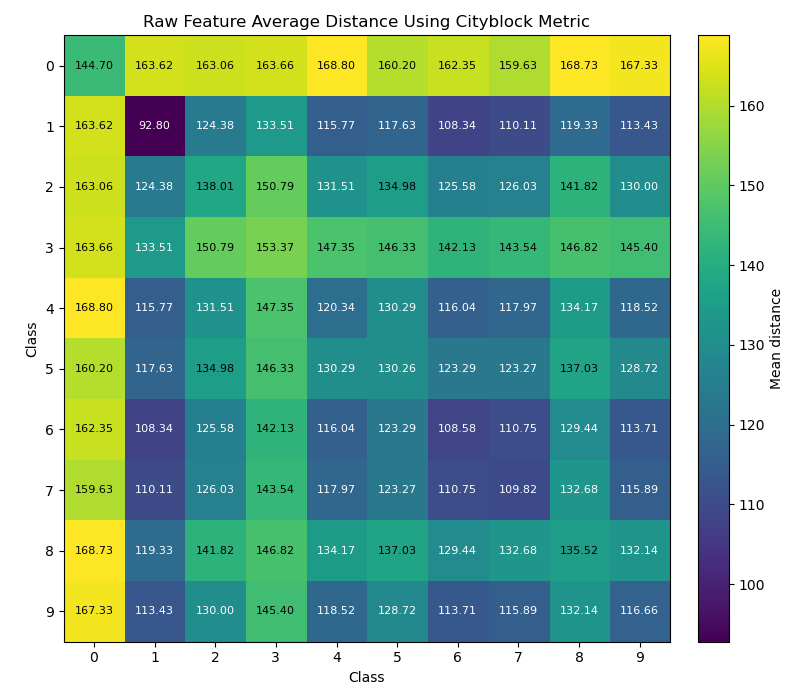}}%
    \quad
    \subfigure[Bispectral, $L_1$]{
      \includegraphics[width=0.22\linewidth]{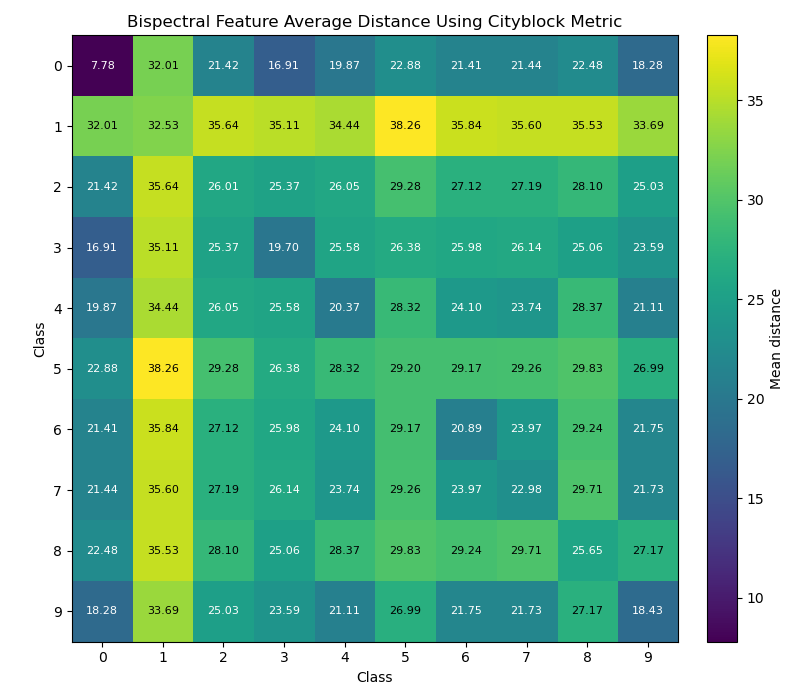}}
      
    \subfigure[Pixels, $L_2^2$]{
      \includegraphics[width=0.22\linewidth]{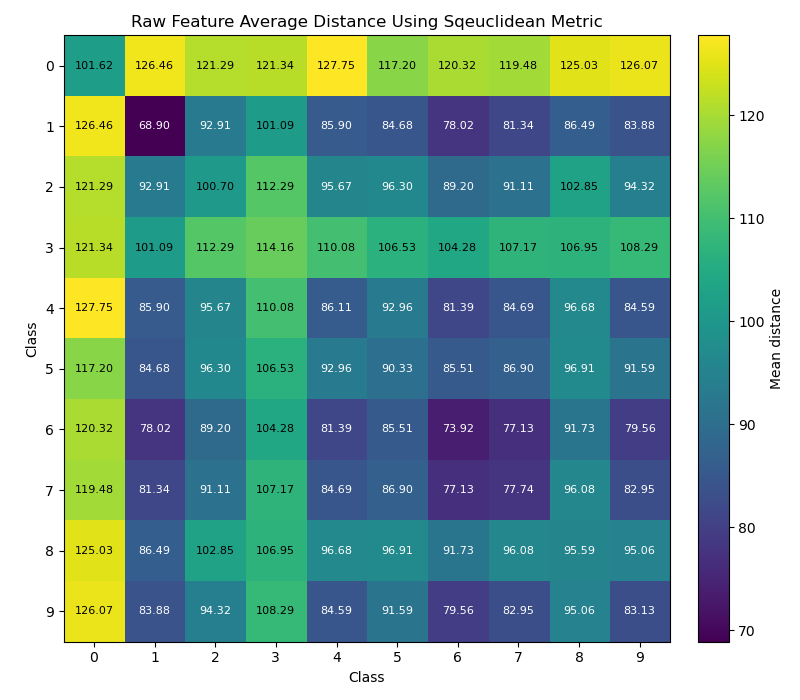}}%
    \quad
    \subfigure[Bispectral, $L_2^2$]{
      \includegraphics[width=0.22\linewidth]{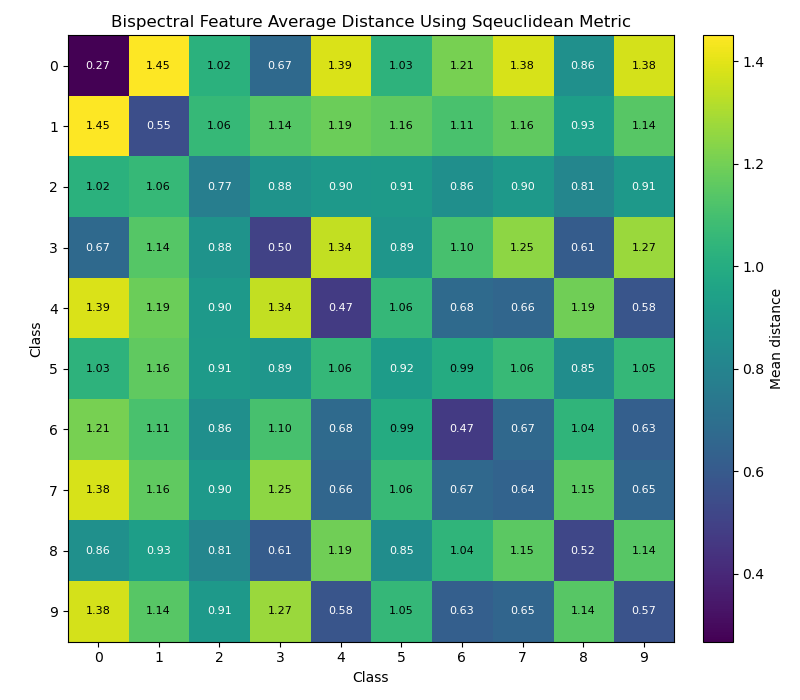}}
    \quad 
      \subfigure[Pixels, $\cos$]{
      \includegraphics[width=0.22\linewidth]{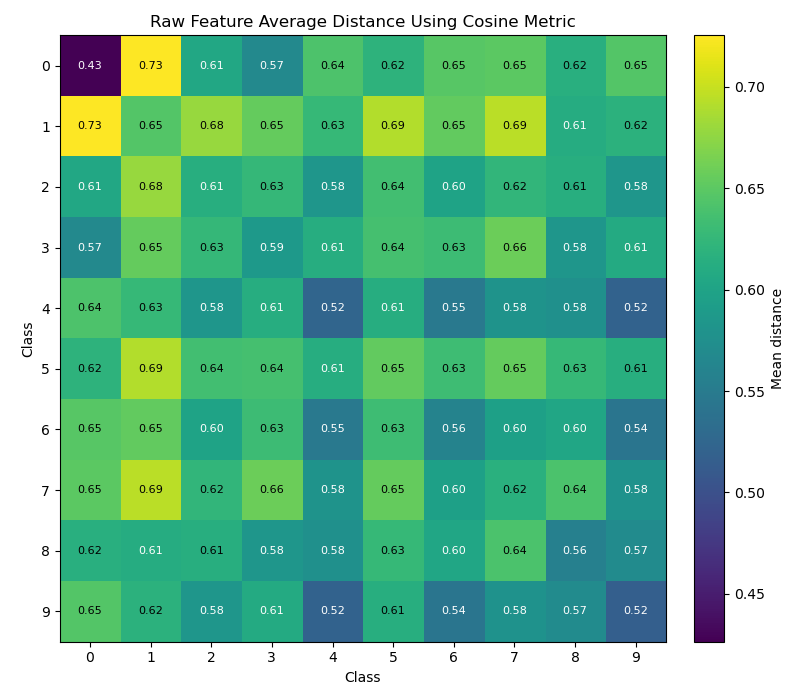}}%
    \quad
    \subfigure[Bispectral, $\cos$]{
      \includegraphics[width=0.22\linewidth]{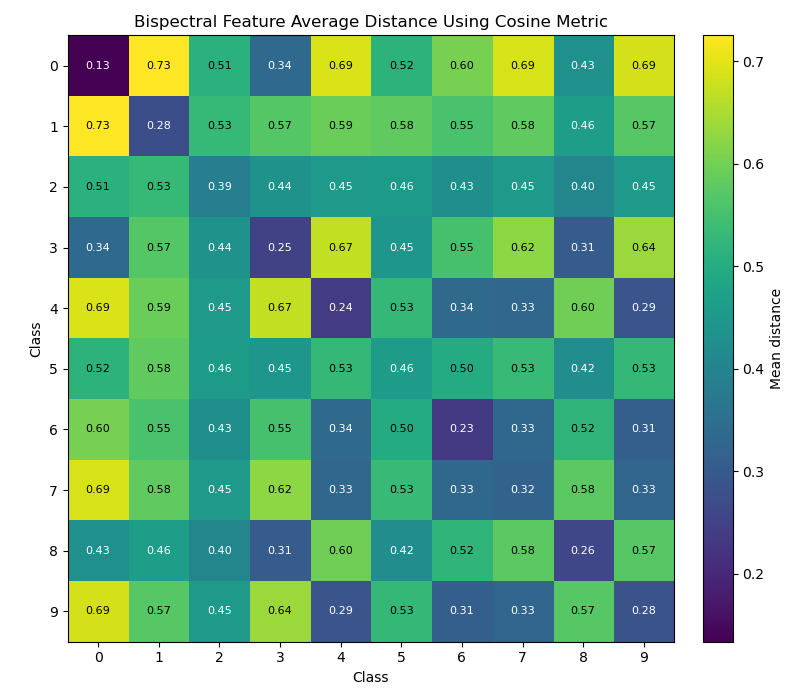}}
  }
\end{figure}

\subsection{Additional Per-Digit Distance Statistics}

In addition to the confusion matrices in Figure \ref{fig:inter-class-distance-stats}, we include the average distance between images of the same class that are rotated by different angles for a better understanding of the local geometry within each class of the bispectral features, in comparison to the raw pixel representations of images. For greater ease of visualization, we rotate each of our 10 sampled images from each class by each multiple of $15^{\circ}$ from $0^{\circ}$ to $360^{\circ},$ and continue to use the $\mathbb{Z}/40\mathbb{Z}$-bispectrum in the bispectral representation of images. The grid-like patterns are likely a function of how we handled clipping due to rotations. 

The distance statistics using the $L_2$ distance are depicted in Figure \ref{fig:l2-per-digit}, the distance statistics using the $L_1$ distance are depicted in Figure \ref{fig:l1-per-digit}, the distance statistics using the $L_2^2$ distance are depicted in Figure \ref{fig:sqeuclidean-per-digit}, and the distance statistics using cosine similarity are depicted in Figure \ref{fig:cos-per-digit}. 

\begin{figure}[htbp]
\floatconts
  {fig:l2-per-digit}
  {\caption{Average distance between bispectral and pixel representations of rotated images from the same class of \textsc{mnist} using euclidean norm.}}
  {%
    \subfigure[Raw feature representations]{%
      \centering
      \begin{tabular}{ccccc}
        \includegraphics[width=0.18\linewidth]{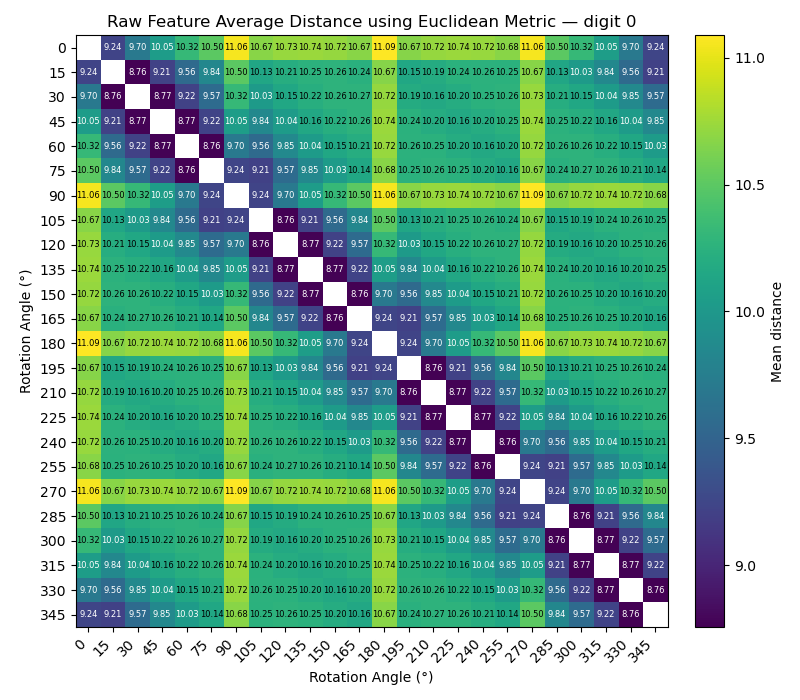} &
        \includegraphics[width=0.18\linewidth]{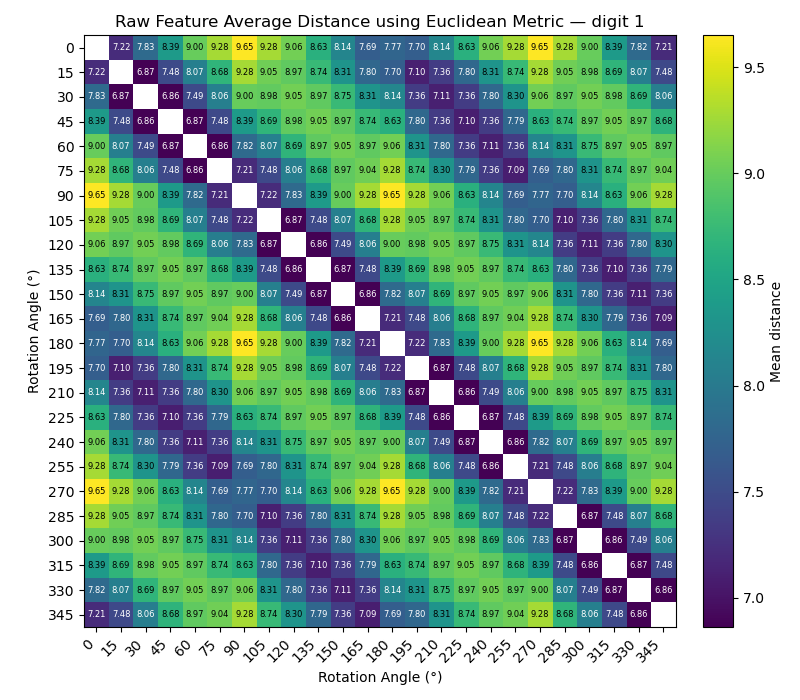} &
        \includegraphics[width=0.18\linewidth]{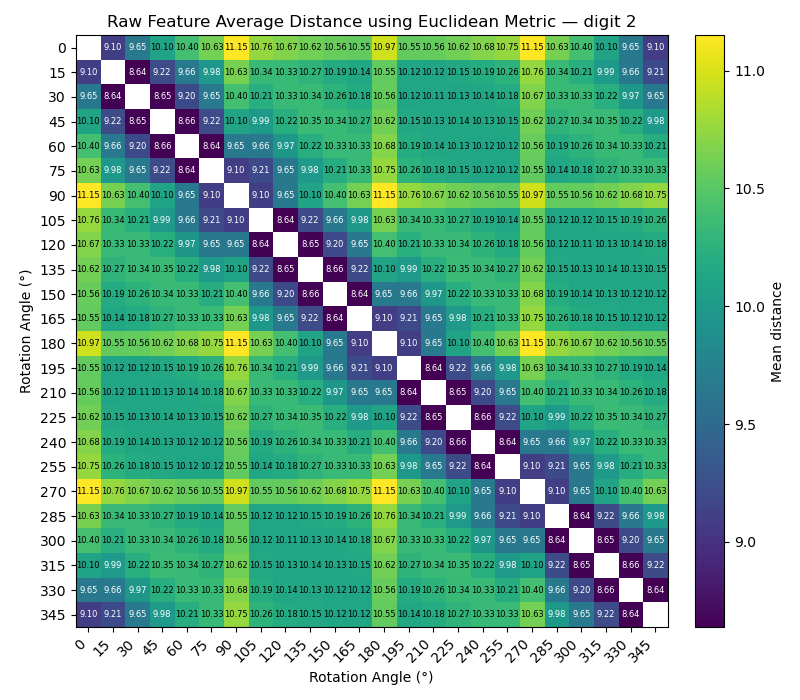} &
        \includegraphics[width=0.18\linewidth]{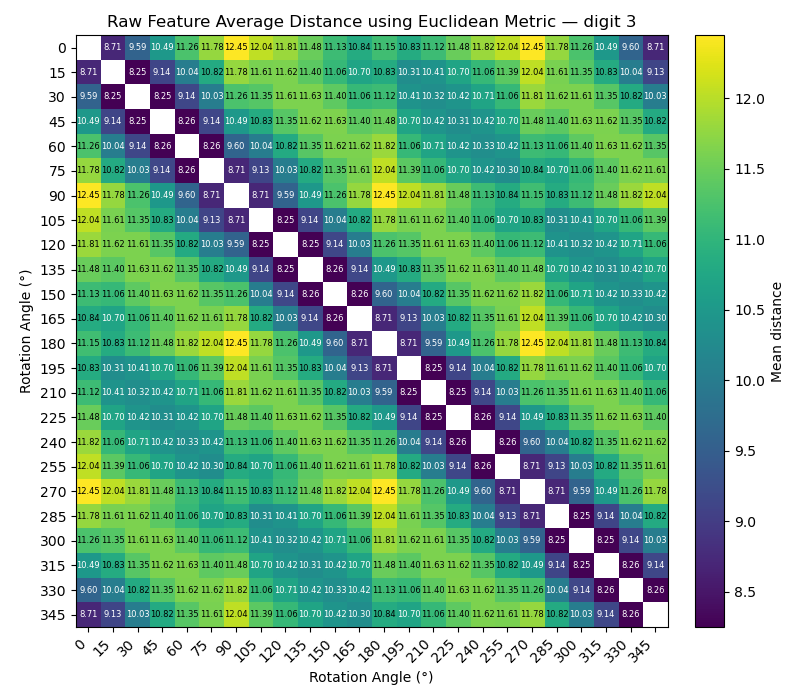} &
        \includegraphics[width=0.18\linewidth]{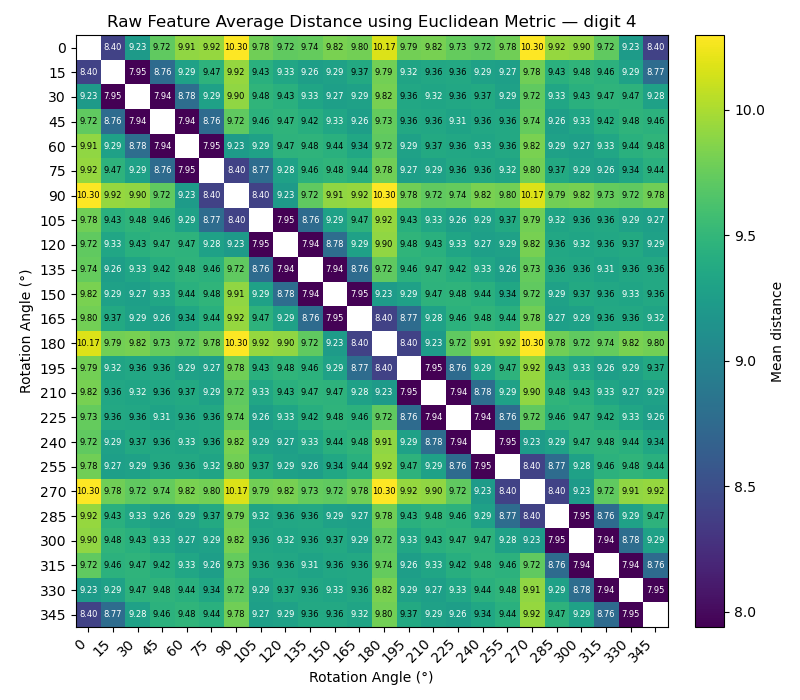} \\
        \includegraphics[width=0.18\linewidth]{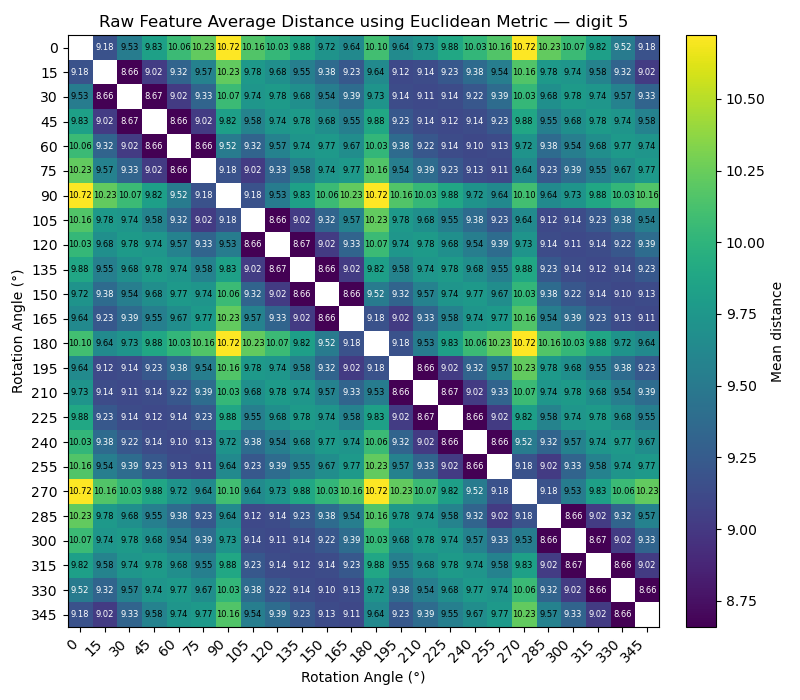} &
        \includegraphics[width=0.18\linewidth]{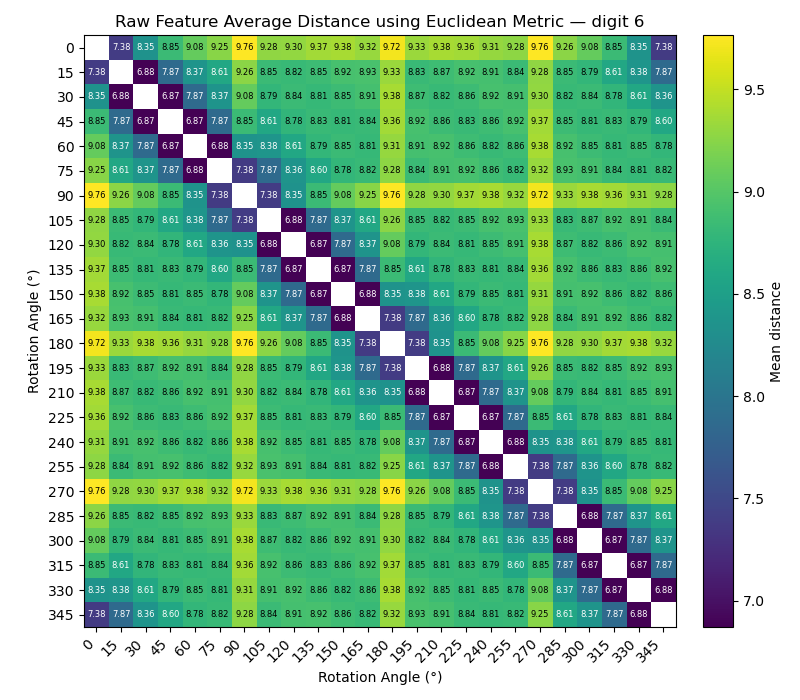} &
        \includegraphics[width=0.18\linewidth]{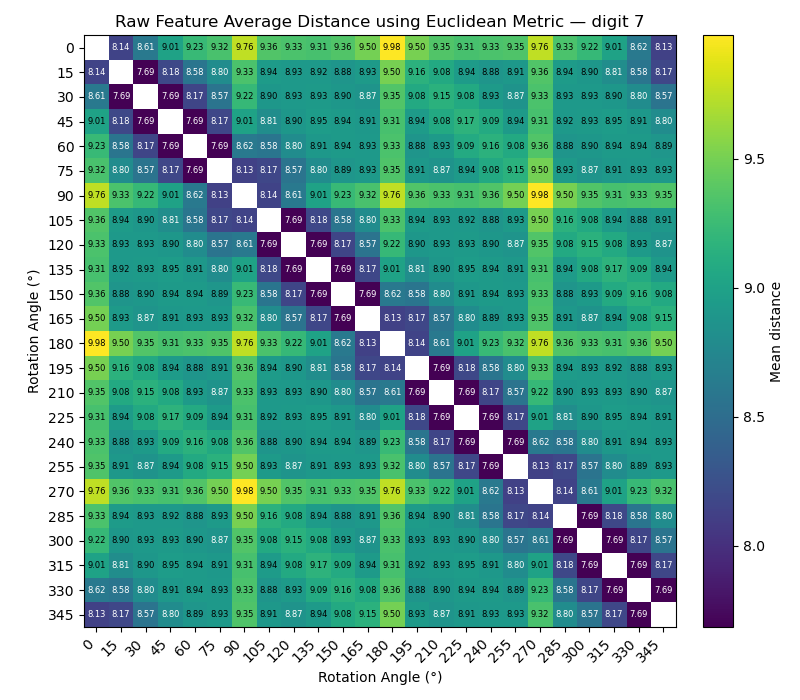} &
        \includegraphics[width=0.18\linewidth]{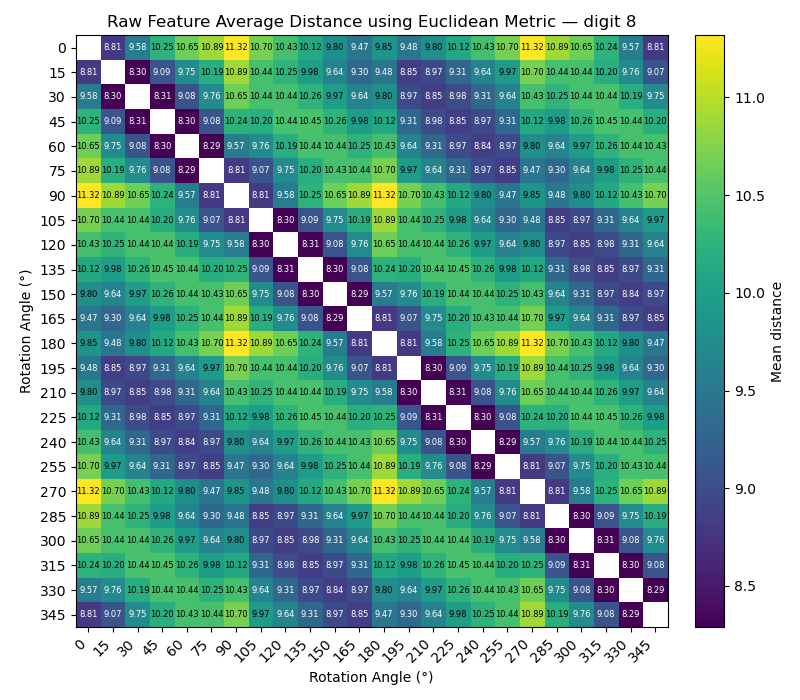} &
        \includegraphics[width=0.18\linewidth]{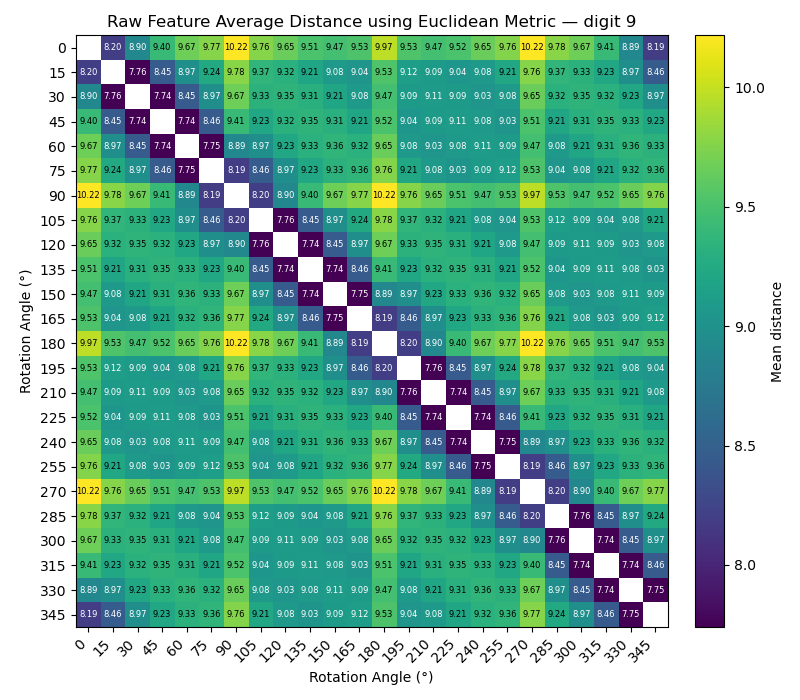} 
      \end{tabular}
    }\quad
    \subfigure[Bispectral feature representations]{%
      \centering
      \begin{tabular}{ccccc}
        \includegraphics[width=0.18\linewidth]{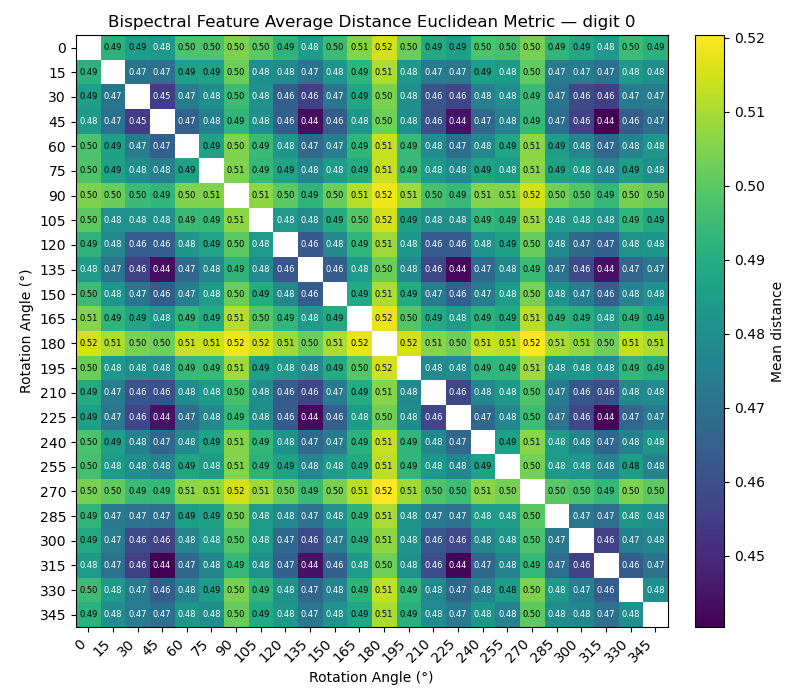} &
        \includegraphics[width=0.18\linewidth]{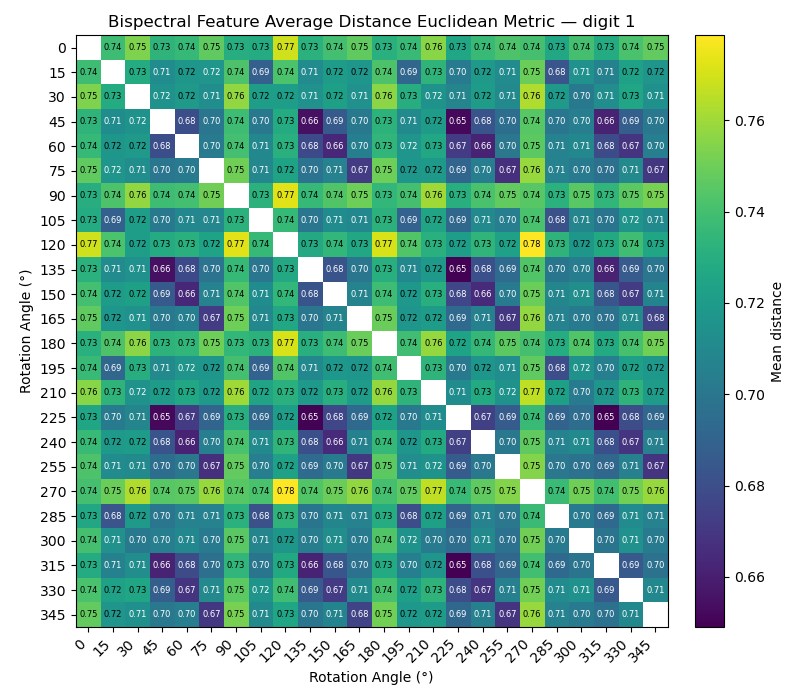} &
        \includegraphics[width=0.18\linewidth]{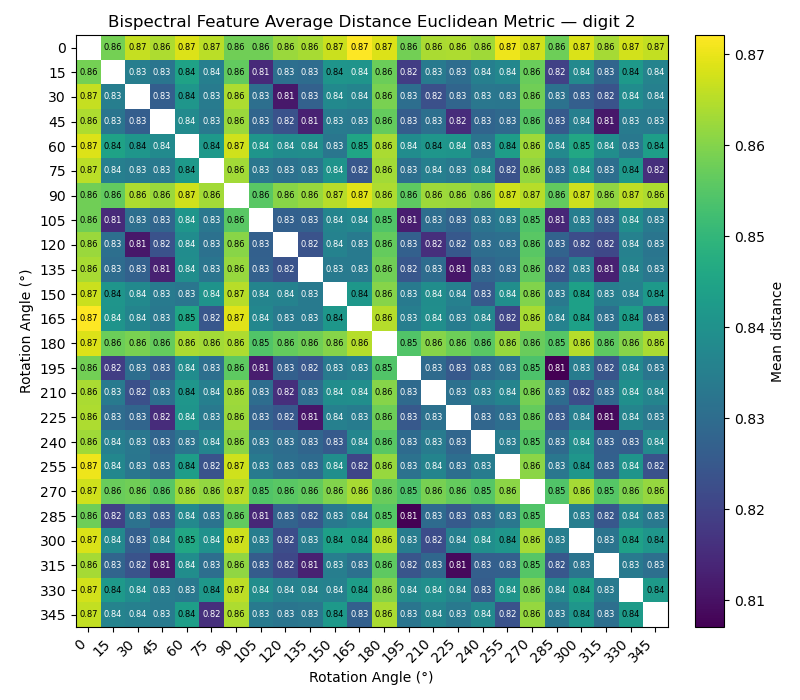} &
        \includegraphics[width=0.18\linewidth]{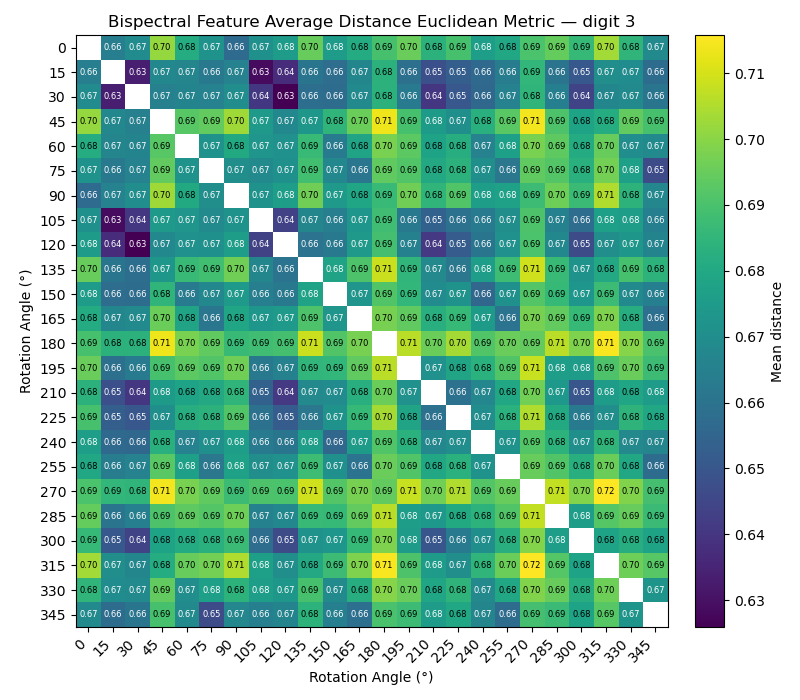} &
        \includegraphics[width=0.18\linewidth]{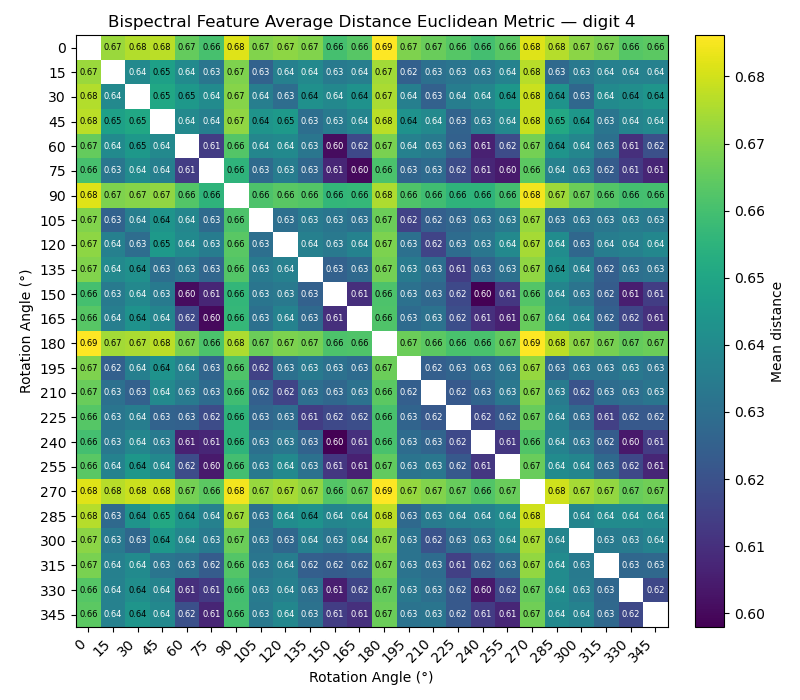} \\
        \includegraphics[width=0.18\linewidth]{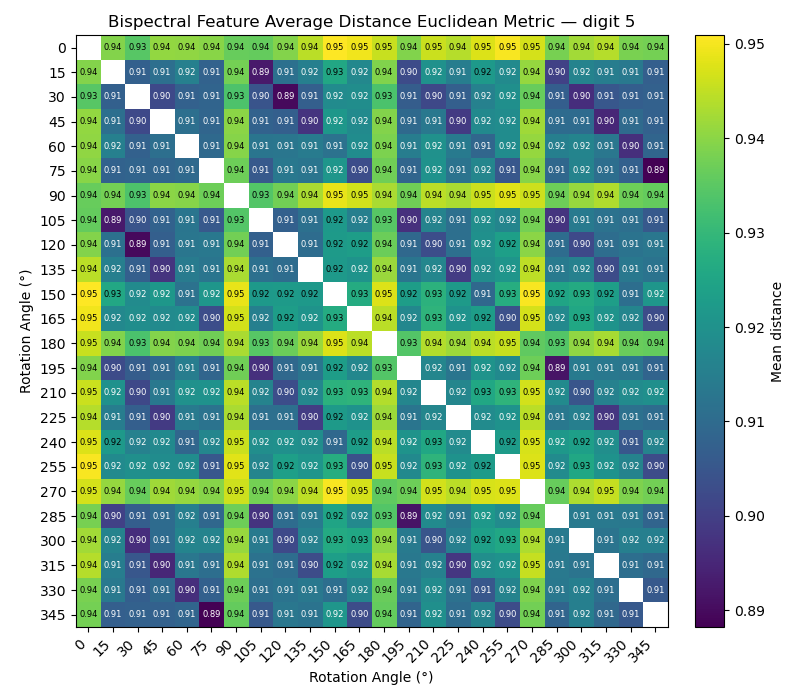} &
        \includegraphics[width=0.18\linewidth]{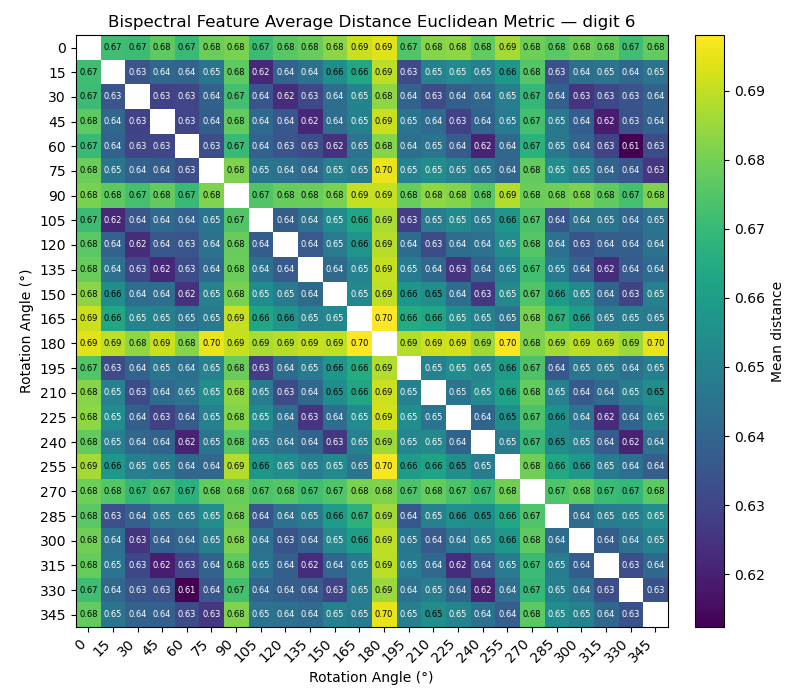} &
        \includegraphics[width=0.18\linewidth]{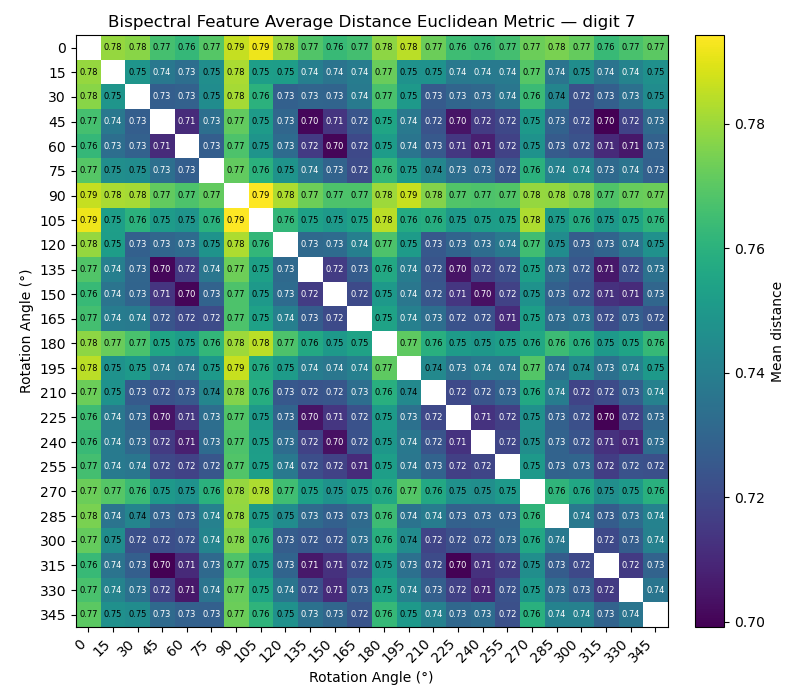} &
        \includegraphics[width=0.18\linewidth]{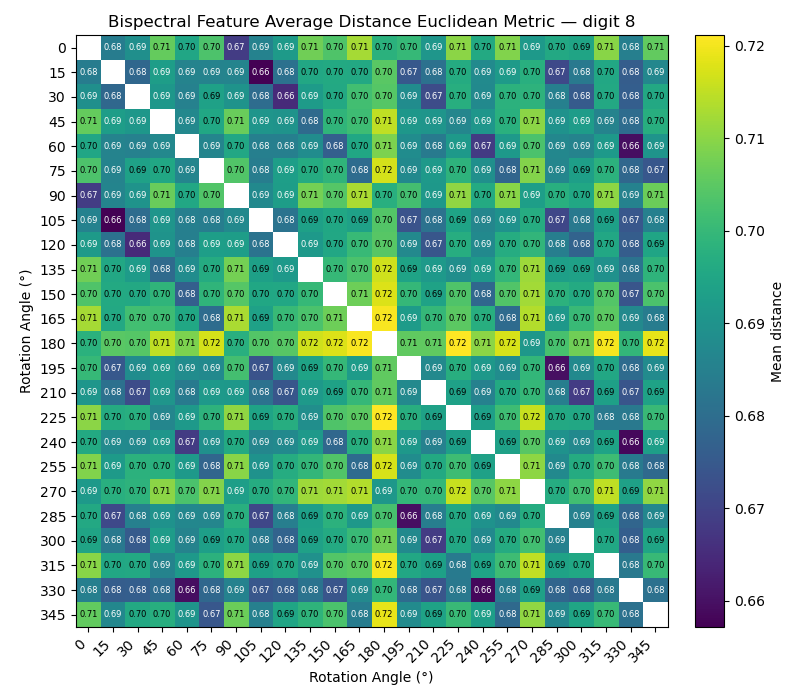} &
        \includegraphics[width=0.18\linewidth]{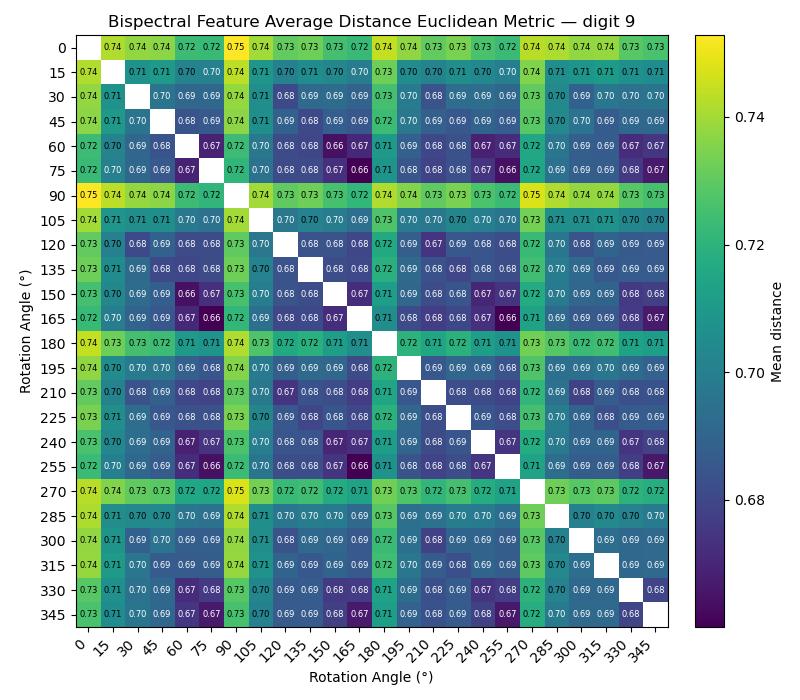} 
      \end{tabular}
    }
  }
\end{figure}

\begin{figure}[htbp]
\floatconts
  {fig:l1-per-digit}
  {\caption{Average distance between bispectral and pixel representations of rotated \textsc{mnist} images using $L_1$ norm.}}
  {%
    \subfigure[Raw feature representations]{%
      \centering
      \begin{tabular}{ccccc}
        \includegraphics[width=0.18\linewidth]{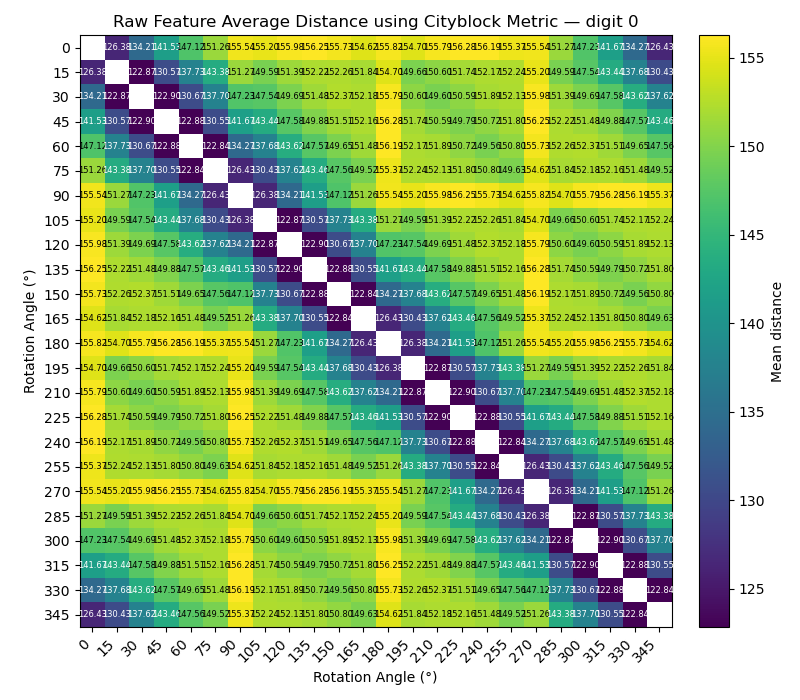} &
        \includegraphics[width=0.18\linewidth]{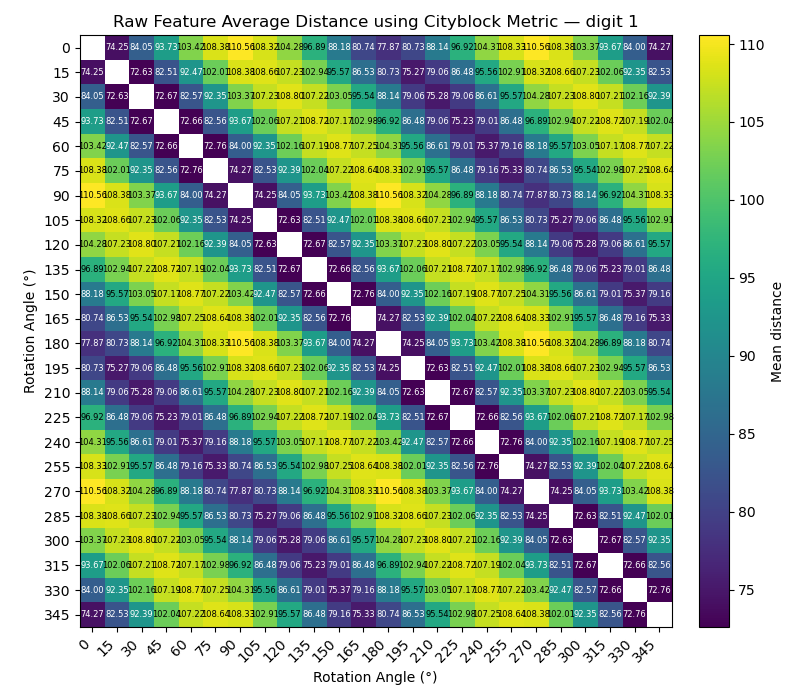} &
        \includegraphics[width=0.18\linewidth]{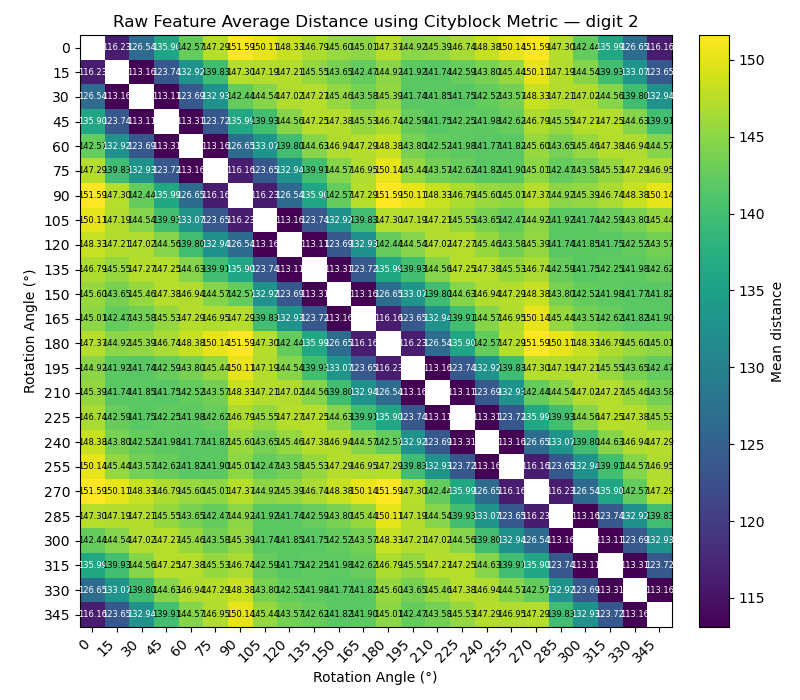} &
        \includegraphics[width=0.18\linewidth]{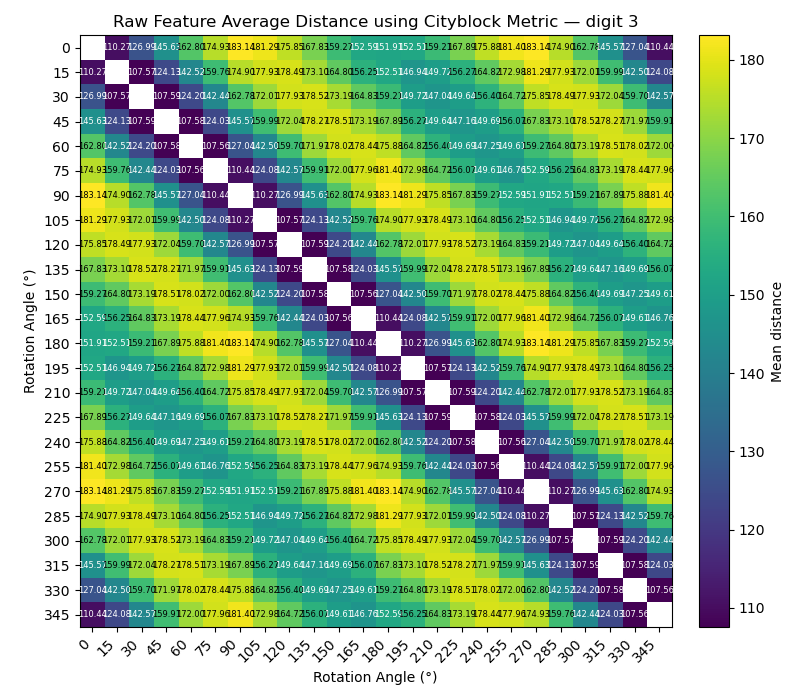} &
        \includegraphics[width=0.18\linewidth]{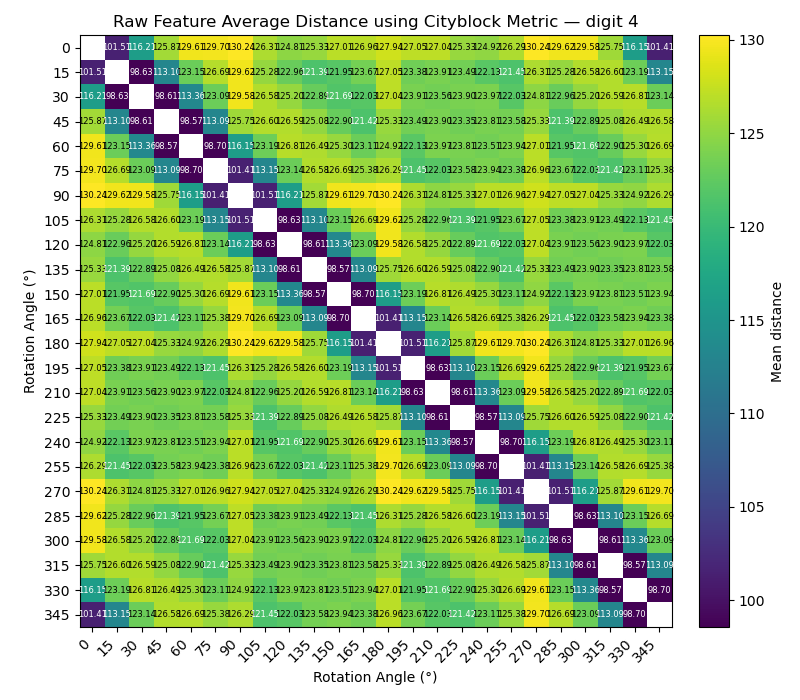} \\
        \includegraphics[width=0.18\linewidth]{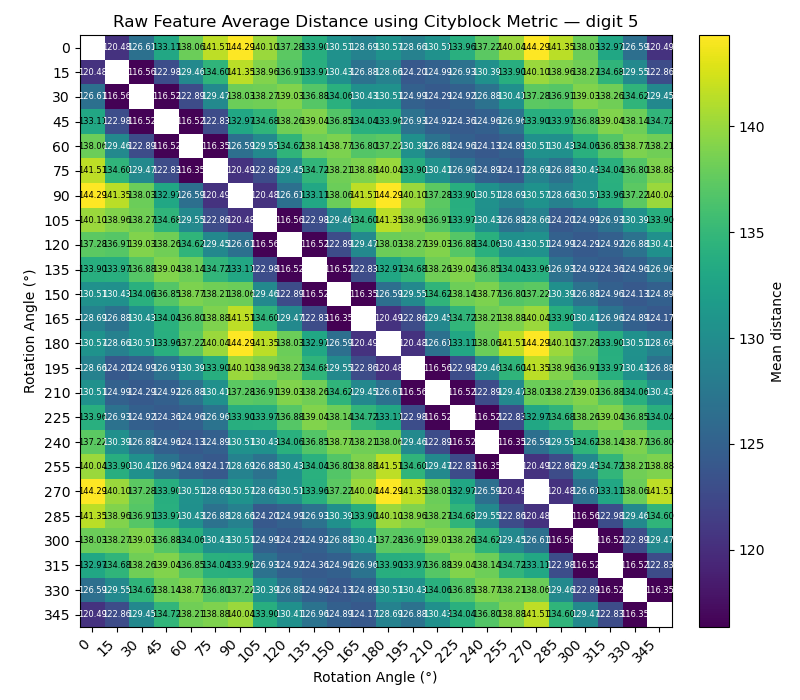} &
        \includegraphics[width=0.18\linewidth]{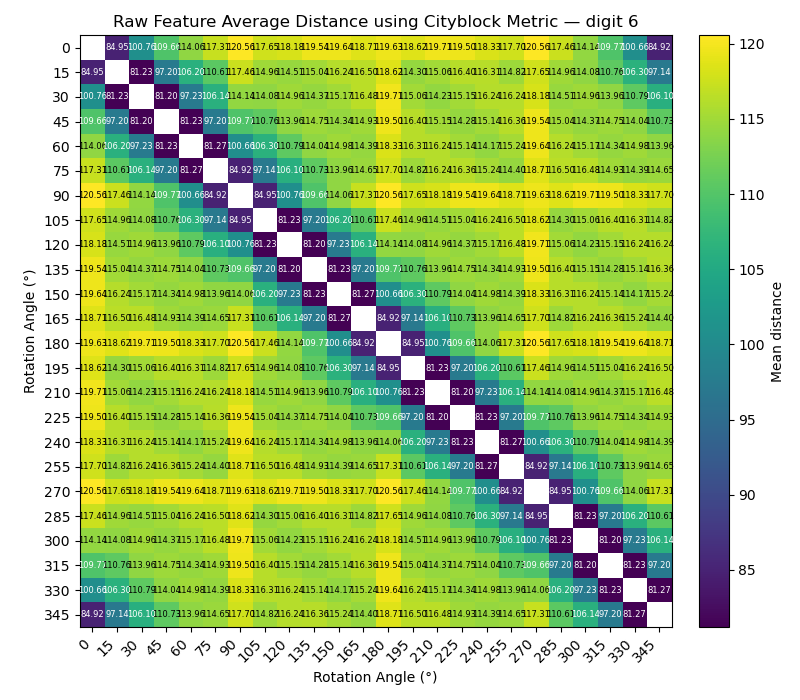} &
        \includegraphics[width=0.18\linewidth]{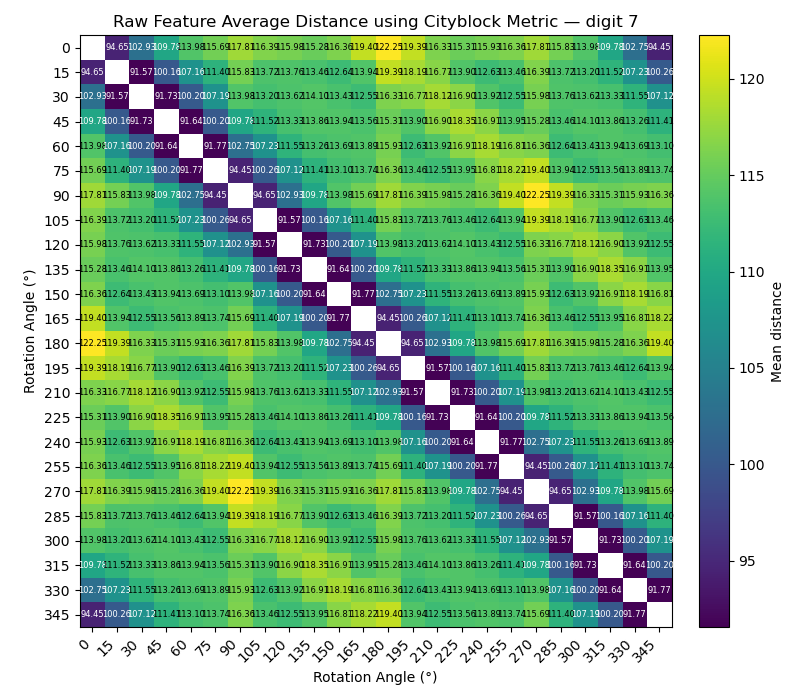} &
        \includegraphics[width=0.18\linewidth]{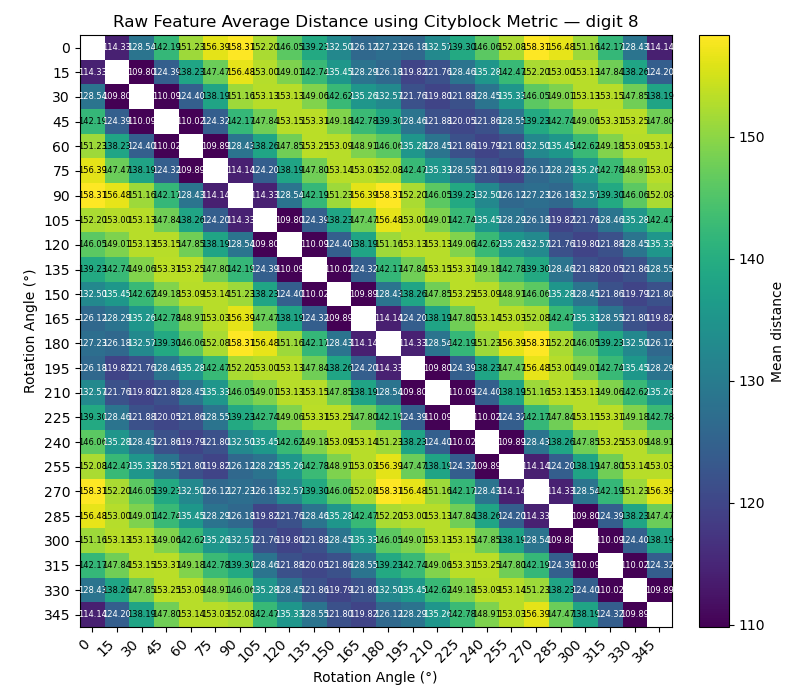} &
        \includegraphics[width=0.18\linewidth]{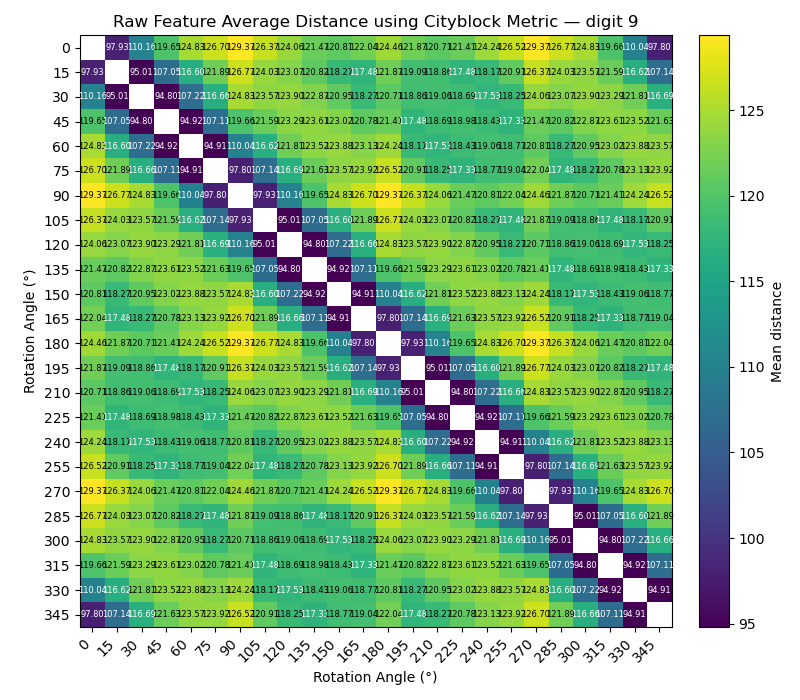}
      \end{tabular}
    }\quad
    \subfigure[Bispectral feature representations]{%
      \centering
      \begin{tabular}{ccccc}
        \includegraphics[width=0.18\linewidth]{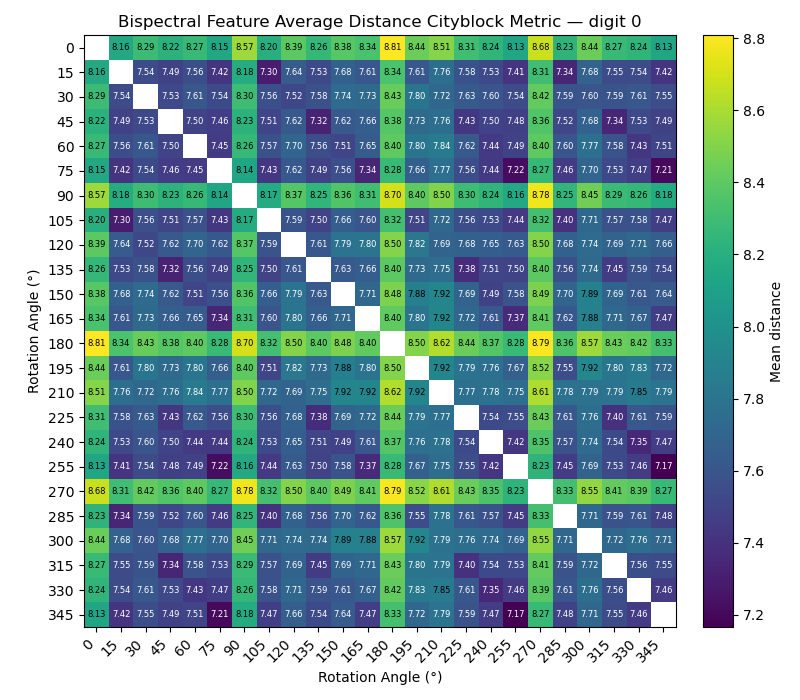} &
        \includegraphics[width=0.18\linewidth]{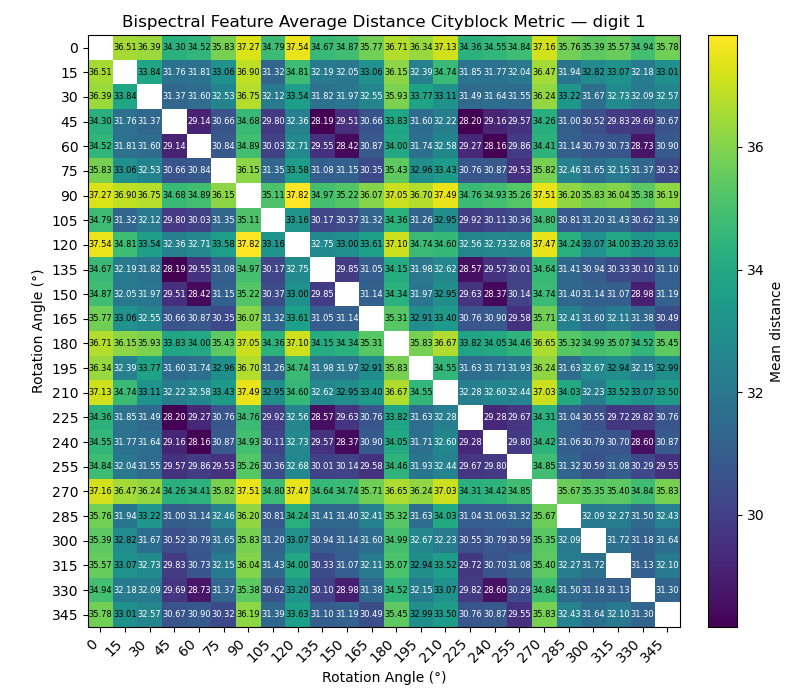} &
        \includegraphics[width=0.18\linewidth]{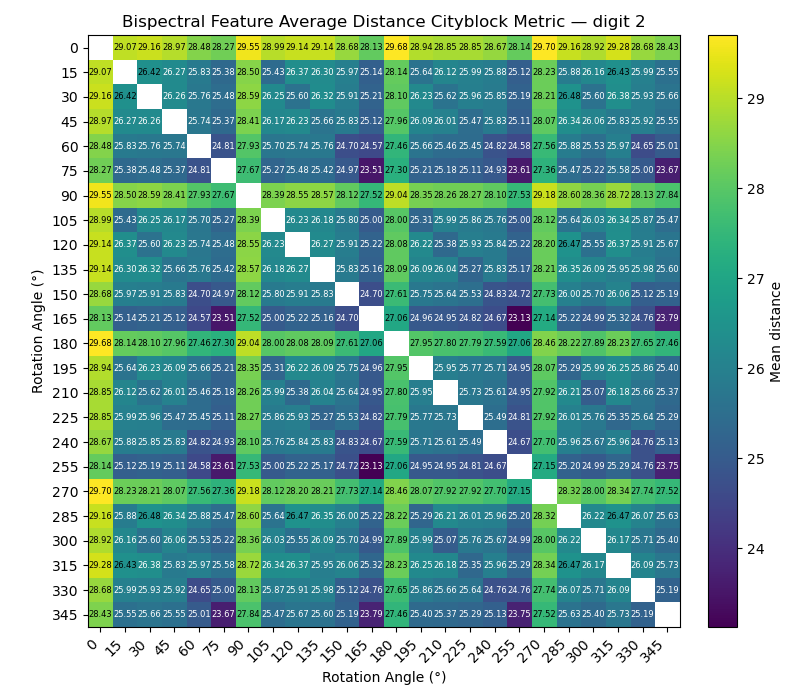} &
        \includegraphics[width=0.18\linewidth]{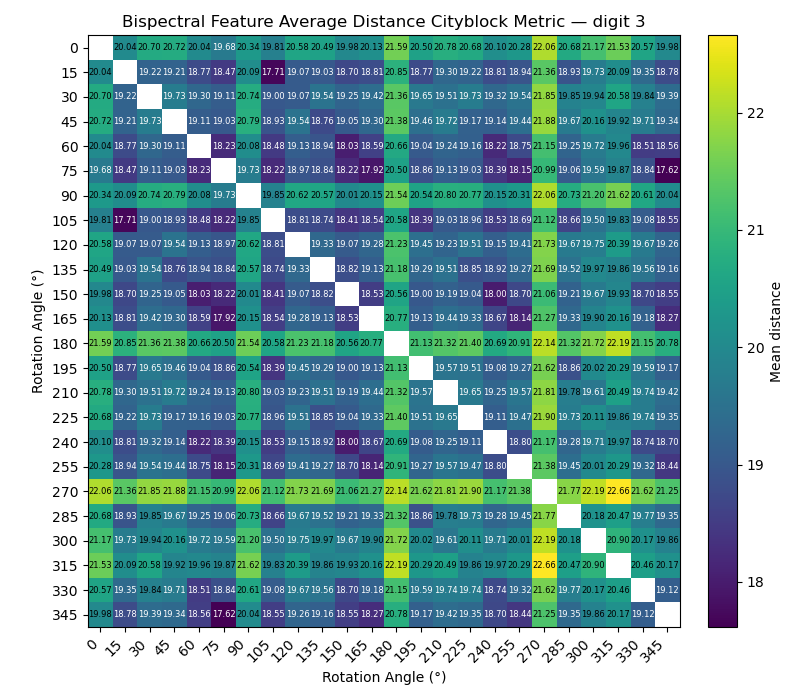} &
        \includegraphics[width=0.18\linewidth]{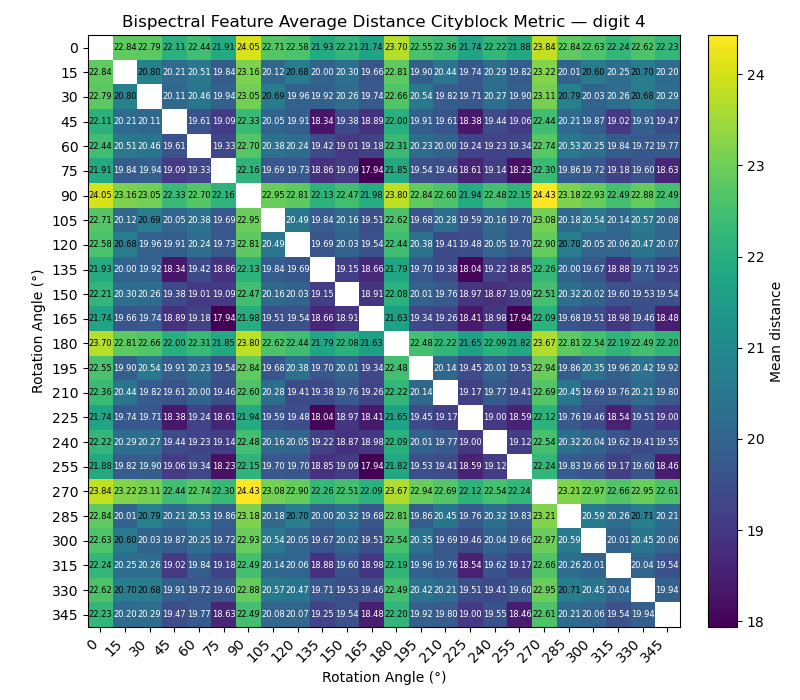} \\
        \includegraphics[width=0.18\linewidth]{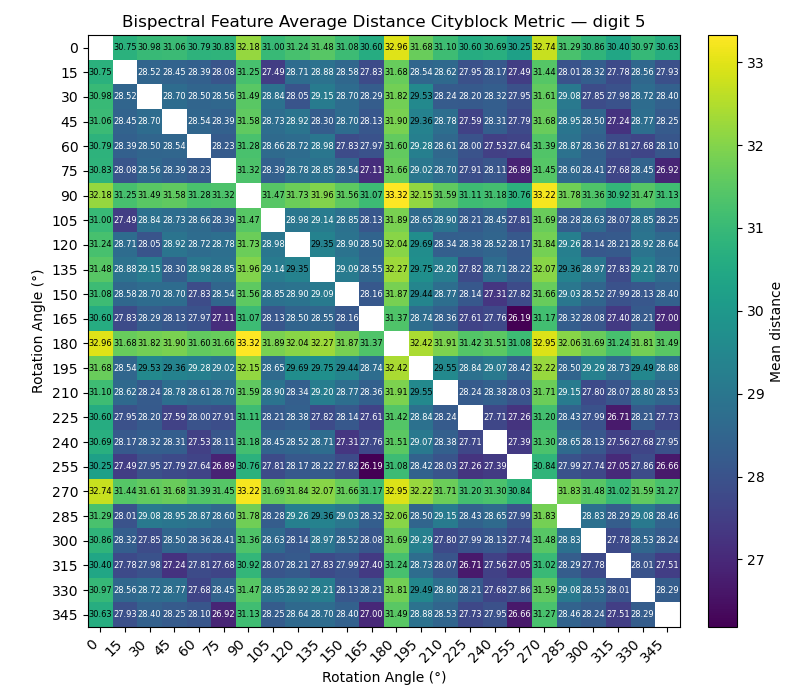} &
        \includegraphics[width=0.18\linewidth]{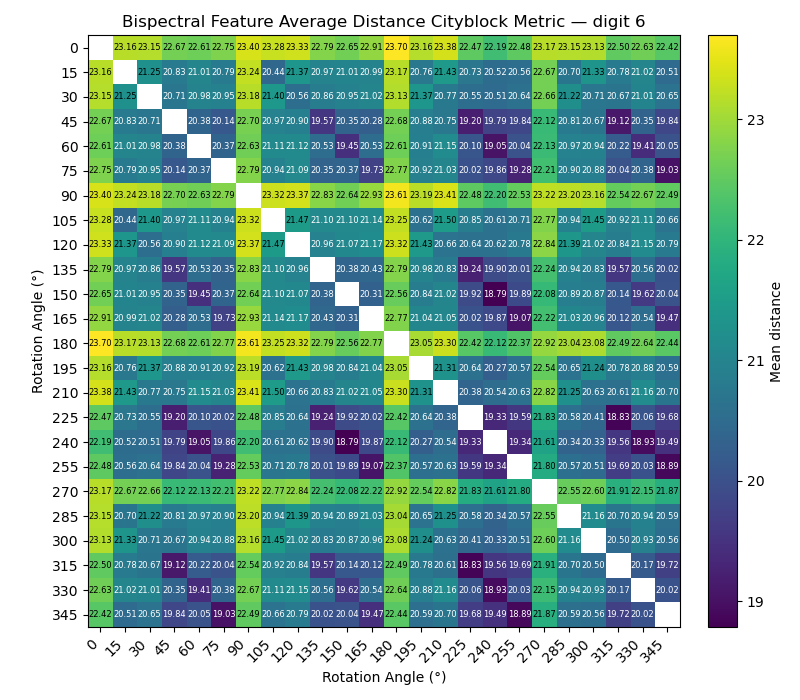} &
        \includegraphics[width=0.18\linewidth]{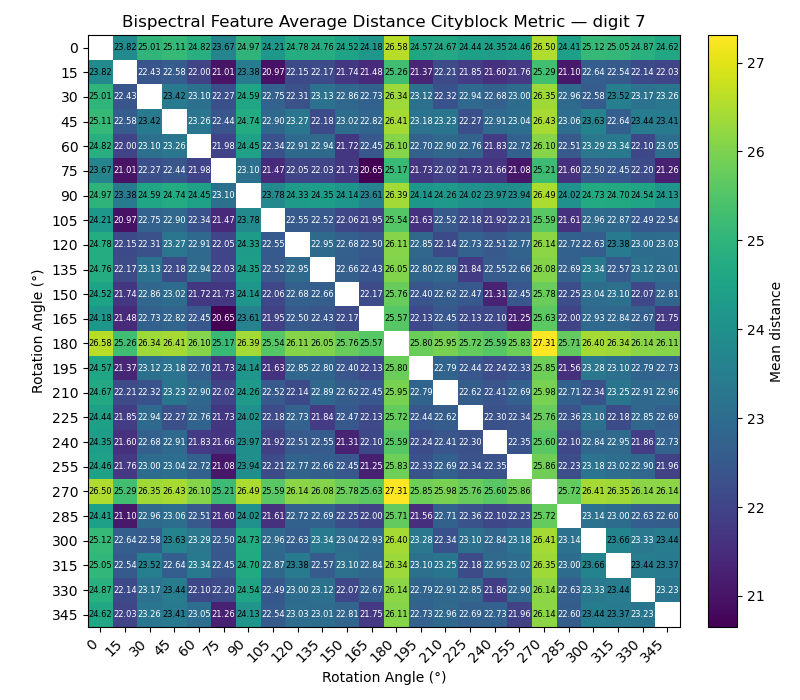} &
        \includegraphics[width=0.18\linewidth]{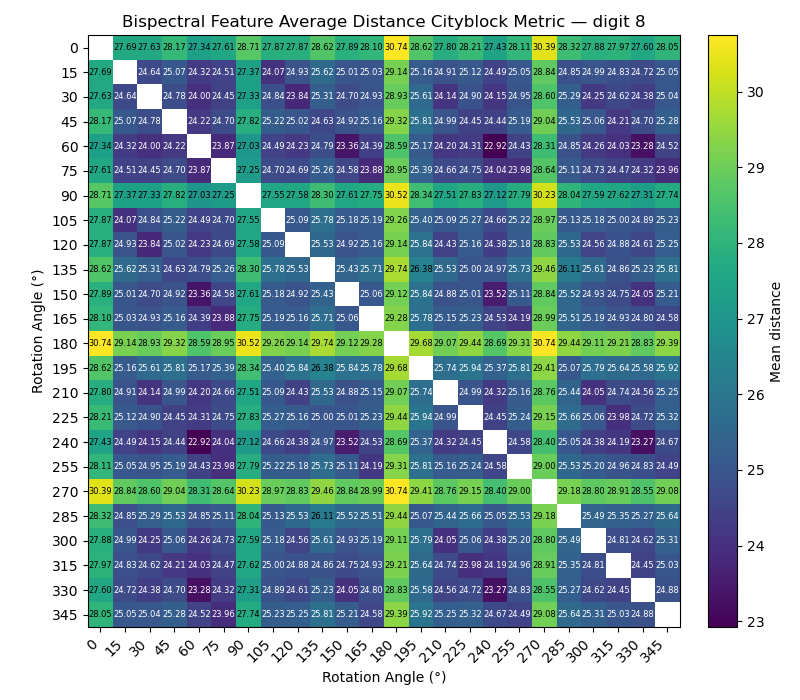} &
        \includegraphics[width=0.18\linewidth]{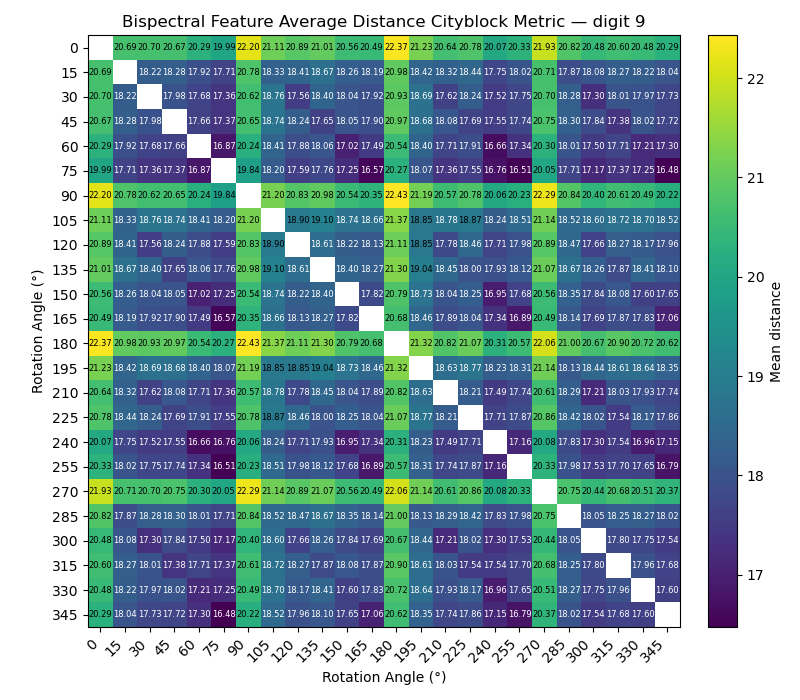}
      \end{tabular}
    }
  }
\end{figure}

\begin{figure}[htbp]
\floatconts
  {fig:sqeuclidean-per-digit}
  {\caption{Average distance between bispectral and pixel representations of rotated \textsc{mnist} images using squared euclidean distance.}}
  {%
    \subfigure[Raw feature representations]{%
      \centering
      \begin{tabular}{ccccc}
        \includegraphics[width=0.18\linewidth]{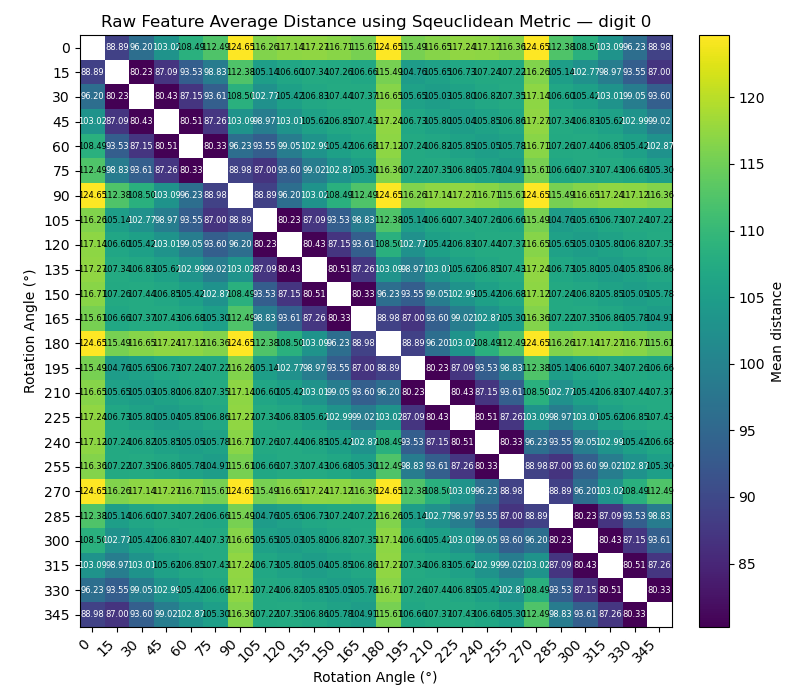} &
        \includegraphics[width=0.18\linewidth]{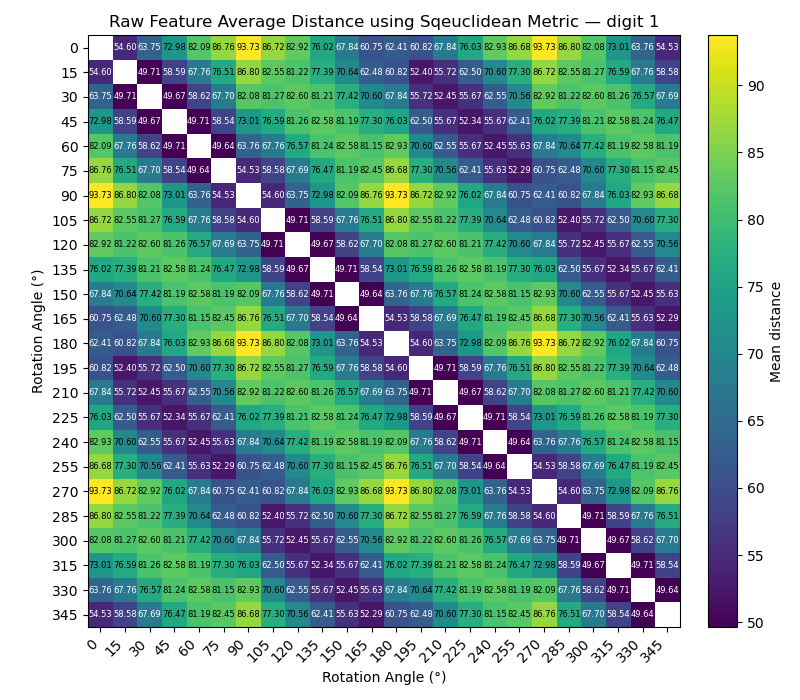} &
        \includegraphics[width=0.18\linewidth]{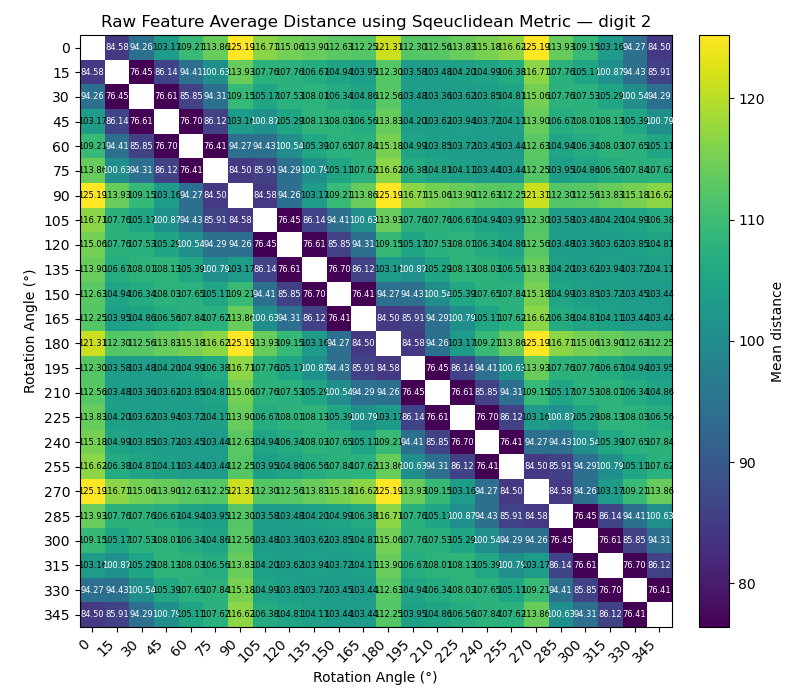} &
        \includegraphics[width=0.18\linewidth]{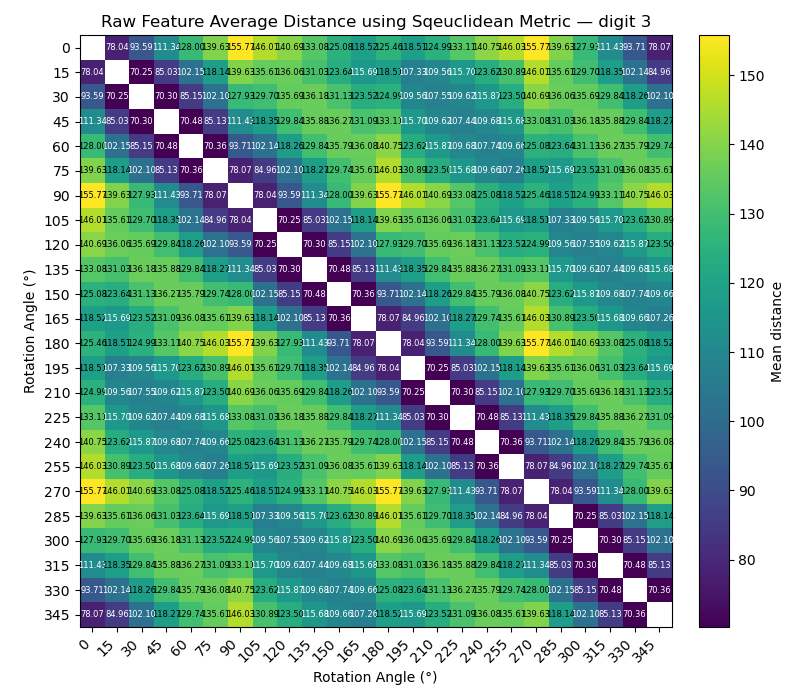} &
        \includegraphics[width=0.18\linewidth]{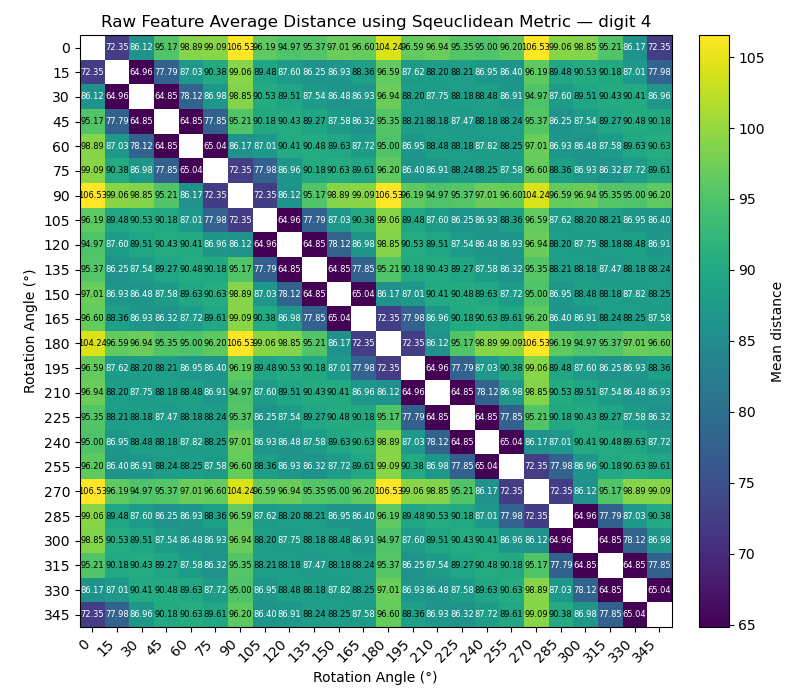} \\
        \includegraphics[width=0.18\linewidth]{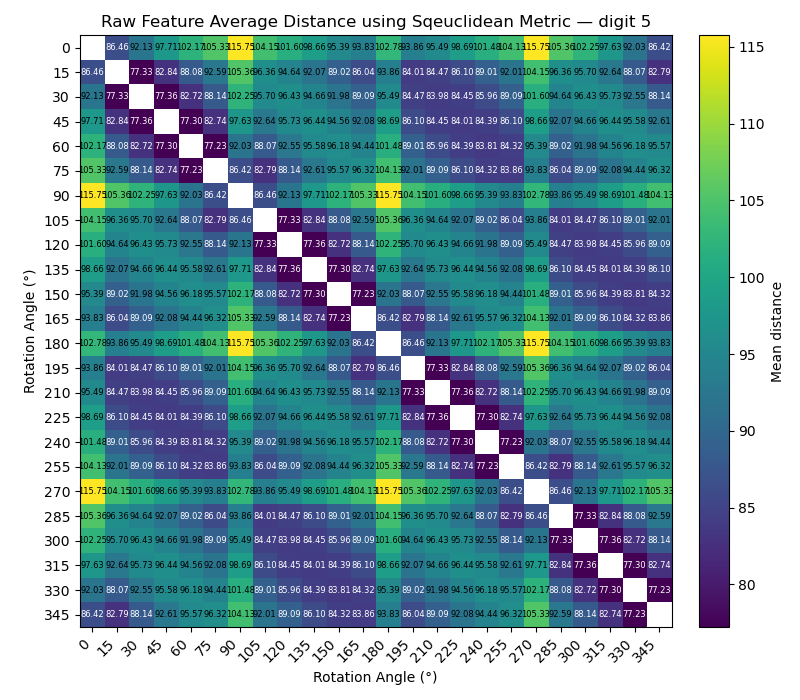} &
        \includegraphics[width=0.18\linewidth]{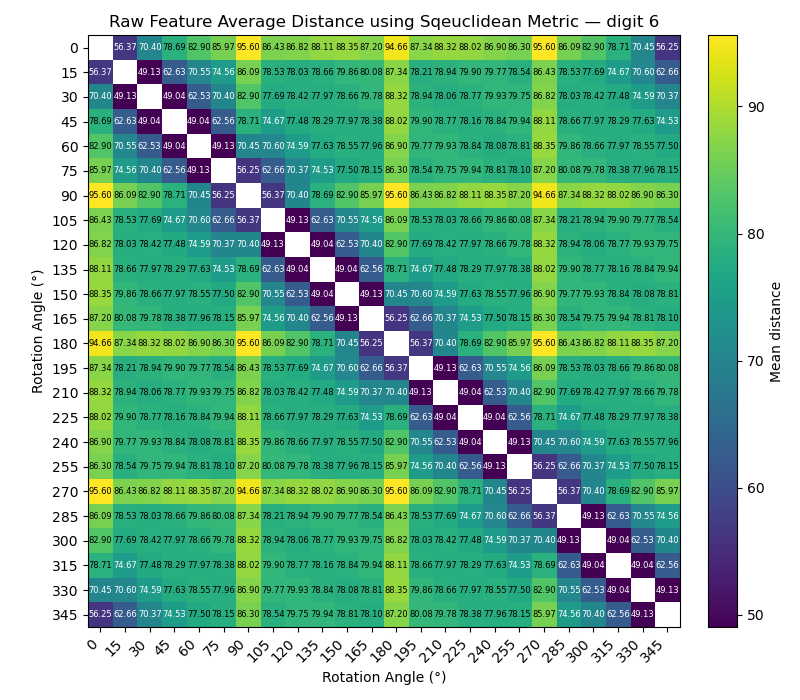} &
        \includegraphics[width=0.18\linewidth]{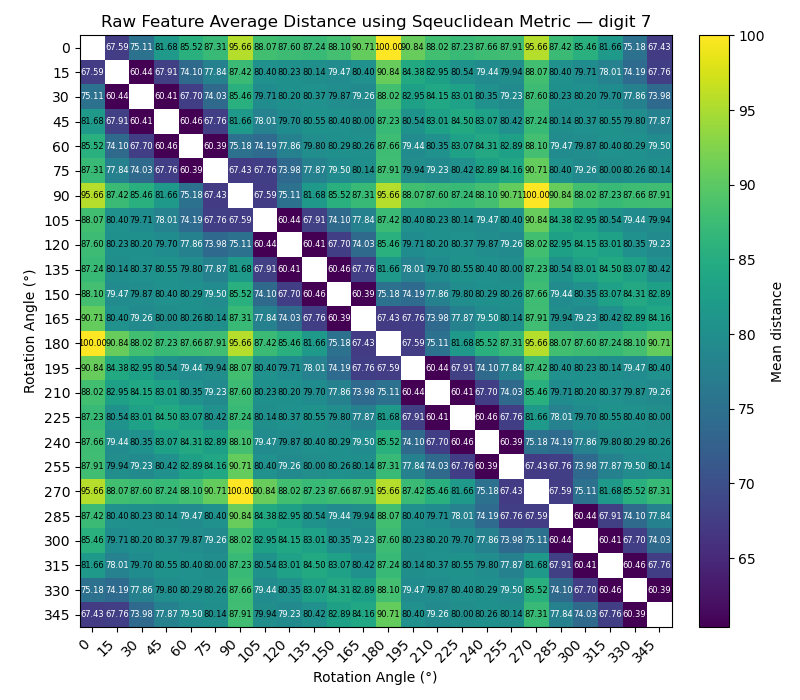} &
        \includegraphics[width=0.18\linewidth]{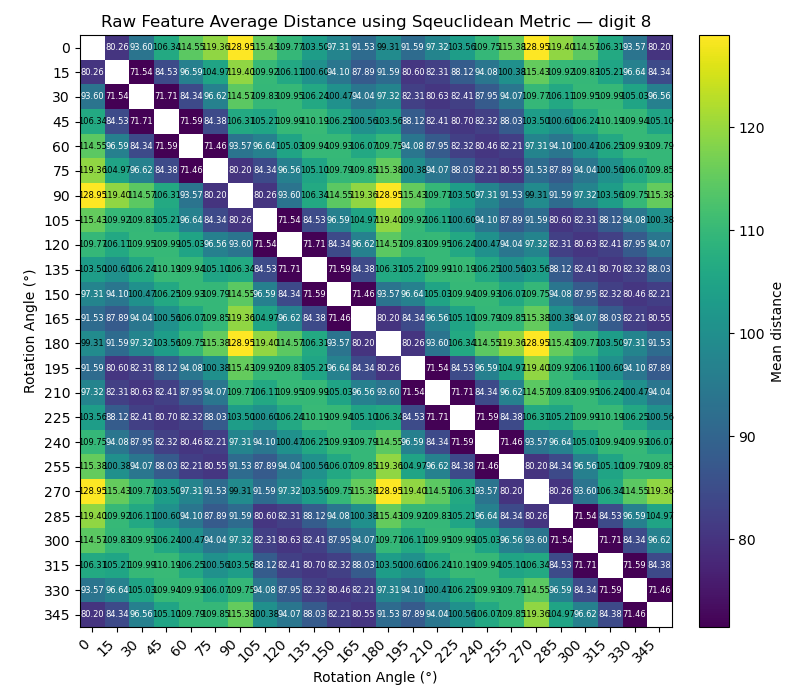} &
        \includegraphics[width=0.18\linewidth]{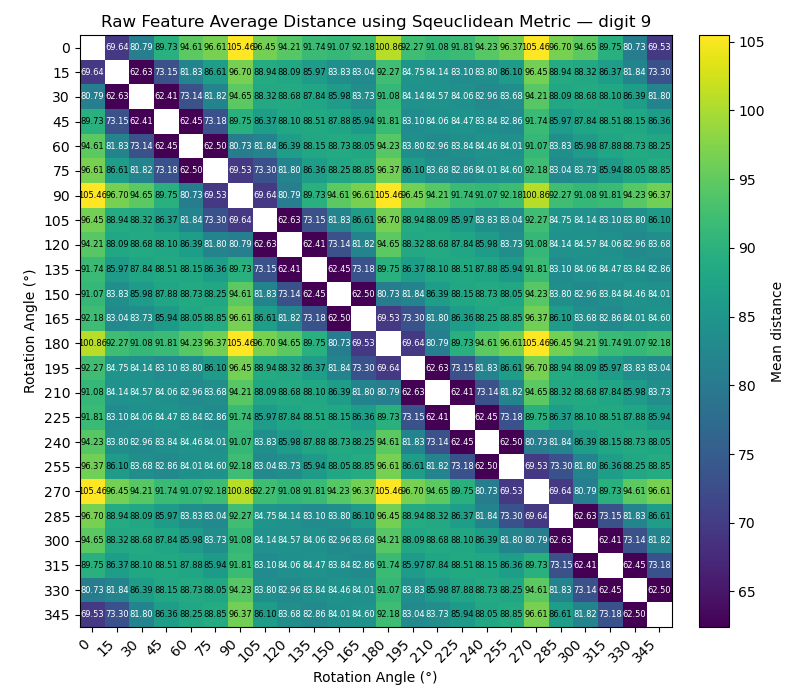}
      \end{tabular}
    }\quad
    \subfigure[Bispectral feature representations]{%
      \centering
      \begin{tabular}{ccccc}
        \includegraphics[width=0.18\linewidth]{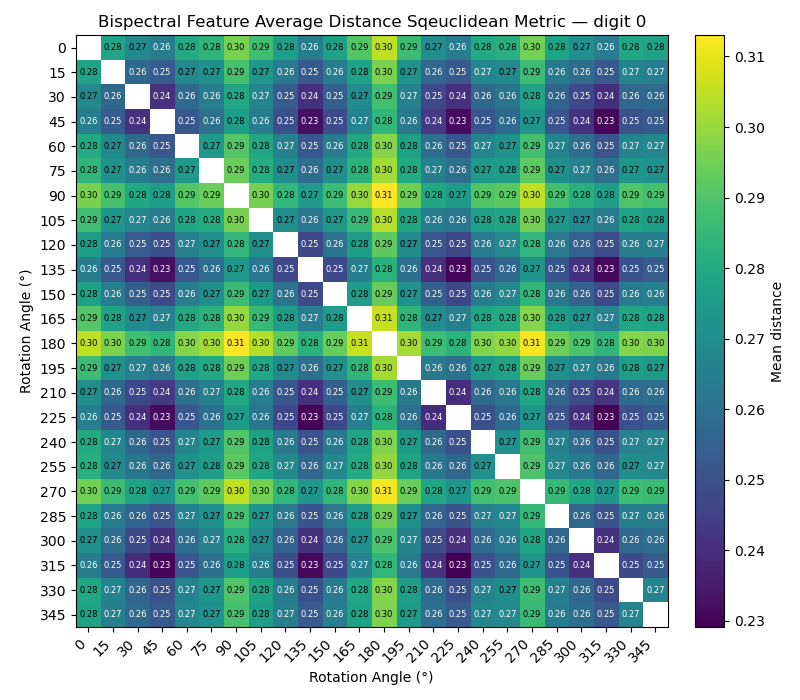} &
        \includegraphics[width=0.18\linewidth]{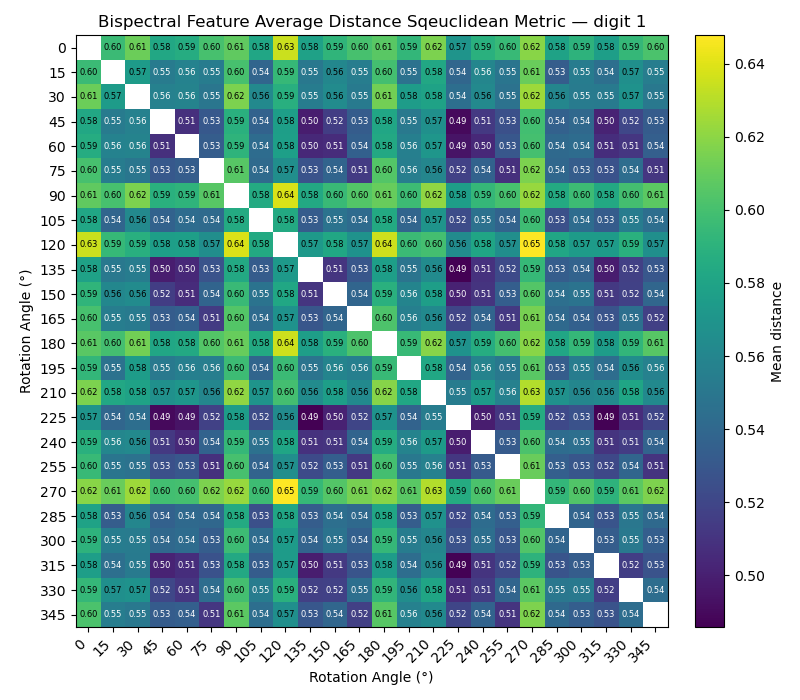} &
        \includegraphics[width=0.18\linewidth]{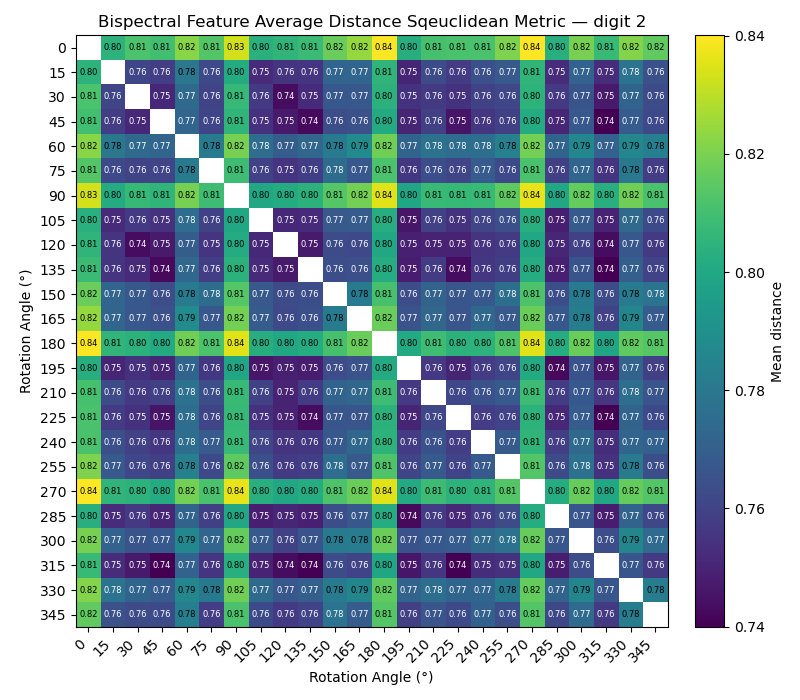} &
        \includegraphics[width=0.18\linewidth]{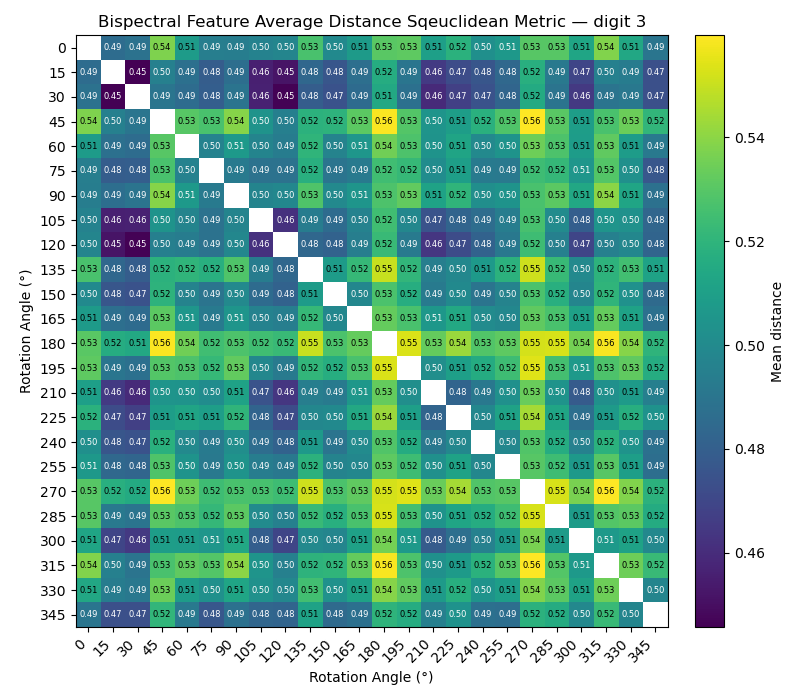} &
        \includegraphics[width=0.18\linewidth]{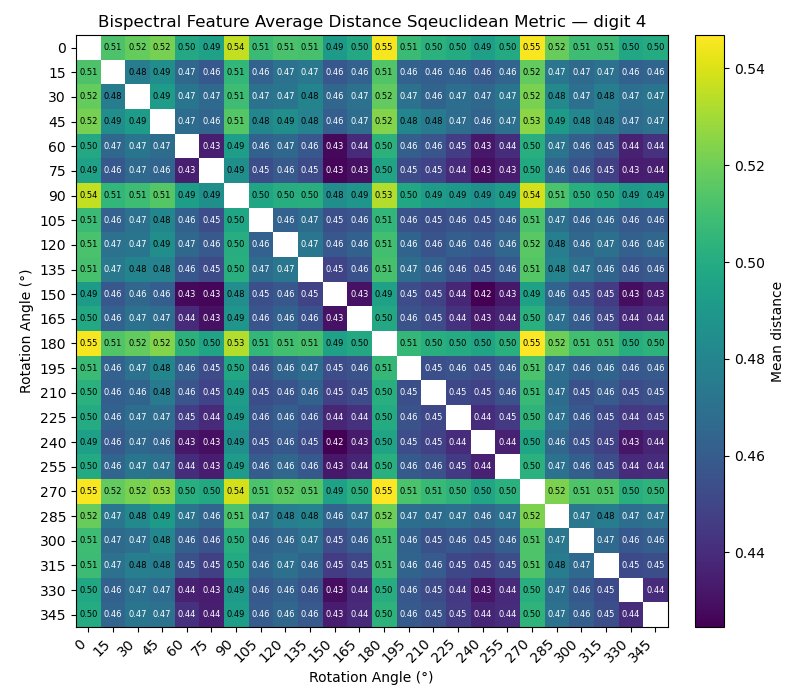} \\
        \includegraphics[width=0.18\linewidth]{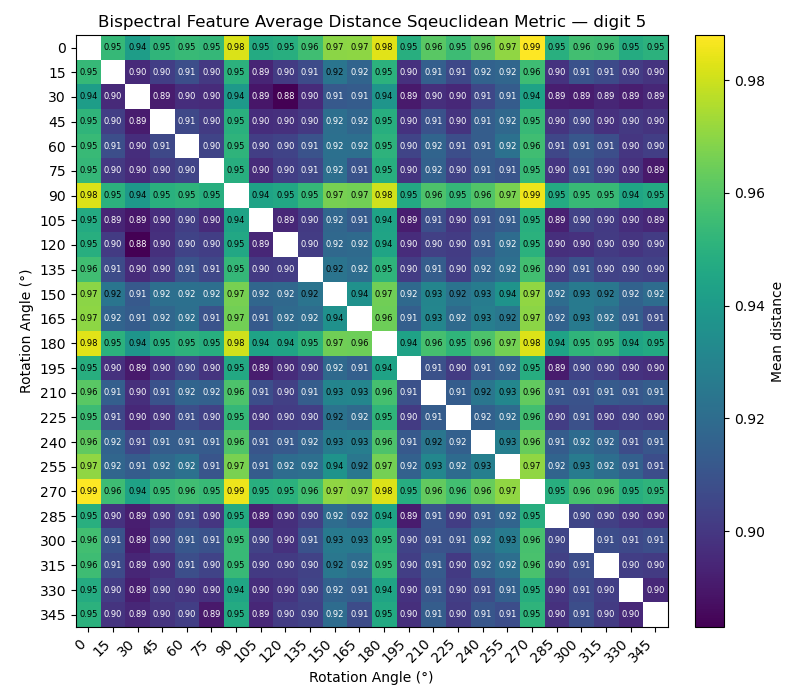} &
        \includegraphics[width=0.18\linewidth]{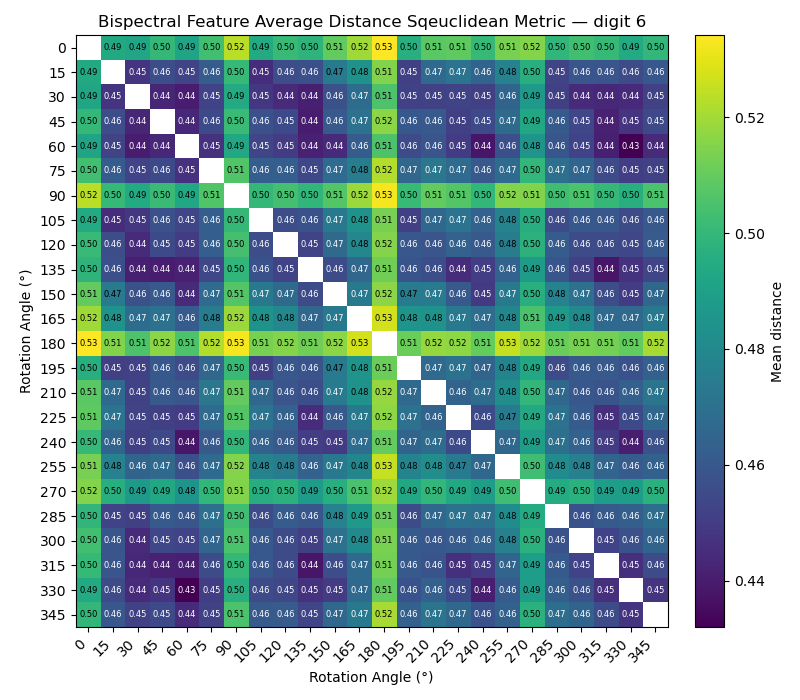} &
        \includegraphics[width=0.18\linewidth]{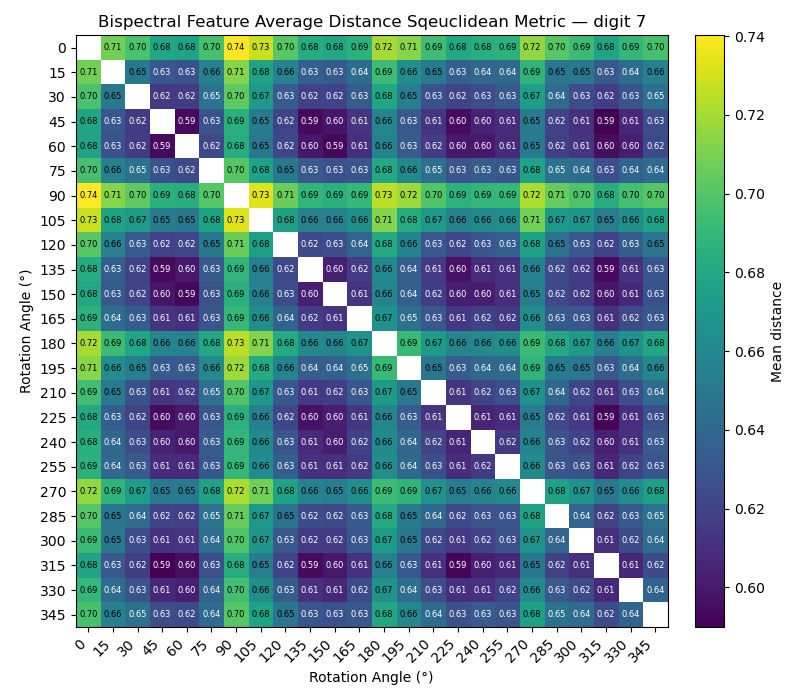} &
        \includegraphics[width=0.18\linewidth]{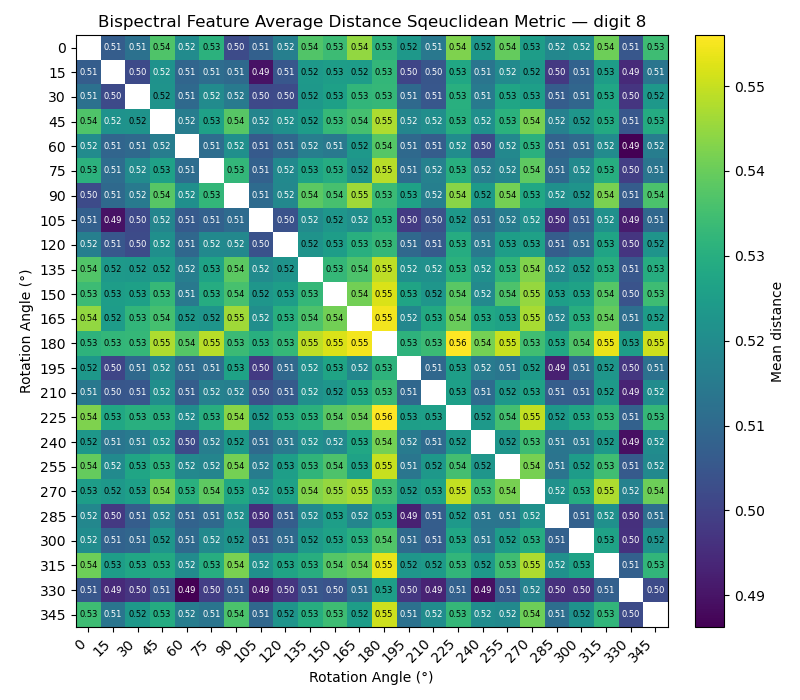} &
        \includegraphics[width=0.18\linewidth]{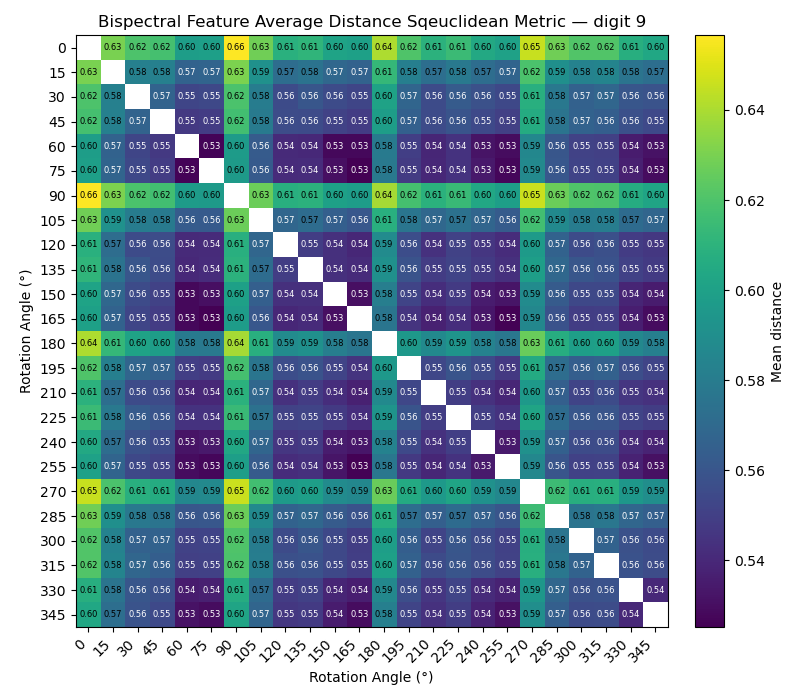}
      \end{tabular}
    }
  }
\end{figure}

\begin{figure}[htbp]
\floatconts
  {fig:cos-per-digit}
  {\caption{Average distance between bispectral and pixel representations of rotated \textsc{mnist} images using cosine similarity.}}
  {%
    \subfigure[Raw feature representations]{%
      \centering
      \begin{tabular}{ccccc}
        \includegraphics[width=0.18\linewidth]{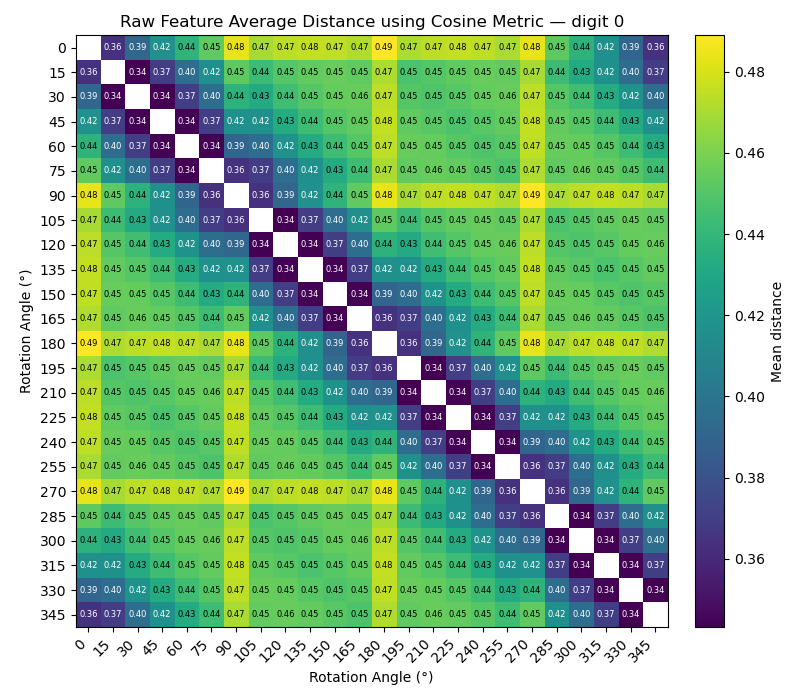} &
        \includegraphics[width=0.18\linewidth]{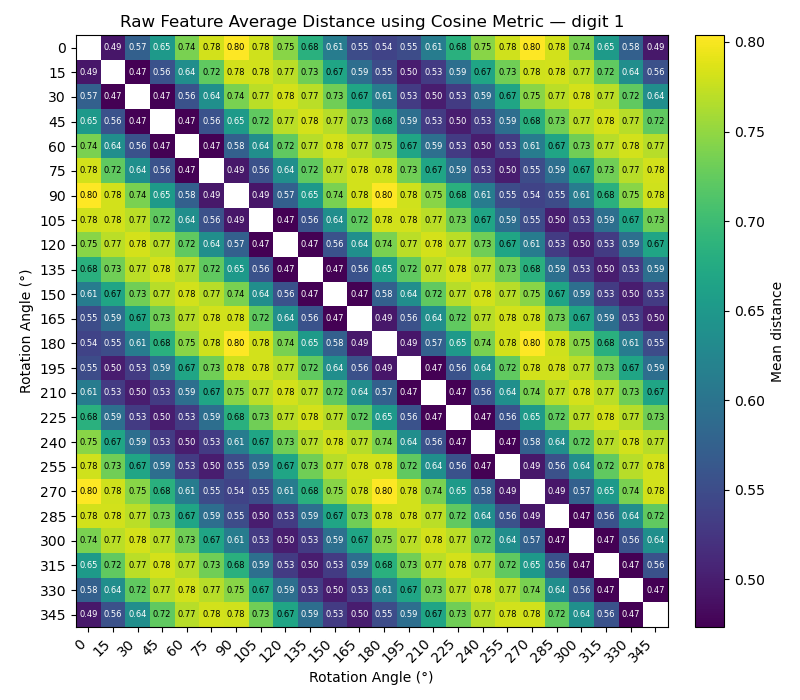} &
        \includegraphics[width=0.18\linewidth]{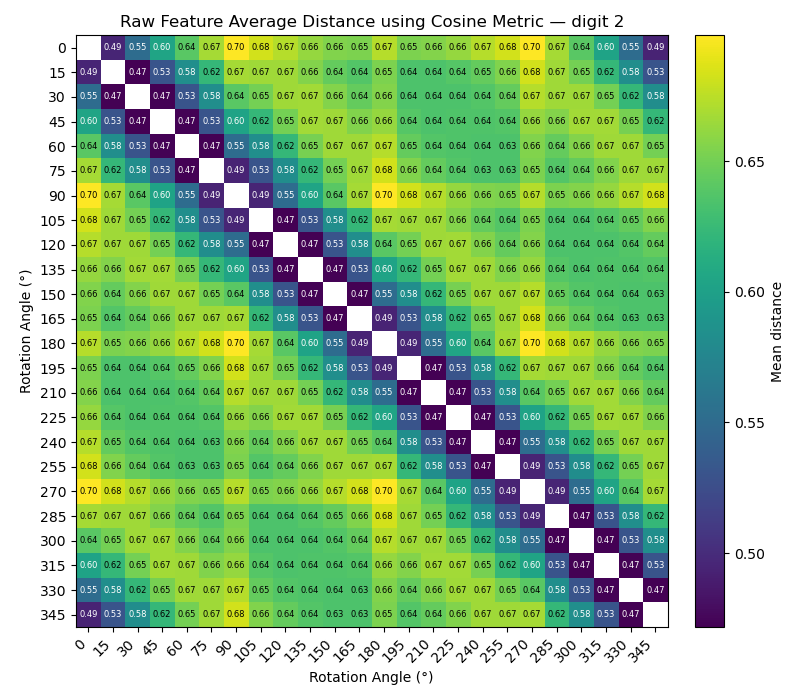} &
        \includegraphics[width=0.18\linewidth]{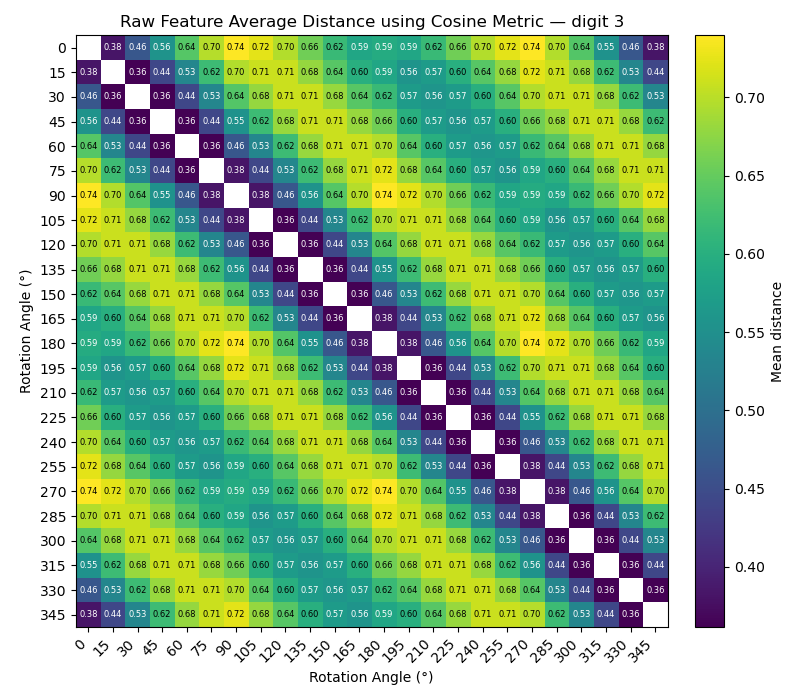} &
        \includegraphics[width=0.18\linewidth]{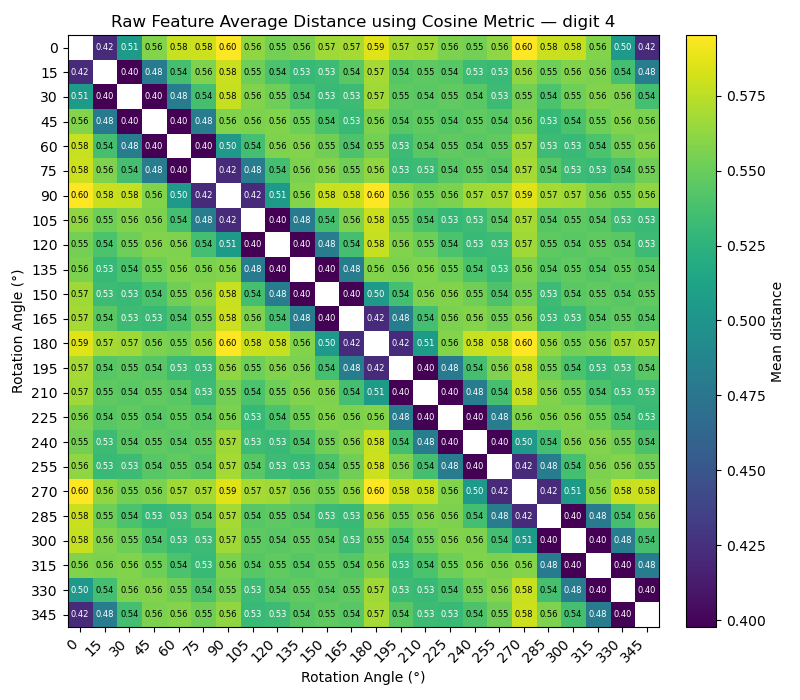} \\
        \includegraphics[width=0.18\linewidth]{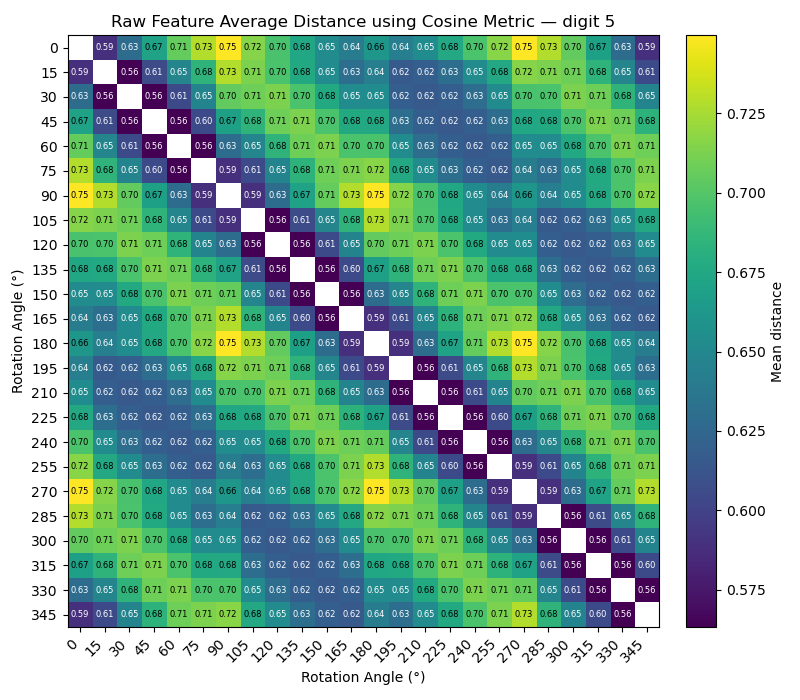} &
        \includegraphics[width=0.18\linewidth]{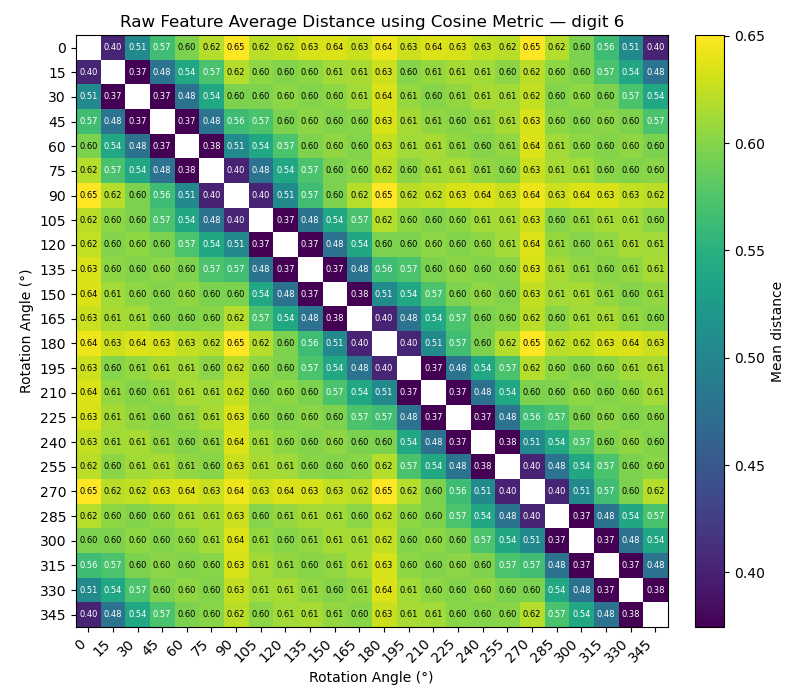} &
        \includegraphics[width=0.18\linewidth]{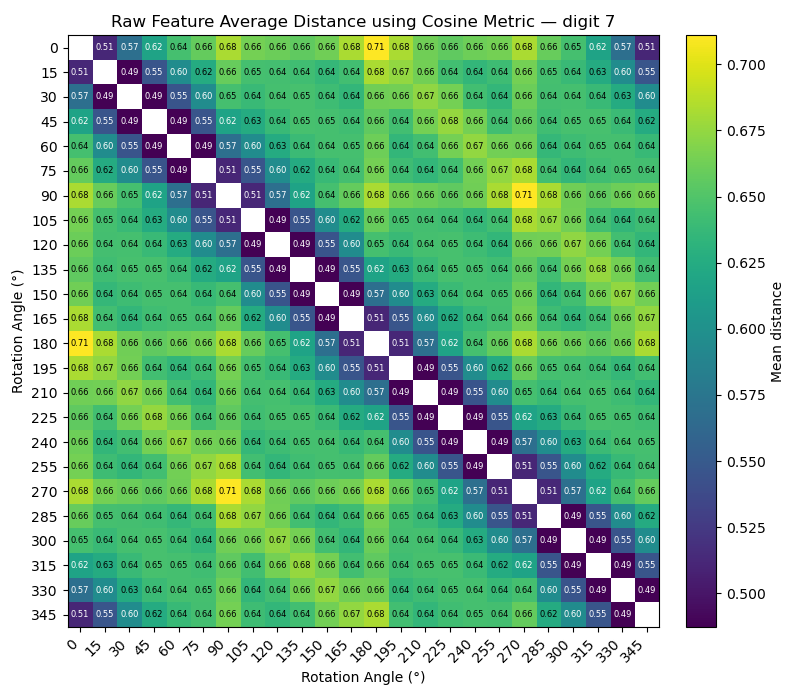} &
        \includegraphics[width=0.18\linewidth]{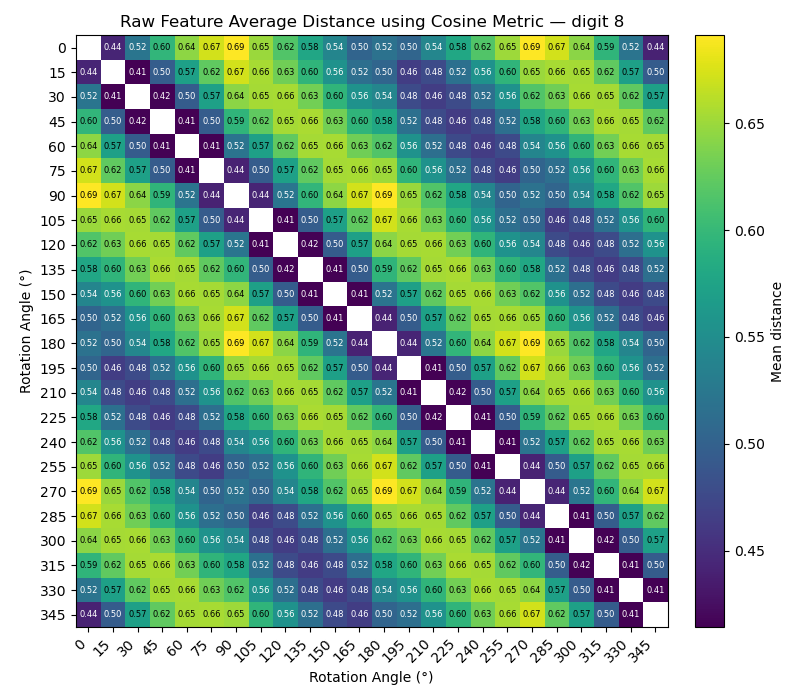} &
        \includegraphics[width=0.18\linewidth]{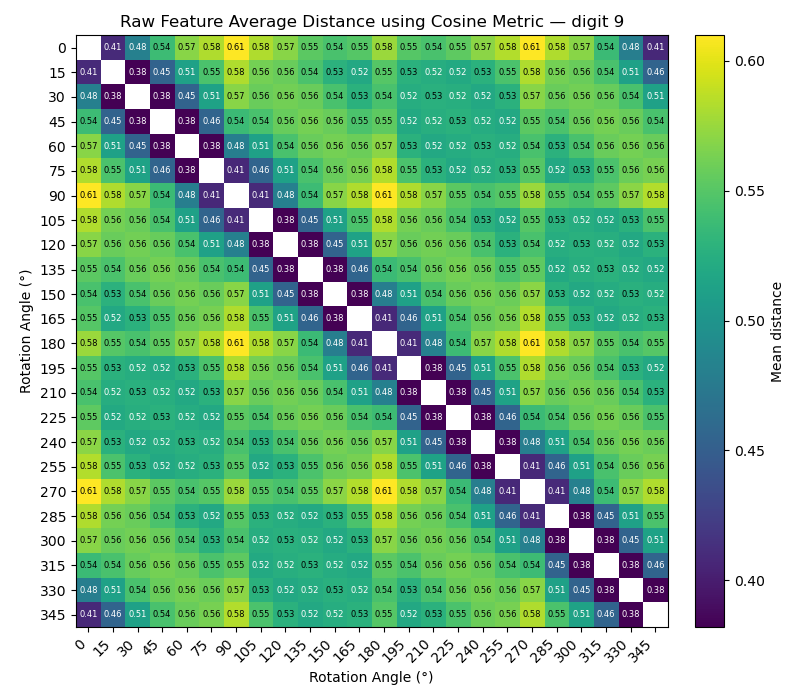}
      \end{tabular}
    }\quad
    \subfigure[Bispectral feature representations]{%
      \centering
      \begin{tabular}{ccccc}
        \includegraphics[width=0.18\linewidth]{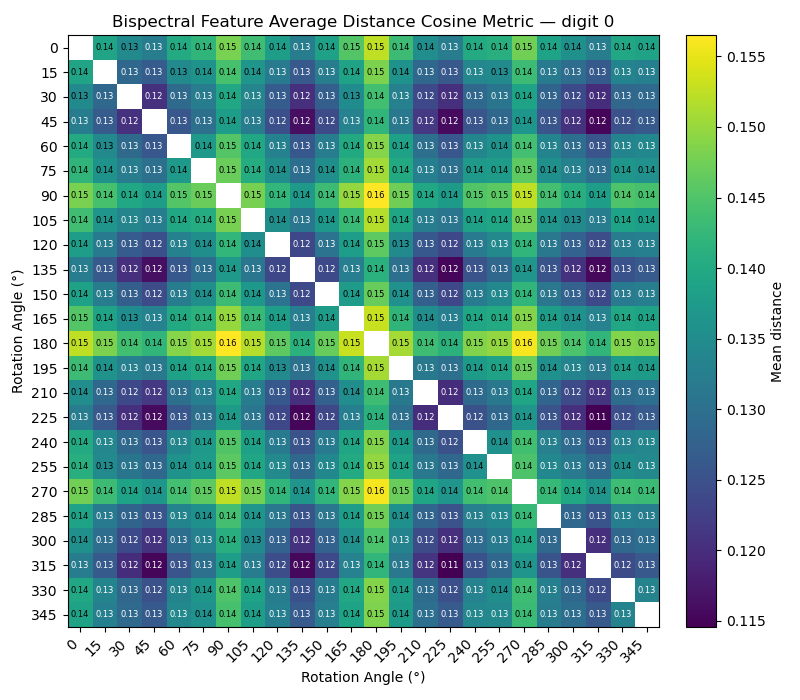} &
        \includegraphics[width=0.18\linewidth]{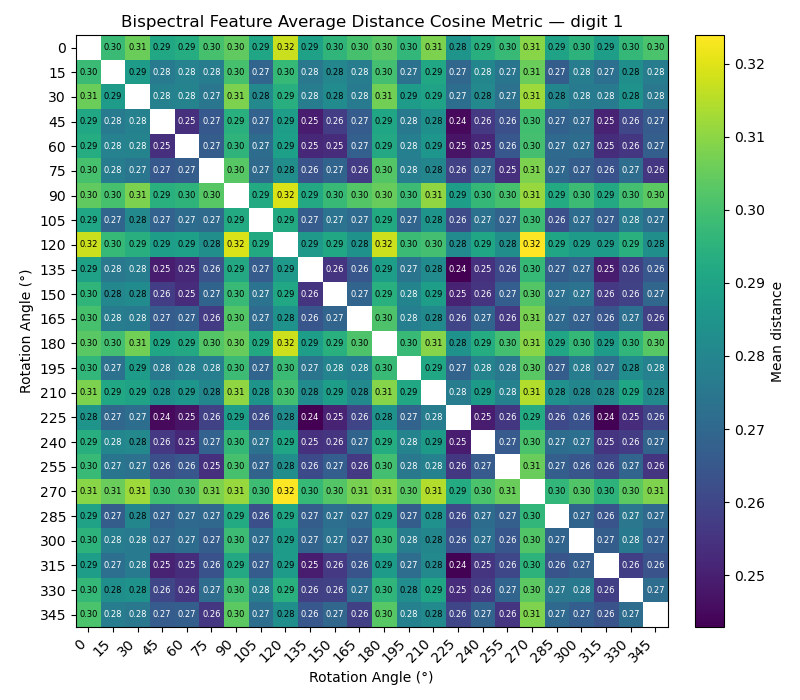} &
        \includegraphics[width=0.18\linewidth]{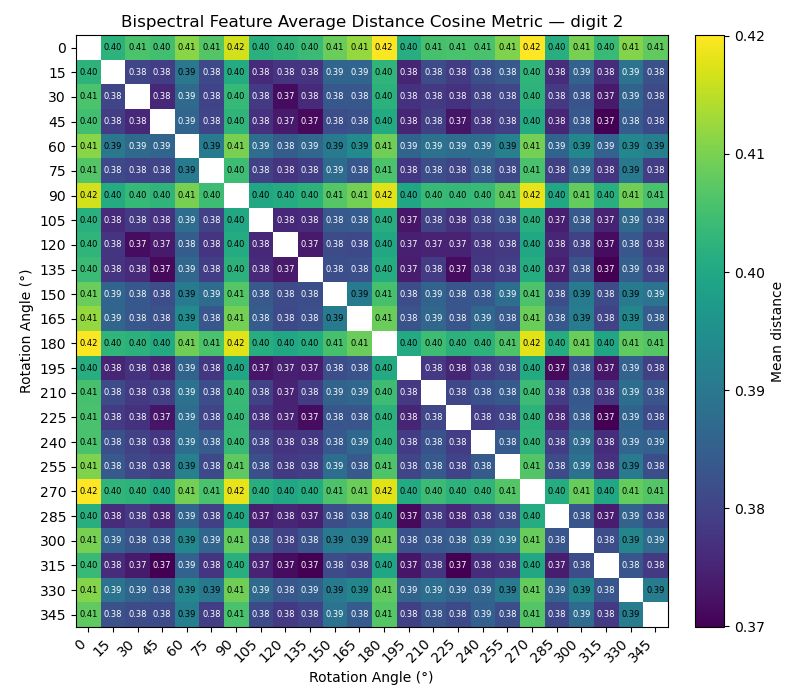} &
        \includegraphics[width=0.18\linewidth]{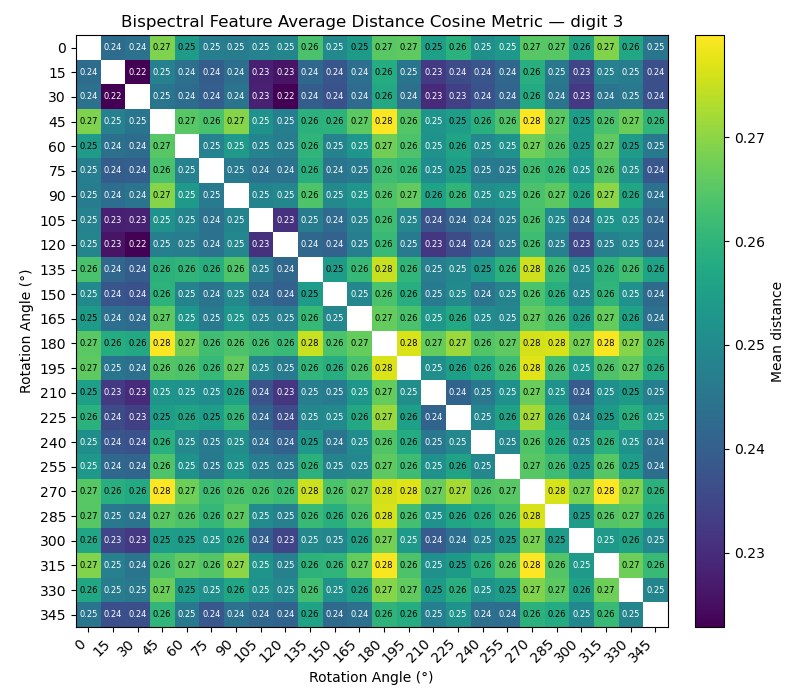} &
        \includegraphics[width=0.18\linewidth]{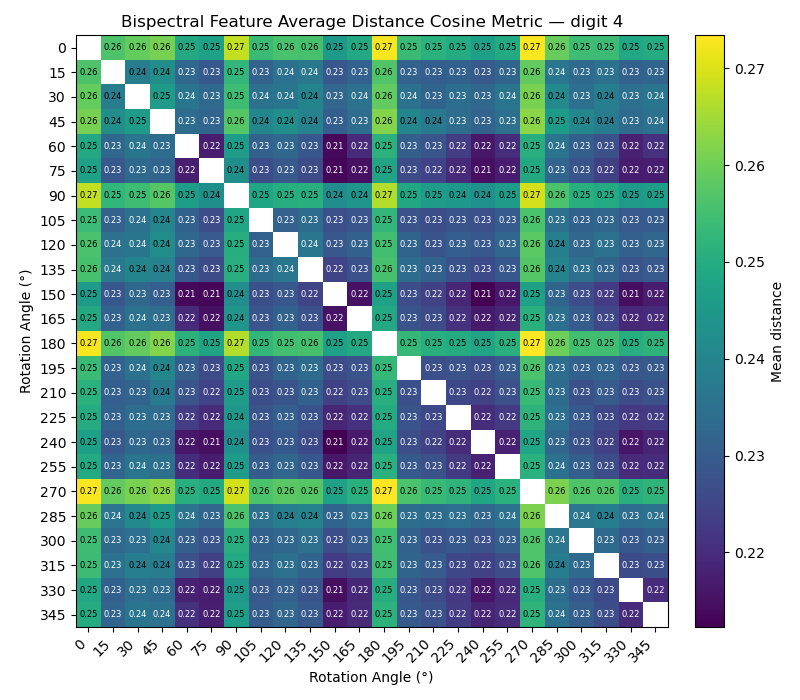} \\
        \includegraphics[width=0.18\linewidth]{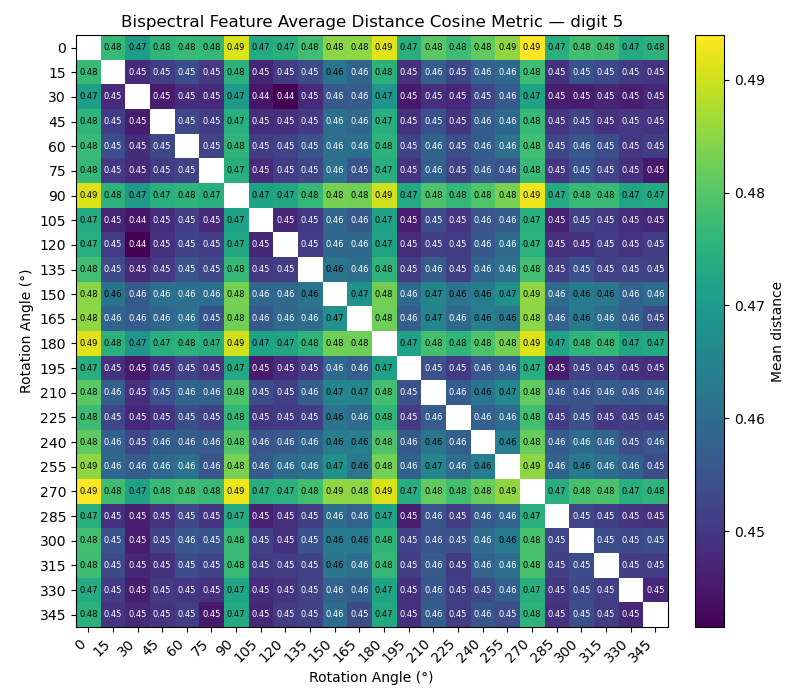} &
        \includegraphics[width=0.18\linewidth]{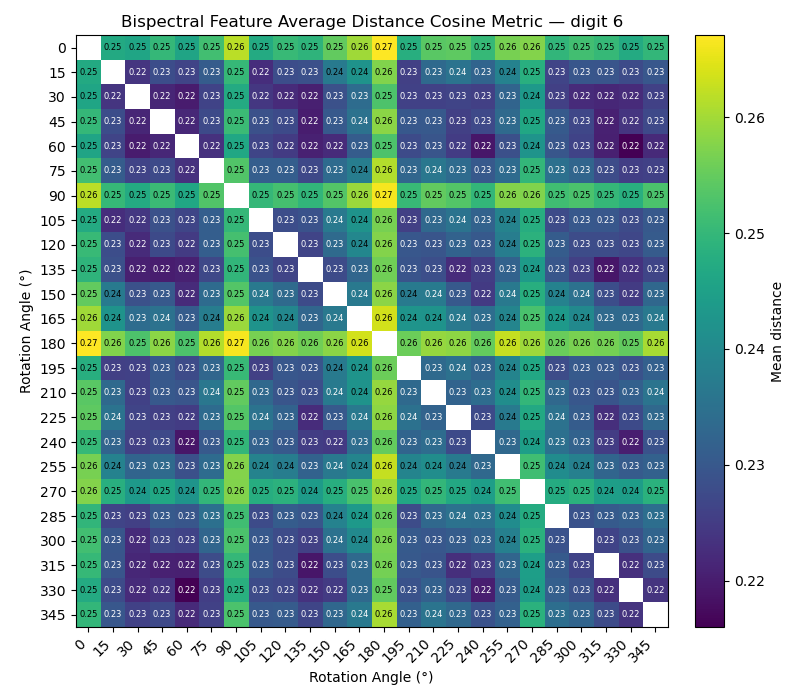} &
        \includegraphics[width=0.18\linewidth]{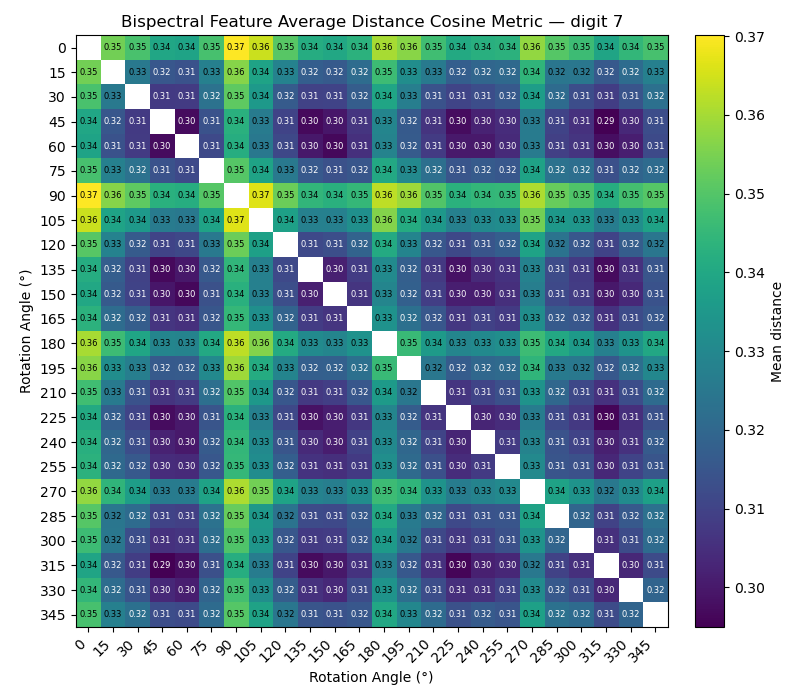} &
        \includegraphics[width=0.18\linewidth]{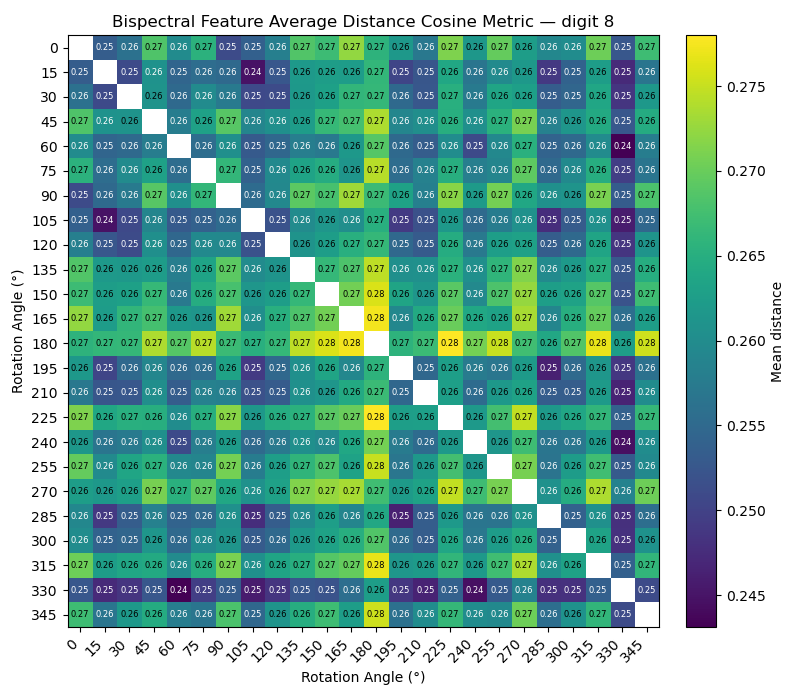} &
        \includegraphics[width=0.18\linewidth]{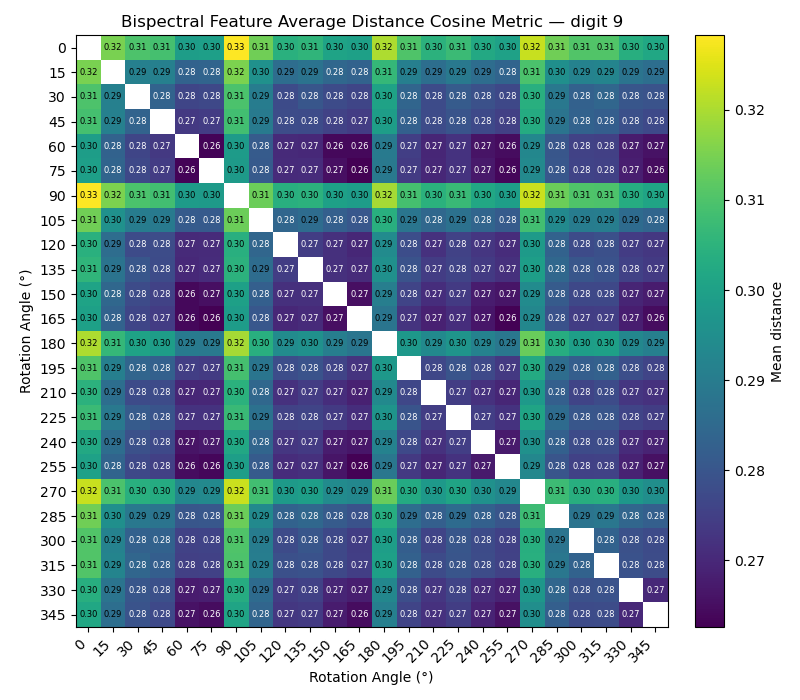}
      \end{tabular}
    }
  }
\end{figure}

\section{Benchmark Dataset Experiments}\label{apd:main-results}

\subsection{Experimental Setup}

As described in Appendix \ref{apd:rotmnist} for the preliminary experiments, we use $R = \frac{\min(M, N)}{2}$ radial bins and $K = 40$ angular bins in the discretized polar representation of our images of size $M \times N$. Thus, the bispectral representations of our features are $\mathbb{Z}/40\mathbb{Z}$ rotation invariant, and constructed using the group bispectrum of $\mathbb{Z}/40\mathbb{Z}.$ 

We perform OT between two disjoint halves of the \texttt{torchvision}'s\footnote{\url{https://docs.pytorch.org/vision/stable/index.html}} training split of each of \textsc{mnist}, \textsc{kmnist}, \textsc{fashion-mnist}, and the letters split of \textsc{emnist}, randomly splitting (with seed $0$) each class in each dataset's training set into two disjoint halves to construct the two datasets that we perform OT between. For \textsc{kmnist} and \textsc{fashion-mnist}, the training split is perfectly balanced between classes, so we obtain two datasets of size $30{,}000,$ with exactly $3{,}000$ images per class. The \textsc{mnist} training split is only approximately balanced, so splitting each class in half yields two datasets of size $29{,}997,$ with class sizes distributed as follows: $\{0:2961, 1:3371, 2:2979, 3:3065, 4:2921, 5:2710, 6:2959, 7:3132, 8:2925, 9:2974\}.$ For \textsc{emnist} (letters), which contains $26$ classes with a total of $124{,}800$ training examples, we similarly split each class in half, yielding two datasets of size $62{,}400$ with exactly $2{,}400$ images per class. 

For the main experiment, we perform OT from a set of images with rotations sampled uniformly at random from $0^{\circ}$ to $360^{\circ}$ to a disjoint unrotated set. As a baseline, also compute the class preservation statistics for the transport plan computed on two sets of unrotated raw and bispectral features. We calculate OT using \texttt{greenkhorn} greedy implementation of the entropically regularized version of OT in the Sinkhorn implementation of the \texttt{pot}\footnote{\url{pythonot.github.io/}} package \citep{flamary2021pot, altschuler2017near} for computational feasibility using a regularization of 
$0.01$ (which maximizes class preservation accuracy while ensuring convergene). Table \ref{tab:main_results} includes only the class preservation accuracies for raw OT using the $L_1$ pairwise cost between images (since this performs the best overall), so for completeness, we include the statistics for the other ground metrics in Table \ref{tab:class_pres_raw}. 

\begin{table}[htbp]
\floatconts
  {tab:class_pres_raw}
  {\caption{Class preservation accuracies for naive OT using $L_1, L_2, L_2^2$ and cosine pairwise metrics for cost.}}
  {{\resizebox{\linewidth}{!}{
    \begin{tabular}{|l||c|c|c|c||c|c|c|c|}
    \hline
    & \multicolumn{4}{|c||}{\bfseries Unrotated to Rotated} & \multicolumn{4}{c|}{\bfseries Baseline} \\
    \cline{2-9}
    \bfseries Dataset & \bfseries $L_1$ &  \bfseries $L_2$ & \bfseries $L_2^2$ & \bfseries $\cos$ &  \bfseries $L_1$ &  \bfseries $L_2$ & \bfseries $L_2^2$ & \bfseries $\cos$ \\
    \hline
    \textsc{mnist} 
  & 0.3297 & 0.3322 & 0.3289 & 0.3239 & 0.9725
 & 0.9697 & 0.9742 & 0.9705\\
    \hline
    \textsc{kmnist} & 0.2420 & 0.2408 & 0.2348 & 0.2412 & 0.9724 & 0.9698 & 0.9712 & 0.9666\\
    \hline
    \textsc{fmnist} & 0.3003 & 0.2787 & 0.2680 & 0.2615  & 0.8726 & 0.8748 & 0.8466 & 0.8736 \\
    \hline
    \textsc{emnist}  & 0.1969 & 0.1991 & 0.1991 & 0.1966 &  0.8754 & 0.8678 & 0.8678 & 0.8730 \\
    \hline
    \end{tabular}}}
  }
\end{table}

\subsection{Per-Class OT Matching Statistics}

For a more granular understanding of the how BOT preserves semantic label structure in each dataset, we include confusion matrices that separate the class preservation accuracies for raw feature OT and Bispectral OT in Table \ref{tab:main_results} by class, detailing the fraction of each class matched to elements in each other class for each of our benchmark datasets. 

\subsubsection{MNIST}

We include the confusion matrices for OT on \textsc{mnist} for the OT matching between rotated and unrotated images in Figures \ref{fig:mnist-bispectral-real} (bispectral features) and \ref{fig:mnist-raw-real} (raw pixel features), and for the baseline experiment (matching unrotated images) in Figures \ref{fig:mnist-bispectral-baseline} (bispectral features) and \ref{fig:mnist-raw-baseline} (raw pixel features)

\begin{figure}[htbp]
\floatconts
  {fig:mnist-bispectral-real}
  {\caption{Class matching statistics for OT plan from rotated \textsc{mnist} to unrotated \textsc{mnist} on bispectral features.}}
  {%
    \subfigure[$L_1$ Cost]{
      \includegraphics[width=0.45\linewidth]{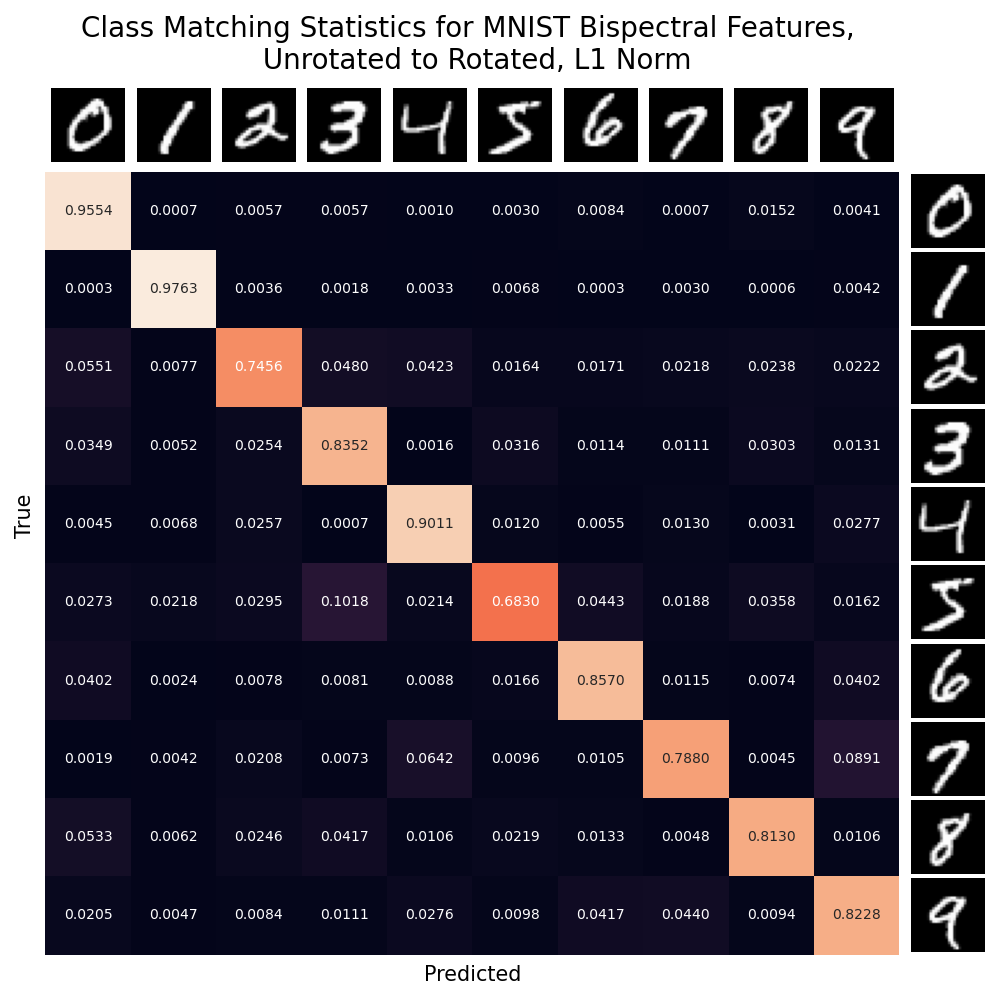}}%
    \quad
    \subfigure[$L_2$ Cost]{
      \includegraphics[width=0.45\linewidth]{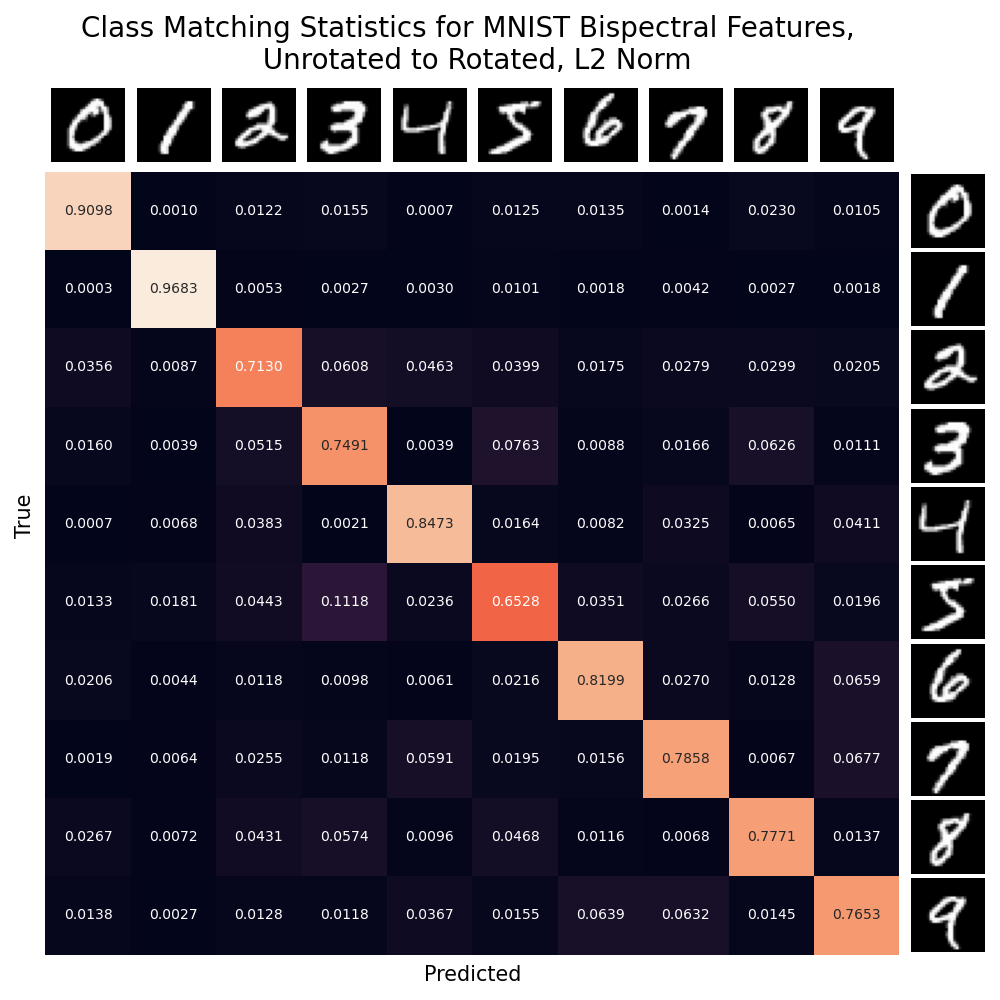}}
      \subfigure[$L_2^2$ Cost]{
      \includegraphics[width=0.45\linewidth]{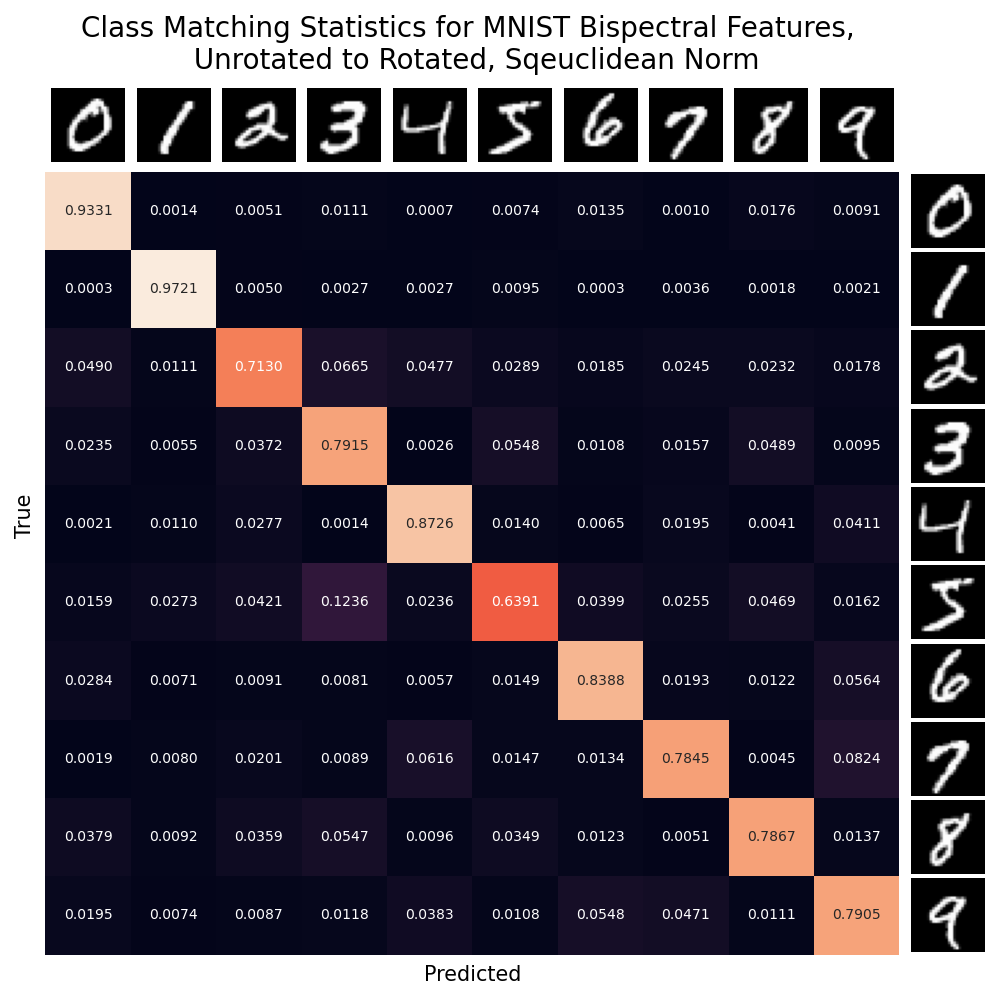}}%
    \quad
    \subfigure[$\cos$ Cost]{
      \includegraphics[width=0.45\linewidth]{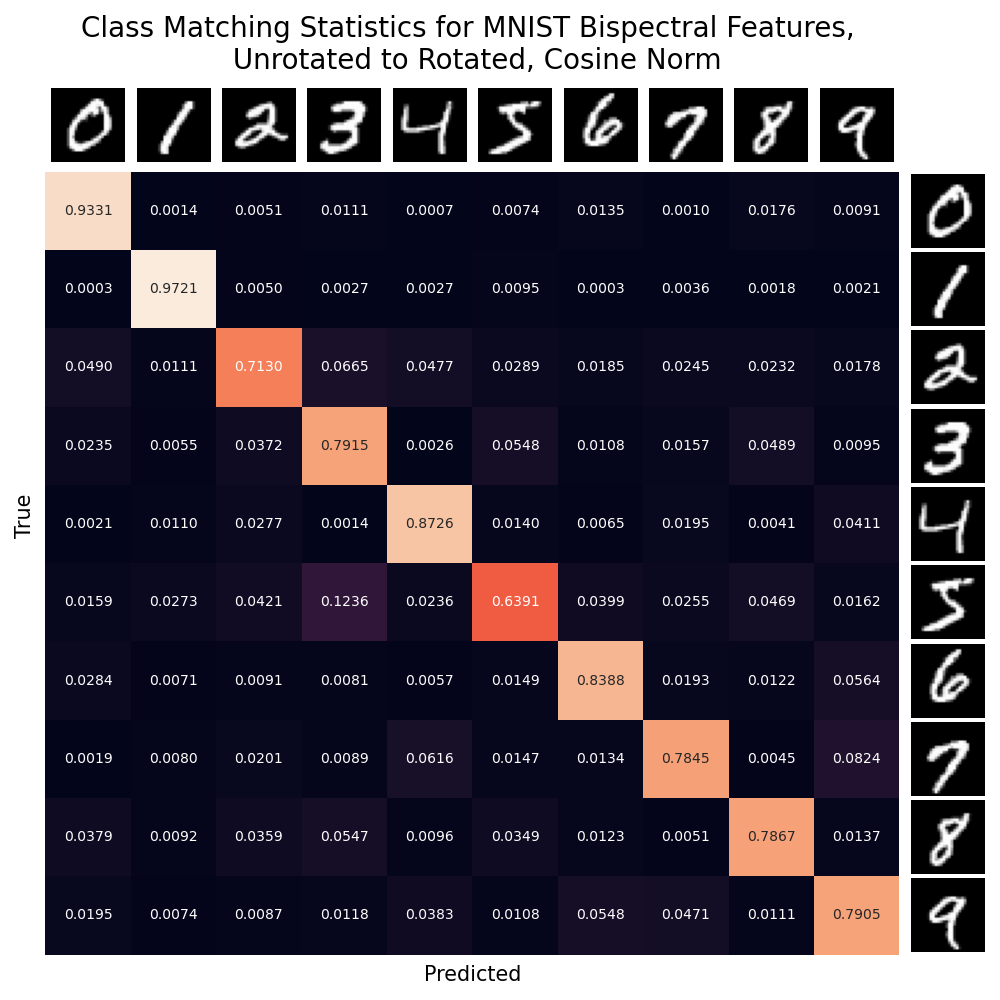}}
    }
\end{figure}

\begin{figure}[htbp]
\floatconts
  {fig:mnist-raw-real}
  {\caption{Class matching statistics for OT plan from rotated \textsc{mnist} to unrotated \textsc{mnist} on raw pixel features.}}
  {%
    \subfigure[$L_1$ Cost]{
      \includegraphics[width=0.45\linewidth]{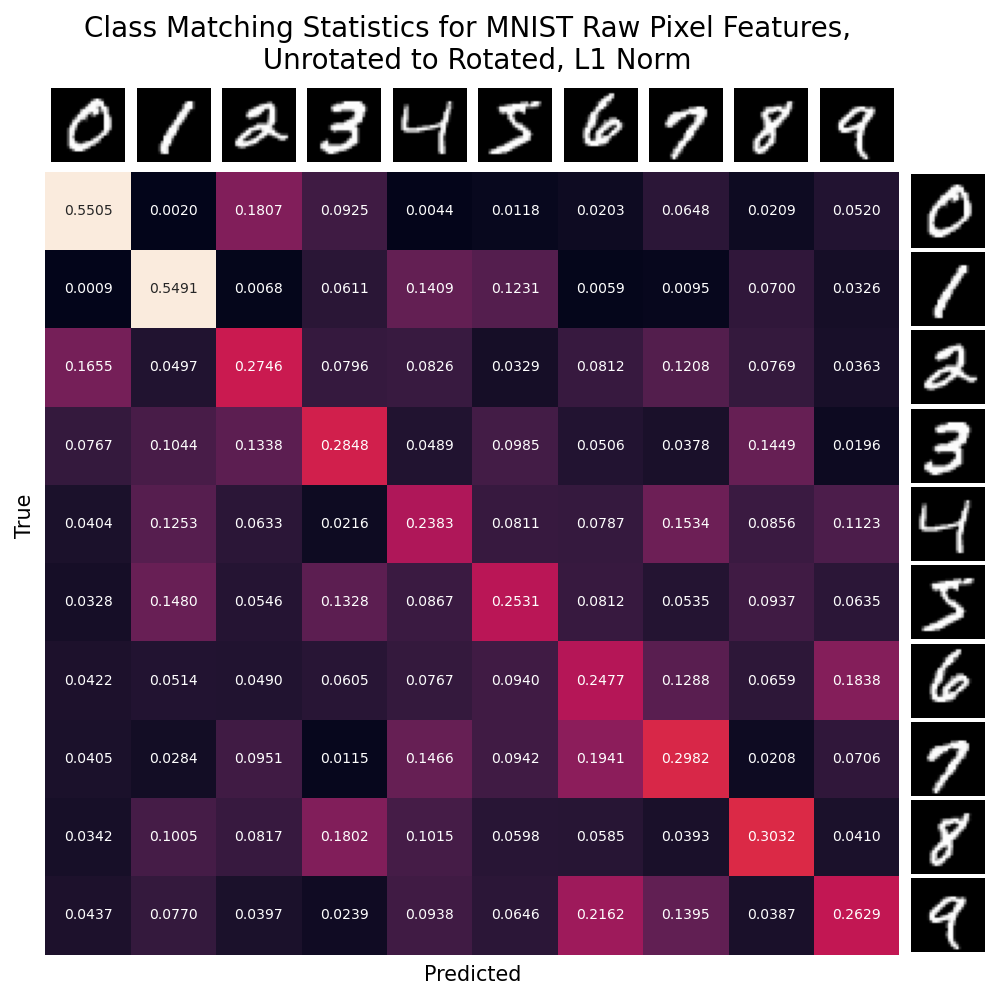}}%
    \quad
    \subfigure[$L_2$ Cost]{
      \includegraphics[width=0.45\linewidth]{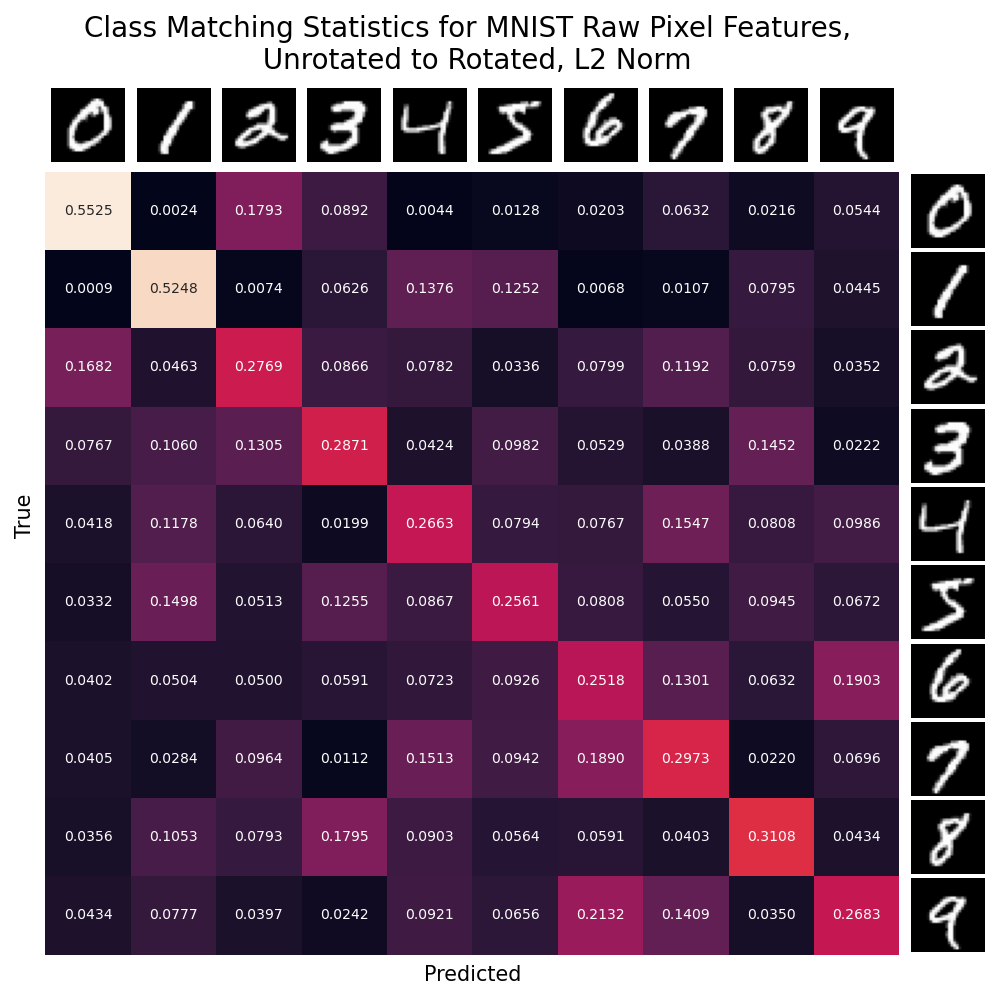}}
      \subfigure[$L_2^2$ Cost]{
      \includegraphics[width=0.45\linewidth]{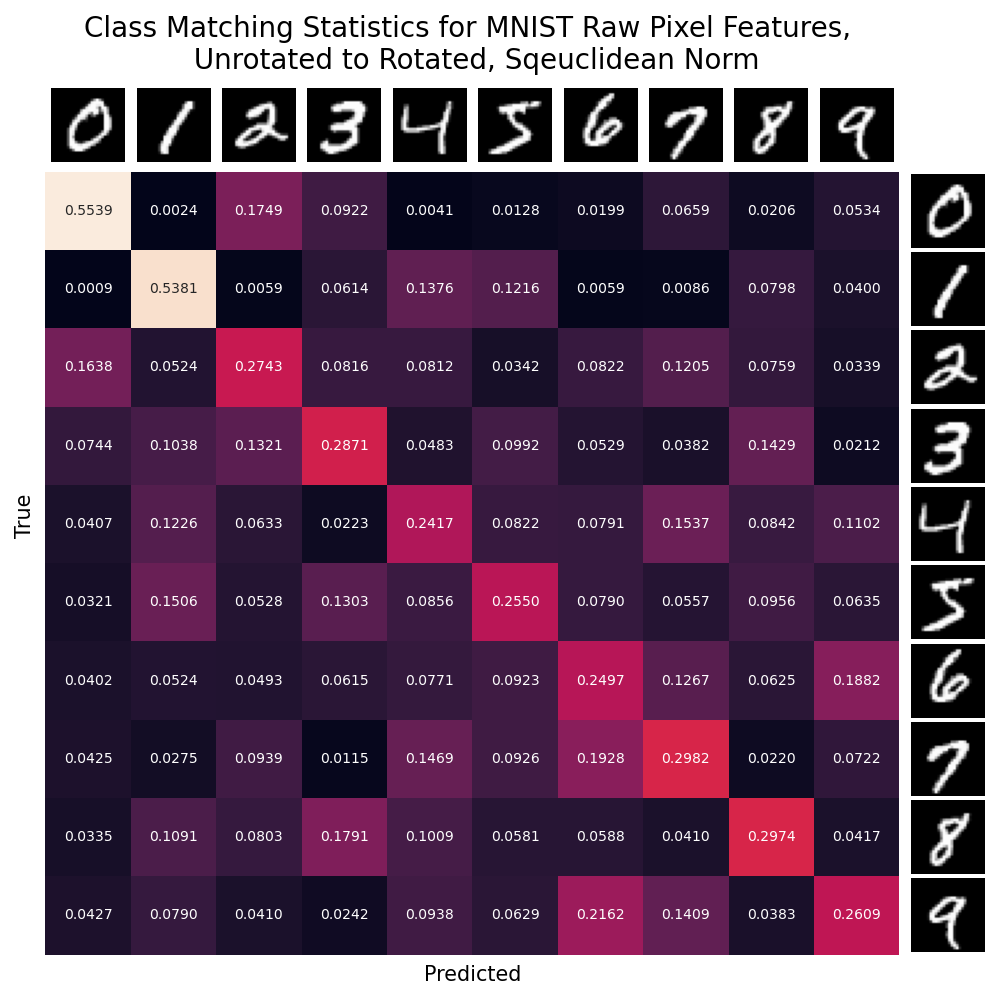}}%
    \quad
    \subfigure[$\cos$ Cost]{
      \includegraphics[width=0.45\linewidth]{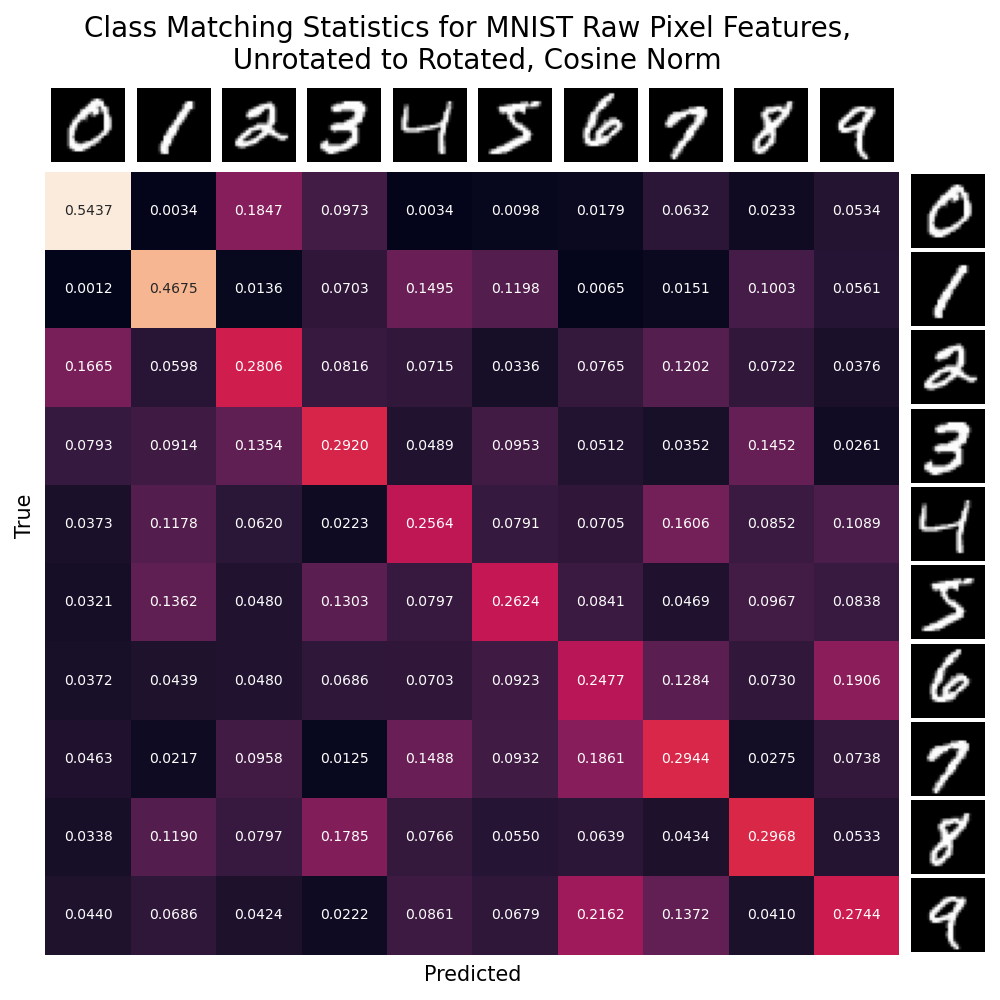}}
      }
\end{figure}

\begin{figure}[htbp]
\floatconts
  {fig:mnist-bispectral-baseline}
  {\caption{Class matching statistics for OT plan from unrotated \textsc{mnist} to unrotated \textsc{mnist} (baseline) on bispectral features.}}
  {%
    \subfigure[$L_1$ Cost]{
      \includegraphics[width=0.45\linewidth]{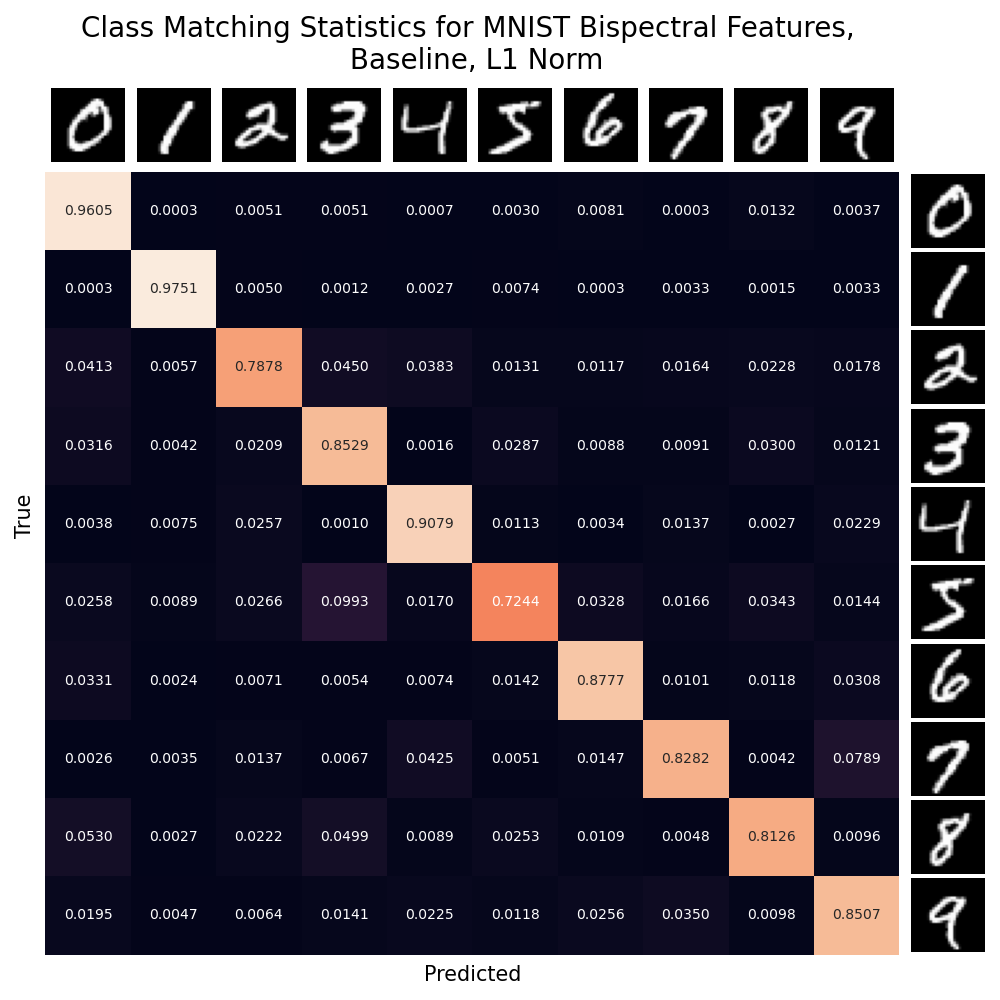}}%
    \quad
    \subfigure[$L_2$ Cost]{
      \includegraphics[width=0.45\linewidth]{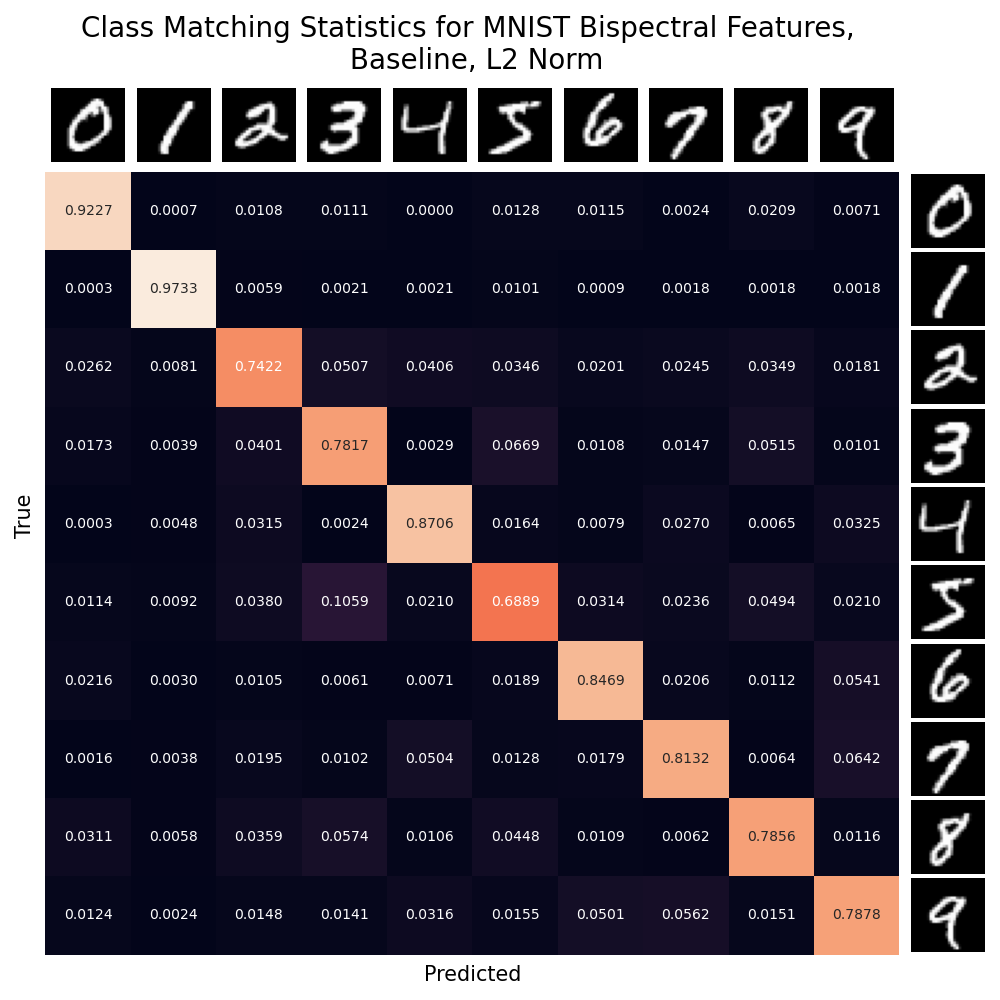}}
      \subfigure[$L_2^2$ Cost]{
      \includegraphics[width=0.45\linewidth]{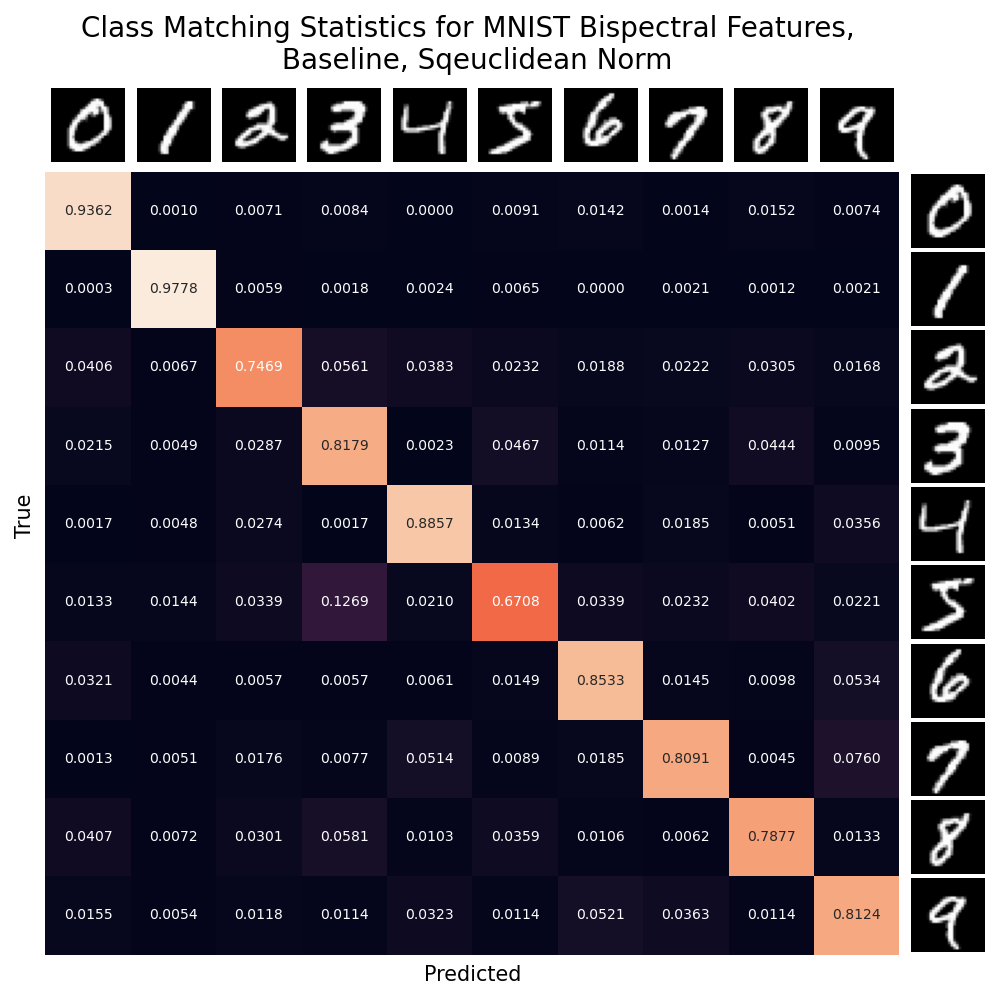}}%
    \quad
    \subfigure[$\cos$ Cost]{
      \includegraphics[width=0.45\linewidth]{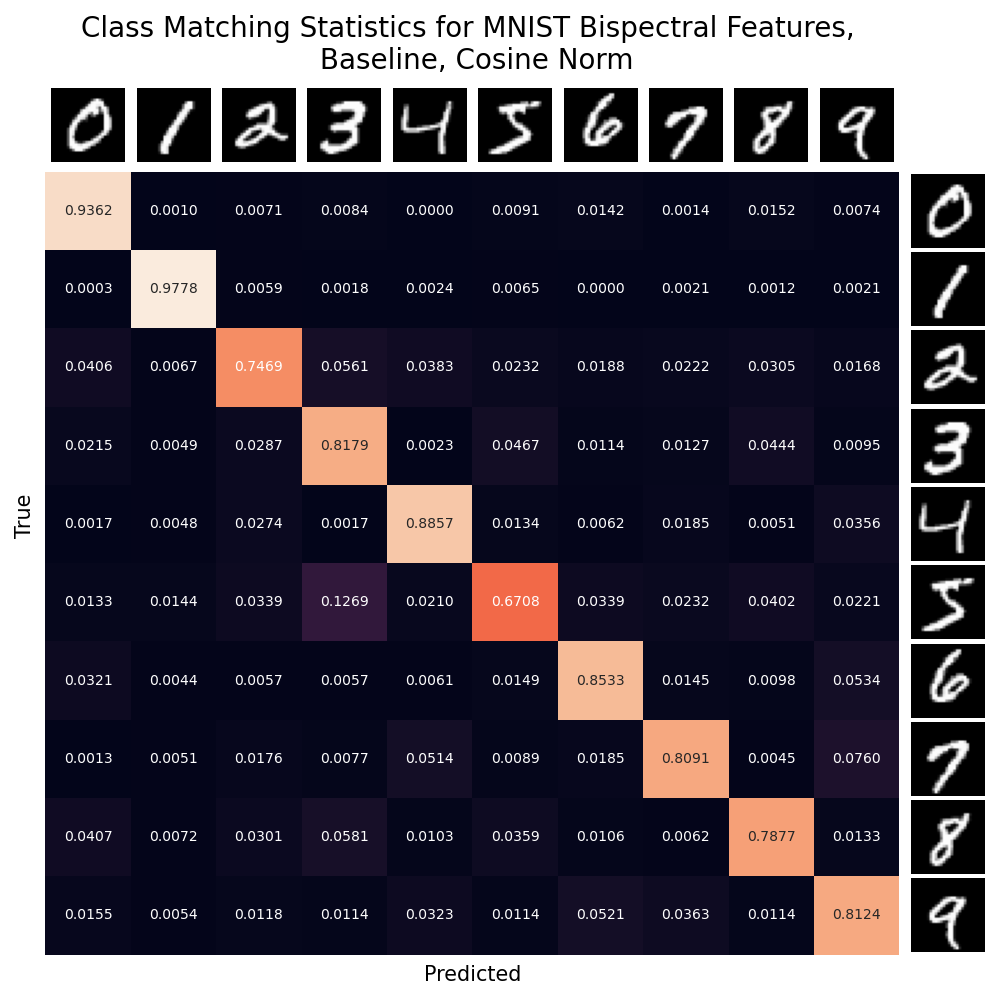}}
    }
\end{figure}

\begin{figure}[htbp]
\floatconts
  {fig:mnist-raw-baseline}
  {\caption{Class matching statistics for OT plan from unrotated \textsc{mnist} to unrotated \textsc{mnist} (baseline) on raw pixel features.}}
  {%
    \subfigure[$L_1$ Cost]{
      \includegraphics[width=0.45\linewidth]{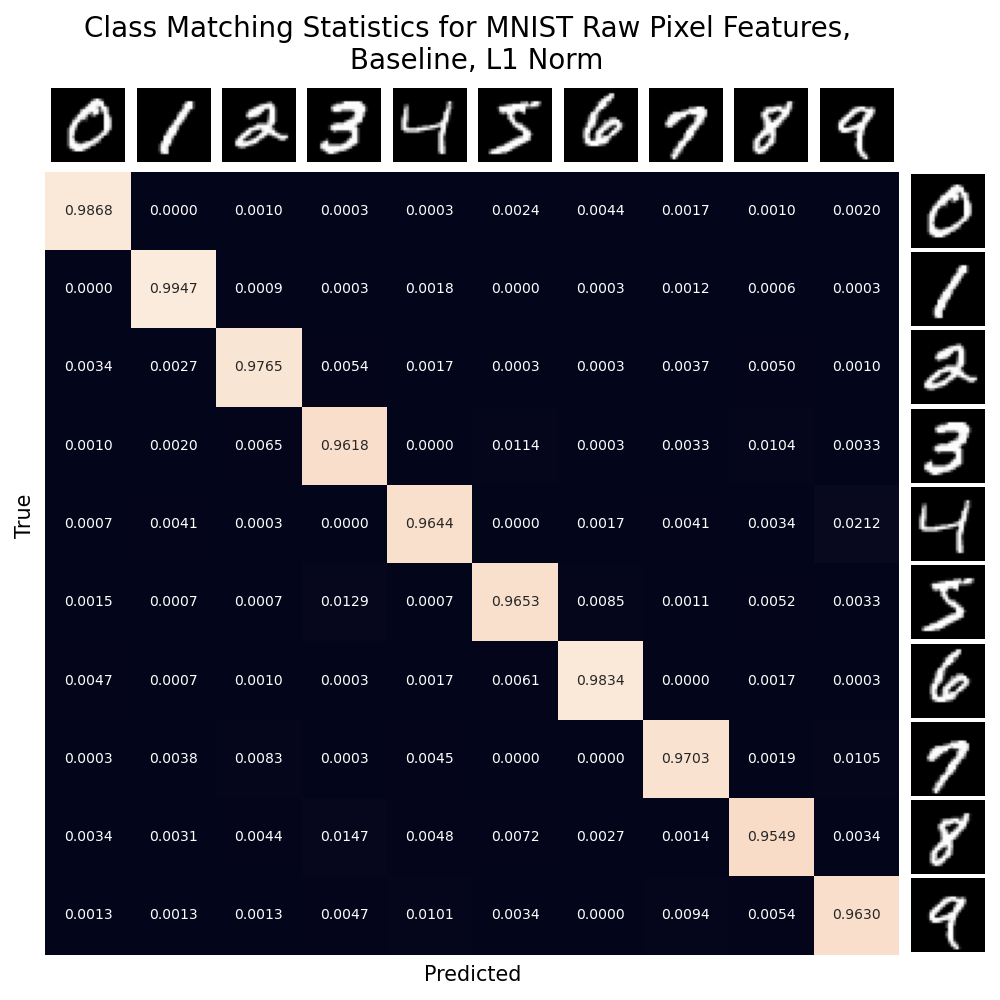}}%
    \quad
    \subfigure[$L_2$ Cost]{
      \includegraphics[width=0.45\linewidth]{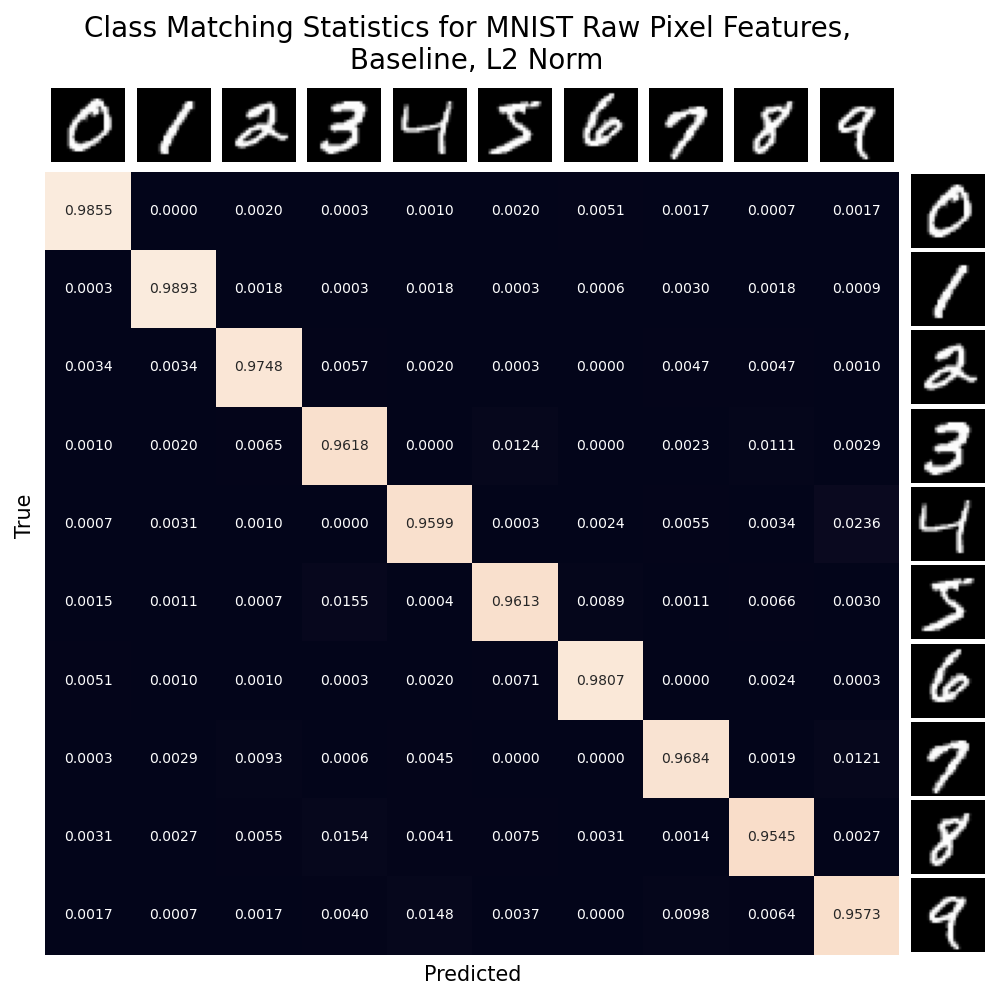}}
      \subfigure[$L_2^2$ Cost]{
      \includegraphics[width=0.45\linewidth]{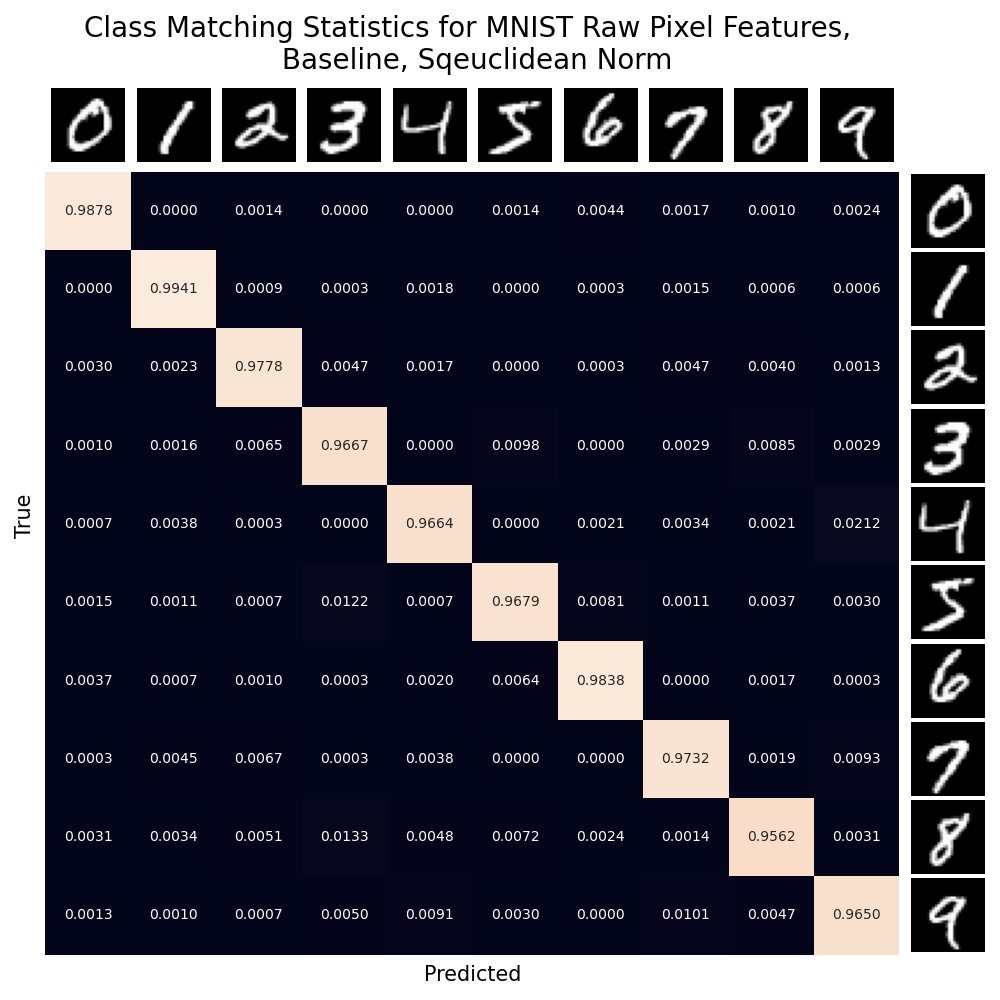}}%
    \quad
    \subfigure[$\cos$ Cost]{
      \includegraphics[width=0.45\linewidth]{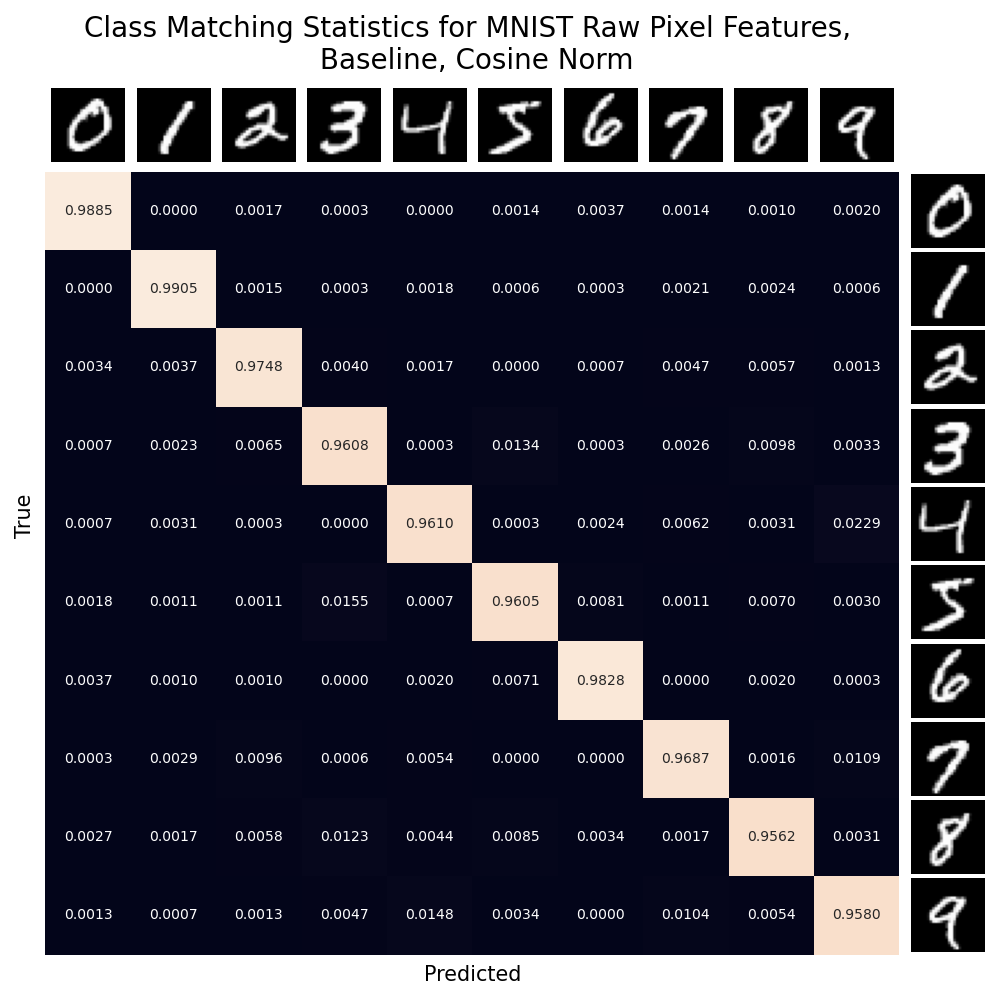}}
    }
\end{figure}

\subsubsection{KMNIST}

We include the confusion matrices for OT on \textsc{kmnist} for the OT matching between rotated and unrotated images in Figures \ref{fig:kmnist-bispectral-real} (bispectral features) and \ref{fig:kmnist-raw-real} (raw pixel features), and for the baseline experiment (matching unrotated images) in Figures \ref{fig:kmnist-bispectral-baseline} (bispectral features) and \ref{fig:kmnist-raw-baseline} (raw pixel features). 

\begin{figure}[htbp]
\floatconts
  {fig:kmnist-bispectral-real}
  {\caption{Class matching statistics for OT plan from rotated \textsc{kmnist} to unrotated \textsc{kmnist} on bispectral features.}}
  {%
    \subfigure[$L_1$ Cost]{
      \includegraphics[width=0.45\linewidth]{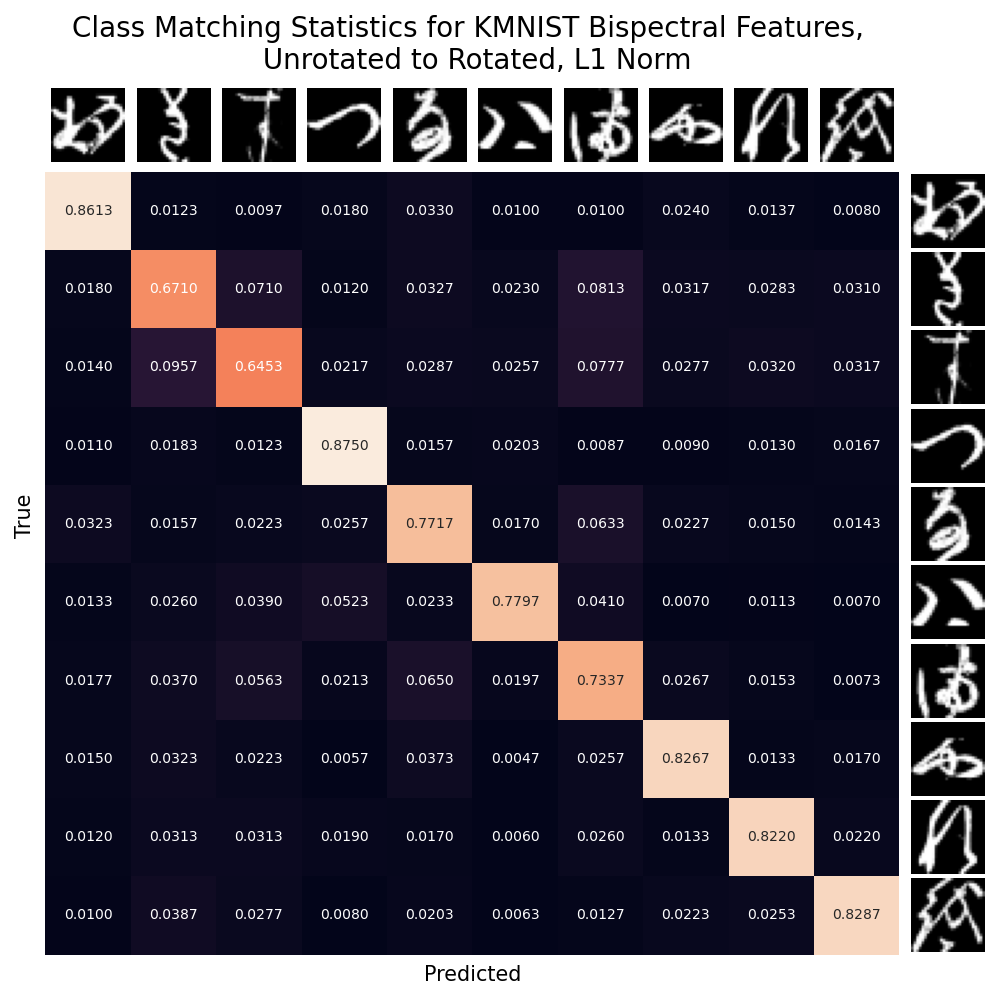}}%
    \quad
    \subfigure[$L_2$ Cost]{
      \includegraphics[width=0.45\linewidth]{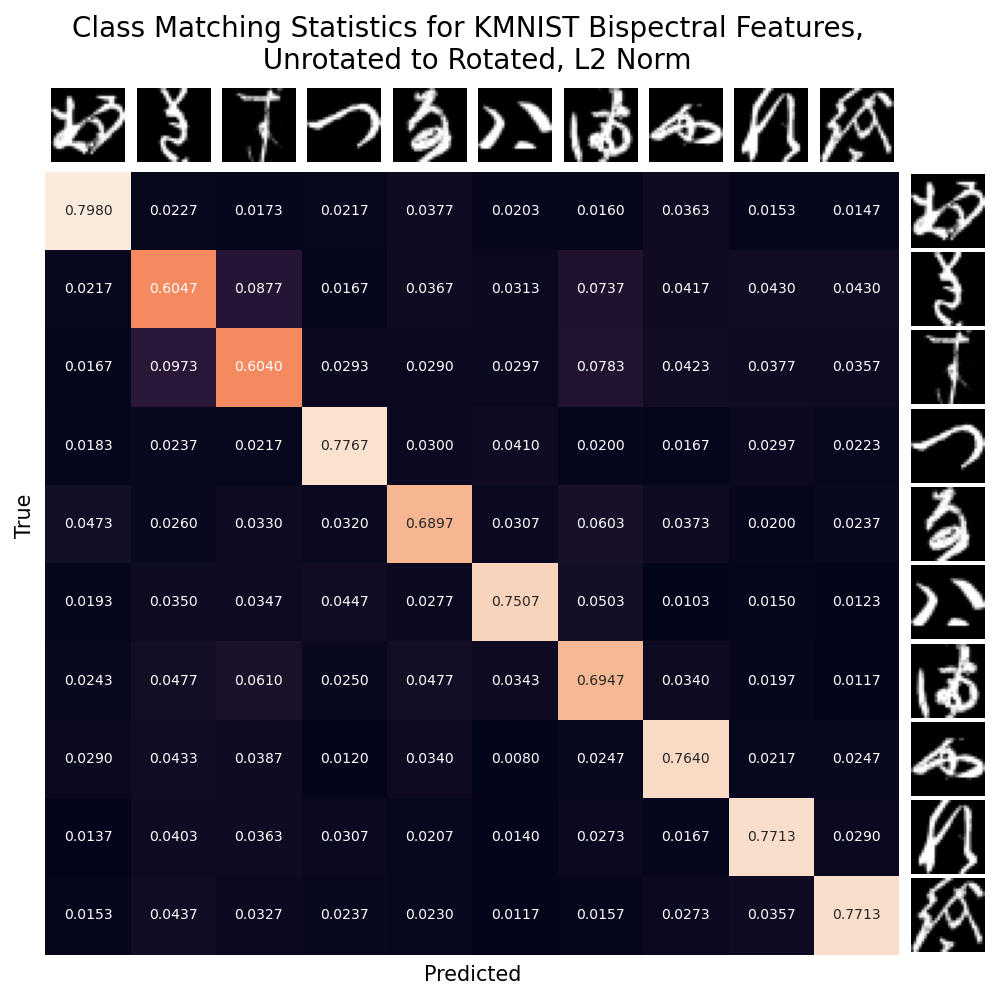}}
      \subfigure[$L_2^2$ Cost]{
      \includegraphics[width=0.45\linewidth]{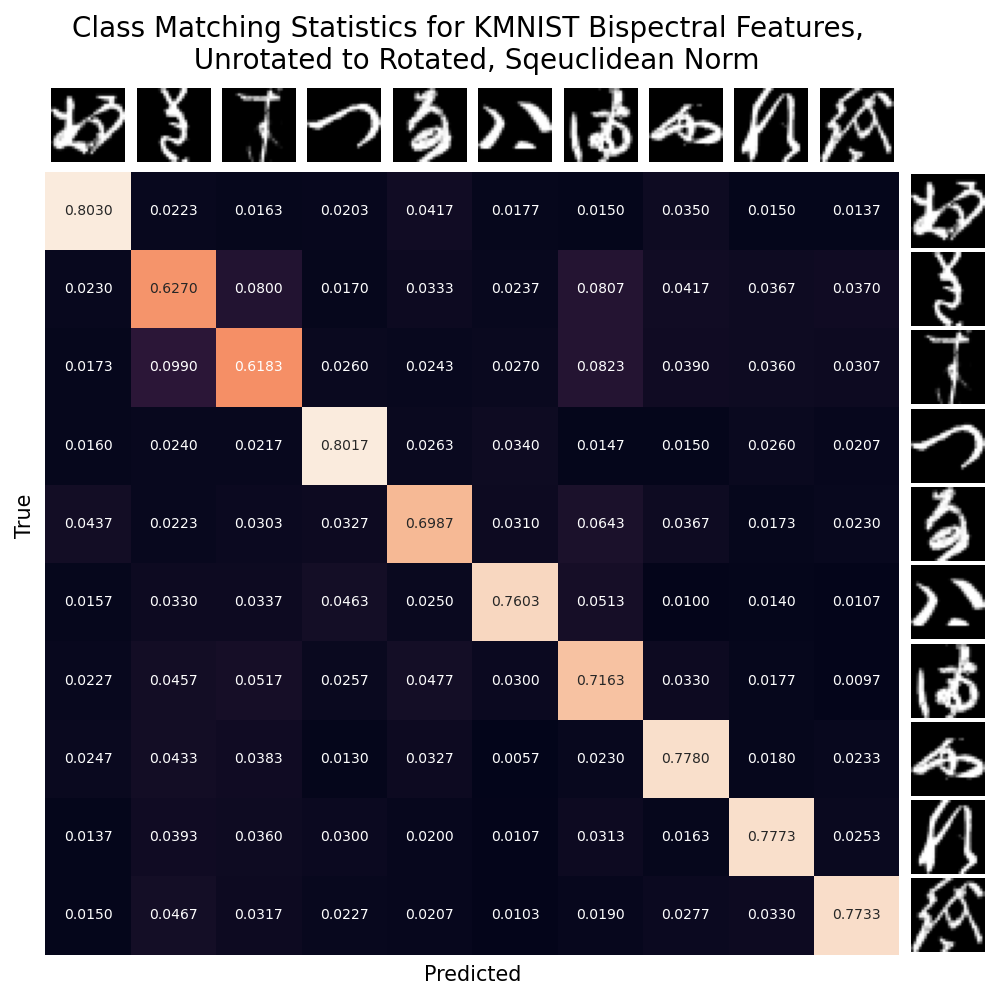}}%
    \quad
    \subfigure[$\cos$ Cost]{
      \includegraphics[width=0.45\linewidth]{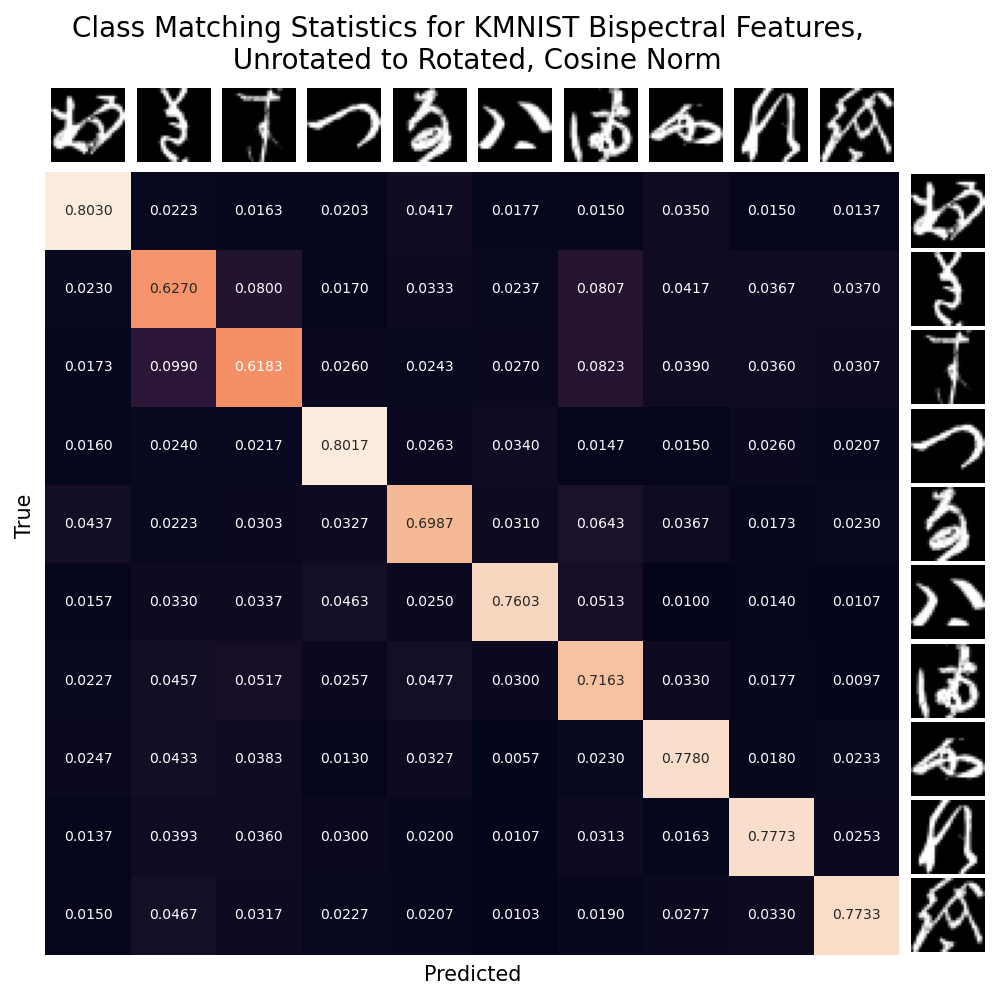}}
    }
\end{figure}

\begin{figure}[htbp]
\floatconts
  {fig:kmnist-raw-real}
  {\caption{Class matching statistics for OT plan from rotated \textsc{kmnist} to unrotated \textsc{kmnist} on raw pixel features.}}
  {%
    \subfigure[$L_1$ Cost]{
      \includegraphics[width=0.45\linewidth]{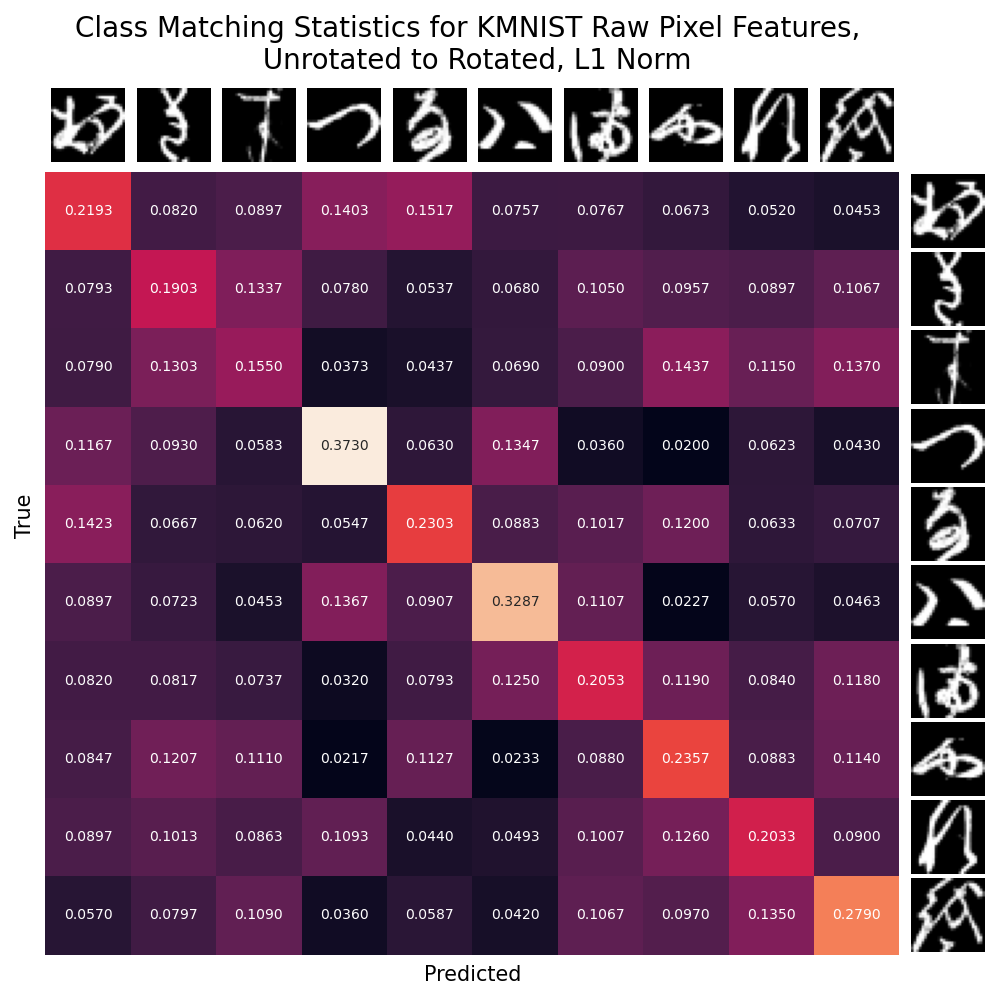}}%
    \quad
    \subfigure[$L_2$ Cost]{
      \includegraphics[width=0.45\linewidth]{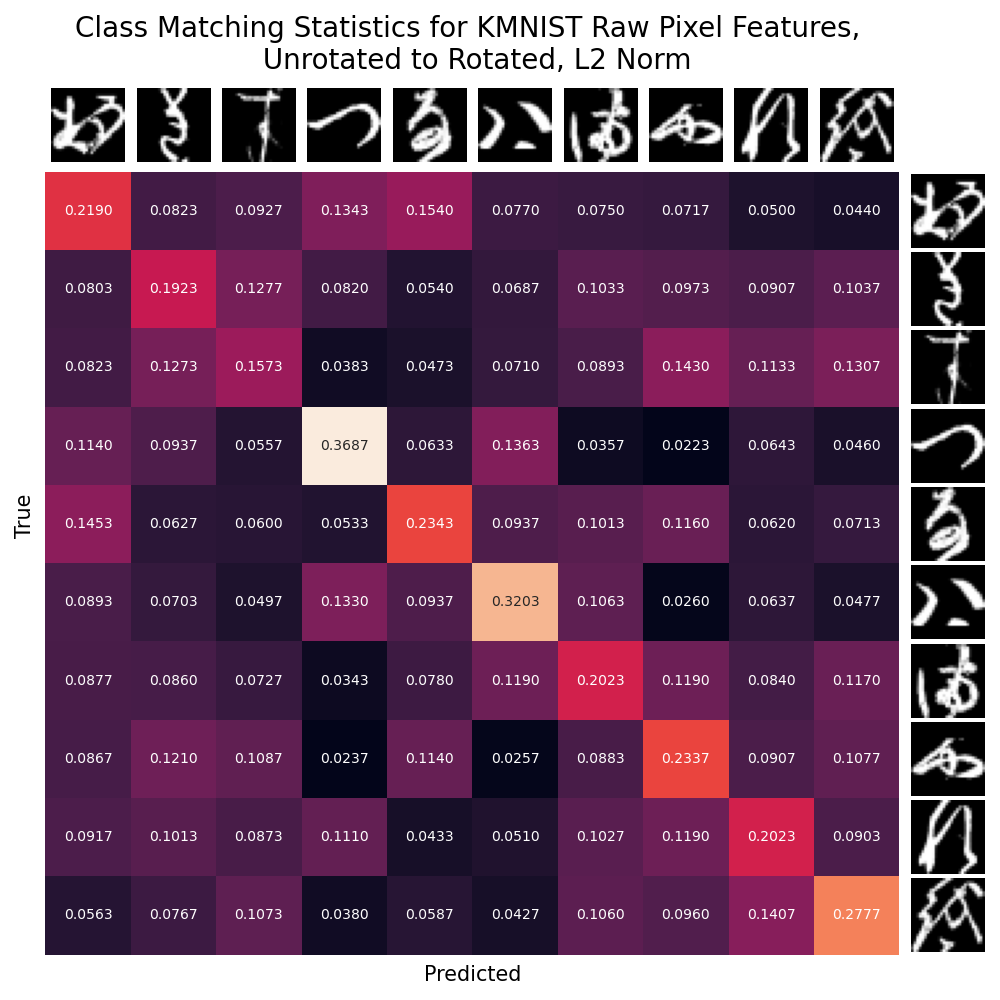}}
      \subfigure[$L_2^2$ Cost]{
      \includegraphics[width=0.45\linewidth]{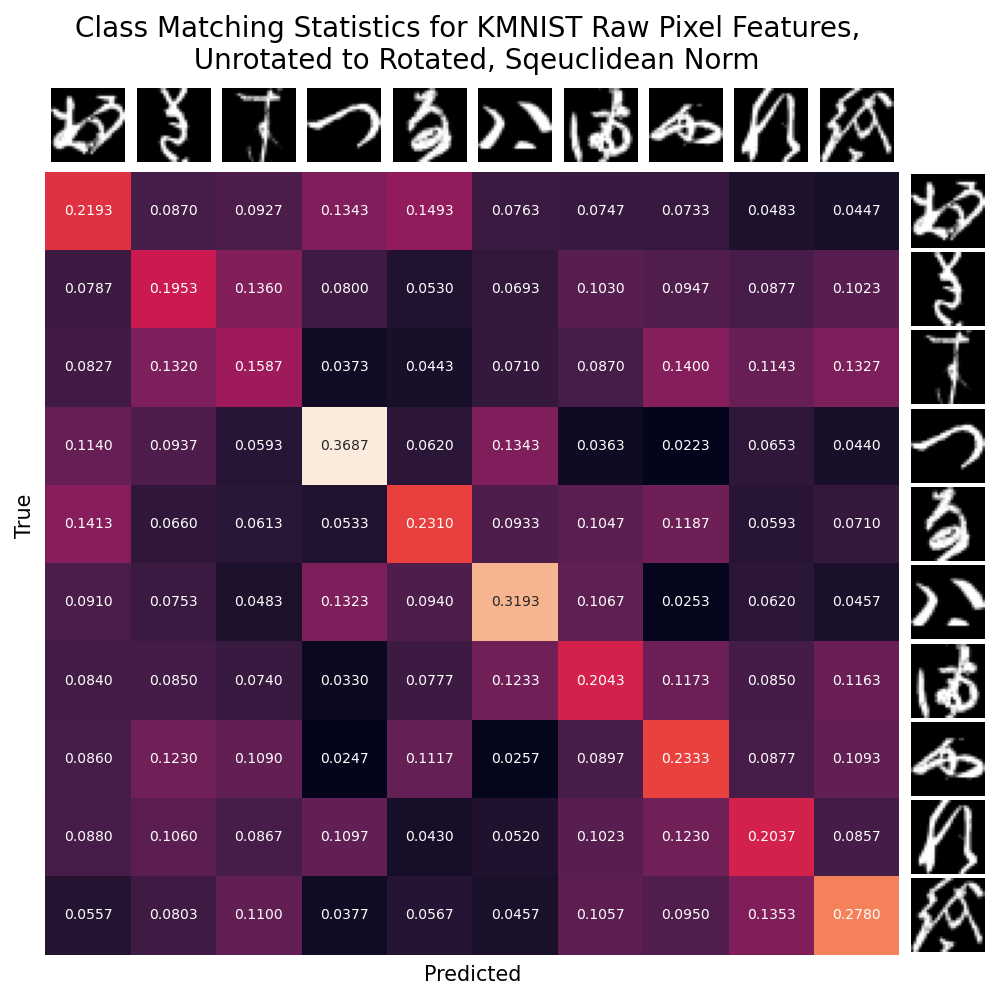}}%
    \quad
    \subfigure[$\cos$ Cost]{
      \includegraphics[width=0.45\linewidth]{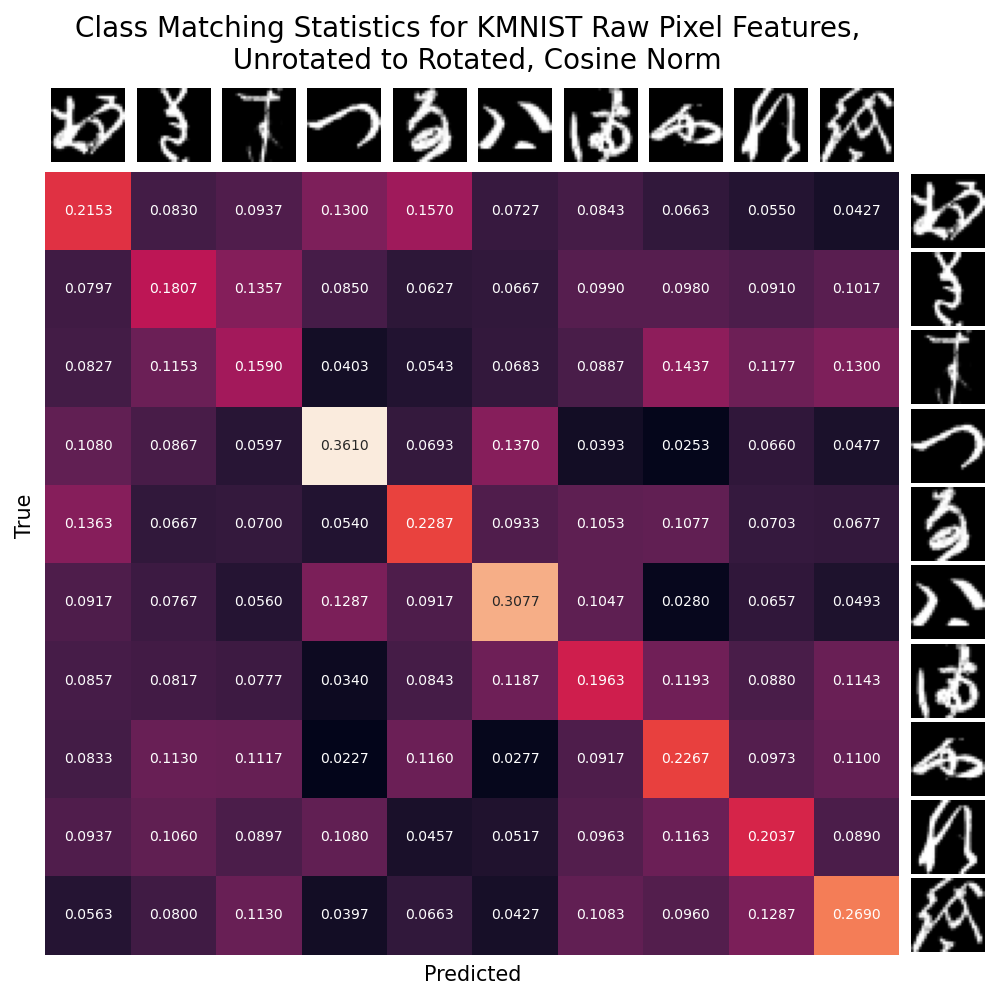}}
      }
\end{figure}

\begin{figure}[htbp]
\floatconts
  {fig:kmnist-bispectral-baseline}
  {\caption{Class matching statistics for OT plan from unrotated \textsc{kmnist} to unrotated \textsc{kmnist} (baseline) on bispectral features.}}
  {%
    \subfigure[$L_1$ Cost]{
      \includegraphics[width=0.45\linewidth]{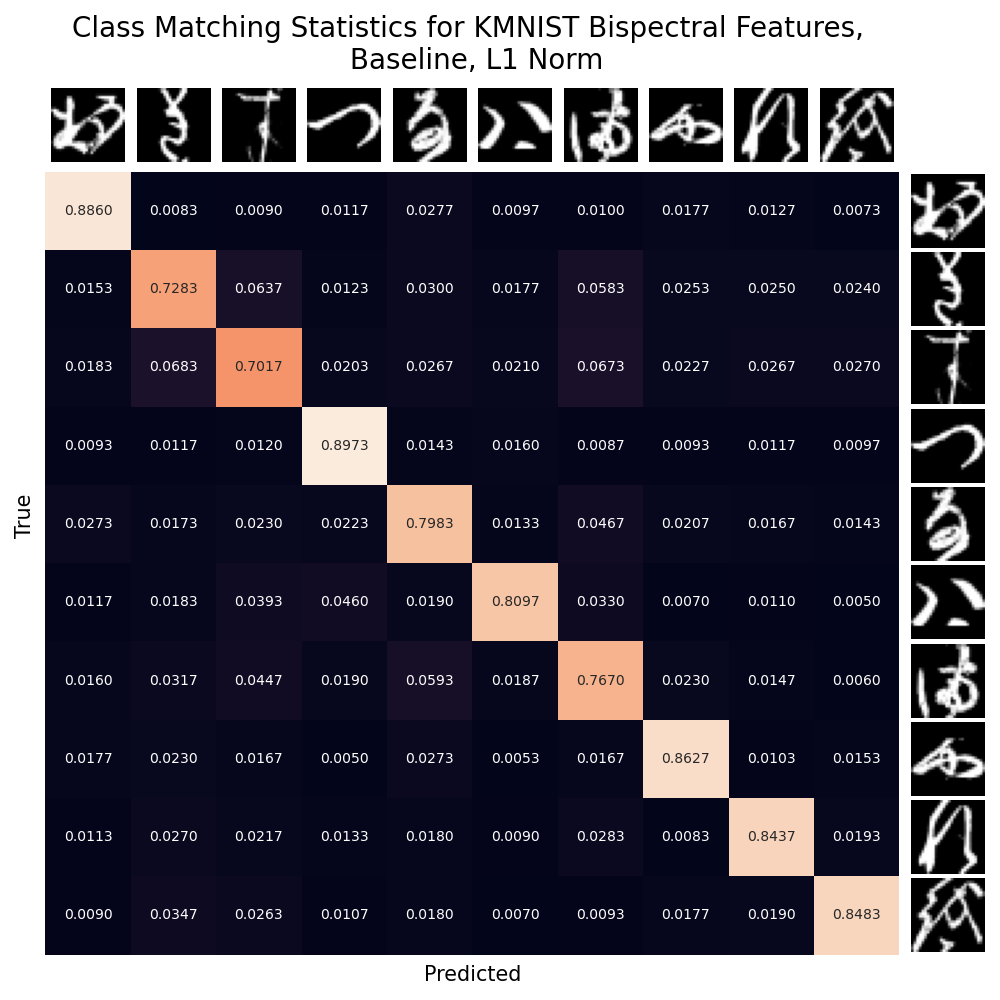}}%
    \quad
    \subfigure[$L_2$ Cost]{
      \includegraphics[width=0.45\linewidth]{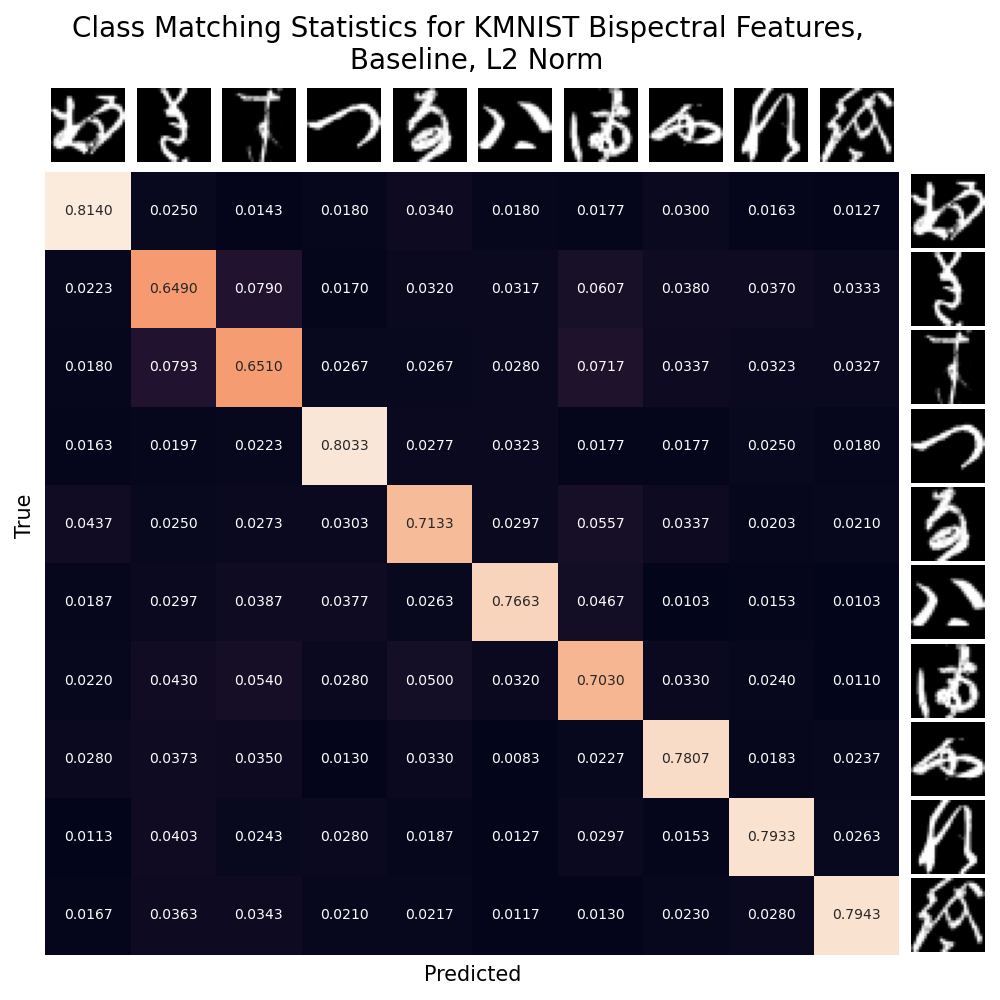}}
      \subfigure[$L_2^2$ Cost]{
      \includegraphics[width=0.45\linewidth]{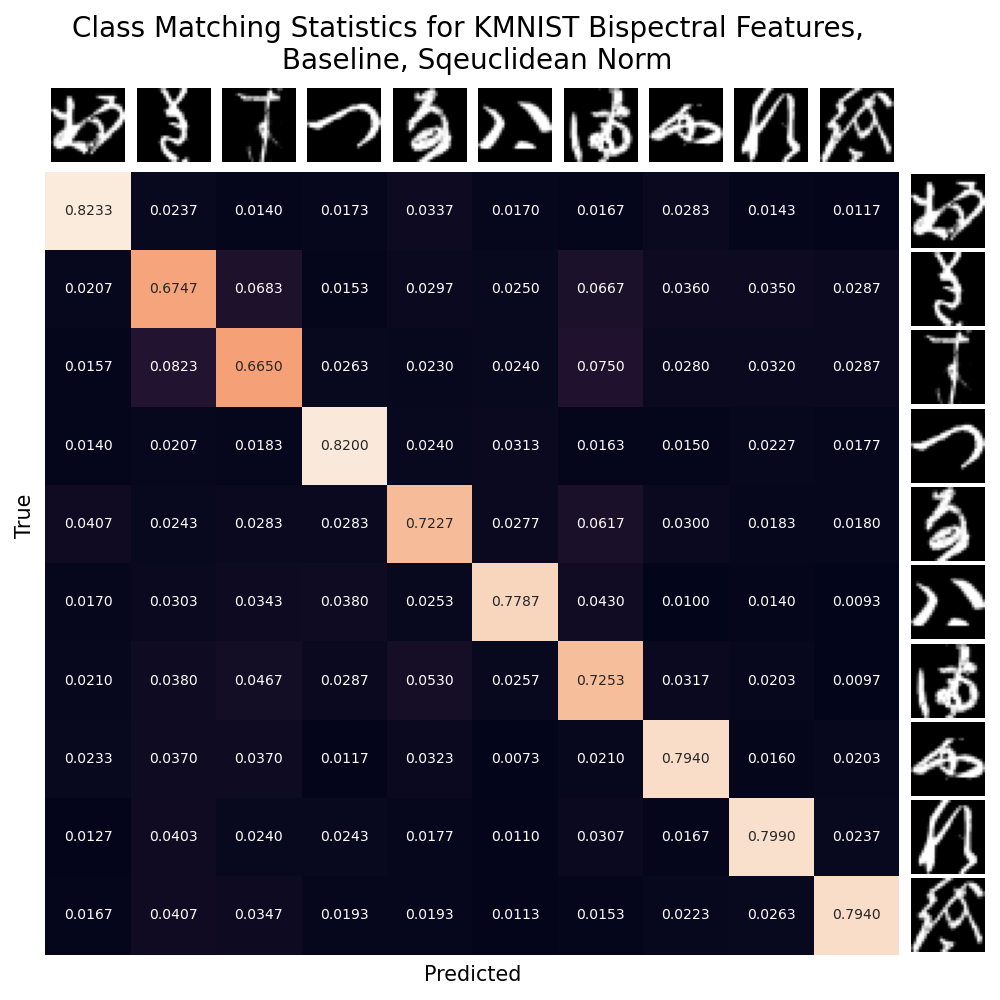}}%
    \quad
    \subfigure[$\cos$ Cost]{
      \includegraphics[width=0.45\linewidth]{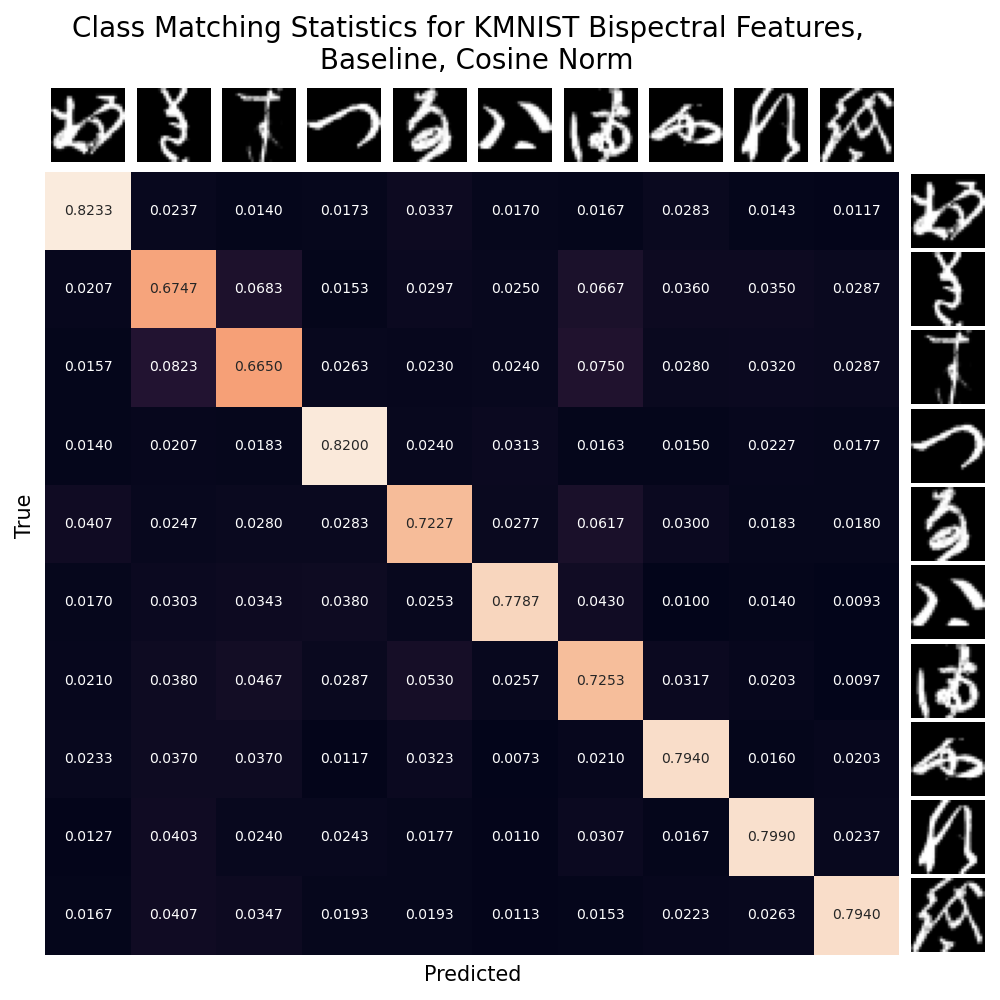}}
    }
\end{figure}

\begin{figure}[htbp]
\floatconts
  {fig:kmnist-raw-baseline}
  {\caption{Class matching statistics for OT plan from unrotated \textsc{kmnist} to unrotated \textsc{kmnist} (baseline) on raw pixel features.}}
  {%
    \subfigure[$L_1$ Cost]{
      \includegraphics[width=0.45\linewidth]{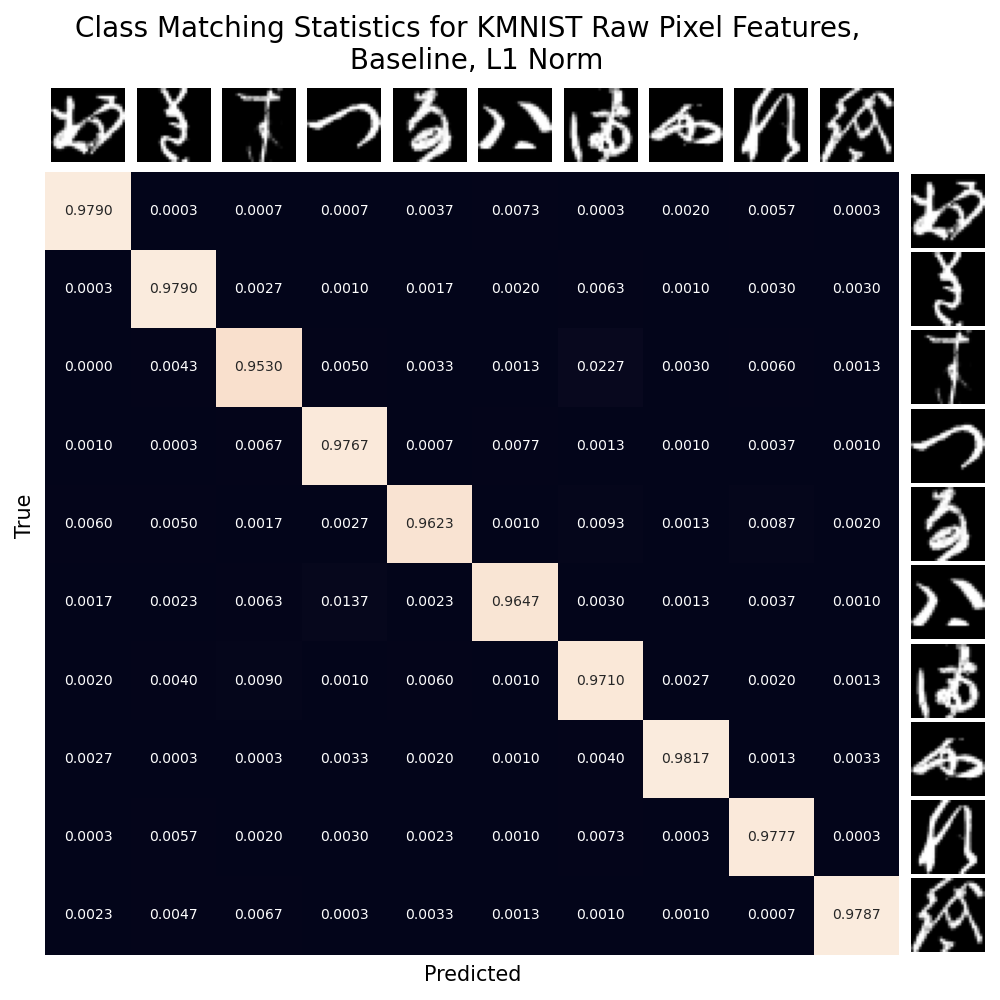}}%
    \quad
    \subfigure[$L_2$ Cost]{
      \includegraphics[width=0.45\linewidth]{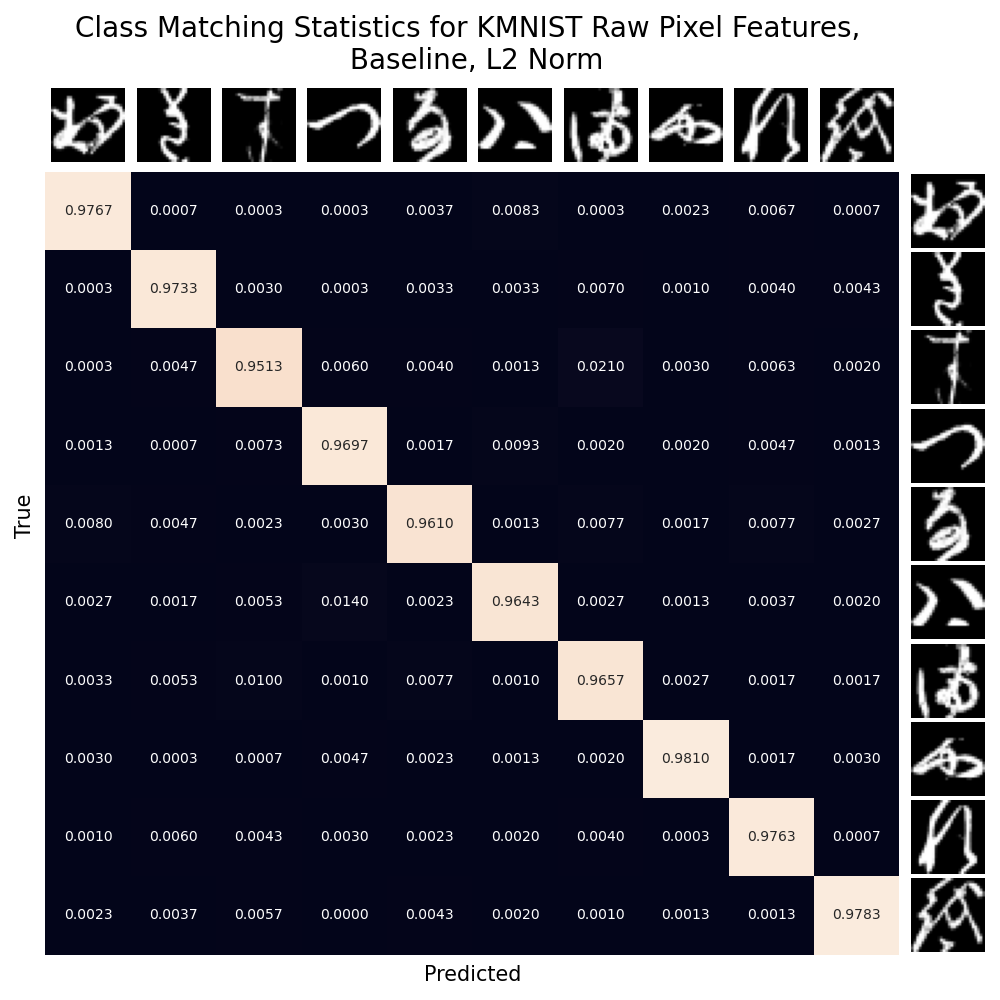}}
      \subfigure[$L_2^2$ Cost]{
      \includegraphics[width=0.45\linewidth]{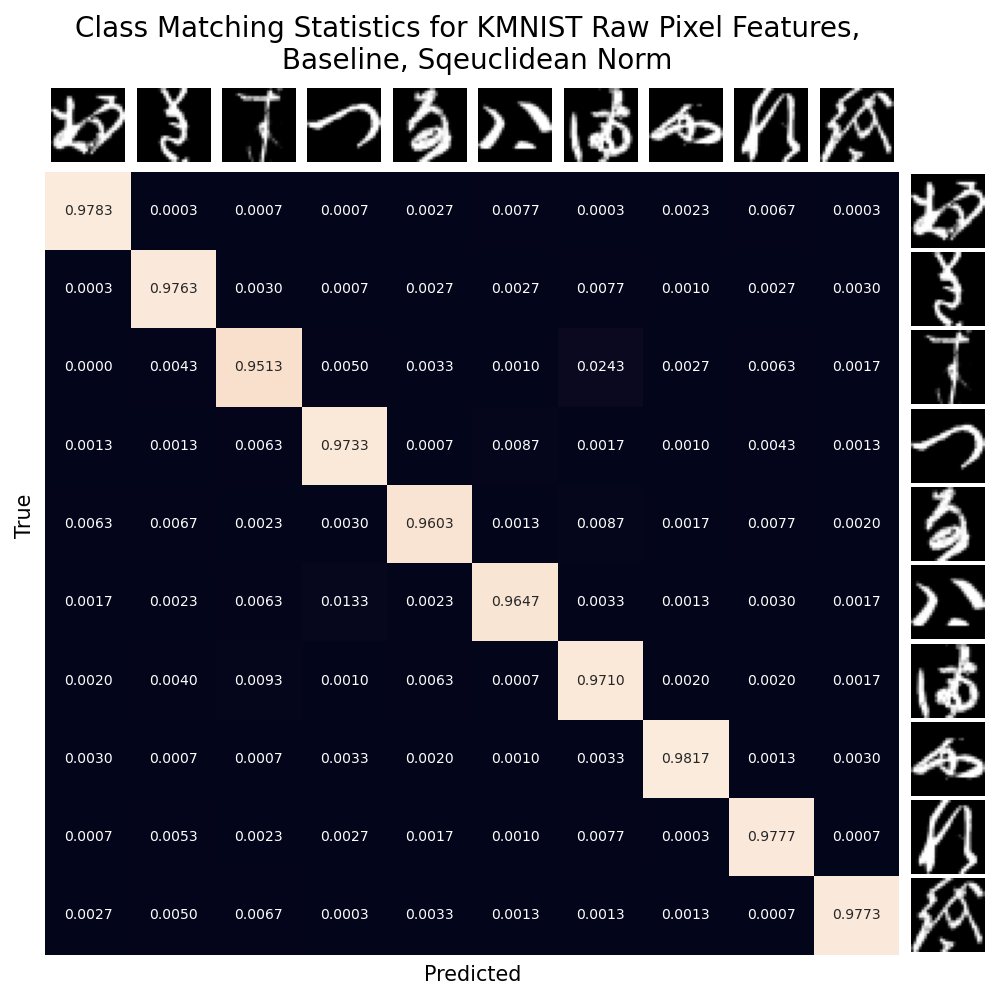}}%
    \quad
    \subfigure[$\cos$ Cost]{
      \includegraphics[width=0.45\linewidth]{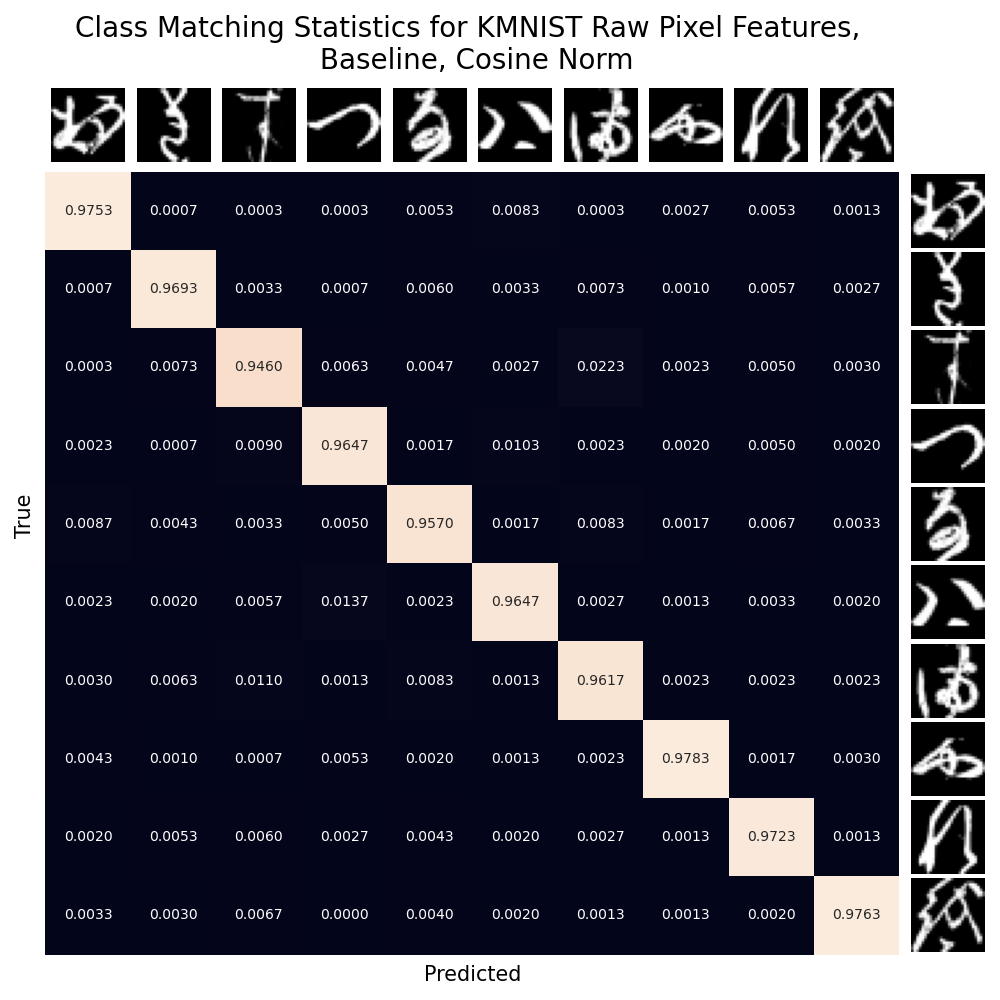}}
    }
\end{figure}

\subsubsection{Fashion-MNIST}

We include the confusion matrices for OT on \textsc{fashion-mnist} for the OT matching between rotated and unrotated images in Figures \ref{fig:fmnist-bispectral-real} (bispectral features) and \ref{fig:fmnist-raw-real} (raw pixel features), and for the baseline experiment (matching unrotated images) in Figures \ref{fig:fmnist-bispectral-baseline} (bispectral features) and \ref{fig:fmnist-raw-baseline} (raw pixel features).

\begin{figure}[htbp]
\floatconts
  {fig:fmnist-bispectral-real}
  {\caption{Class matching statistics for OT plan from rotated \textsc{fashion-mnist} to unrotated \textsc{fashion-mnist} on bispectral features.}}
  {%
    \subfigure[$L_1$ Cost]{
      \includegraphics[width=0.45\linewidth]{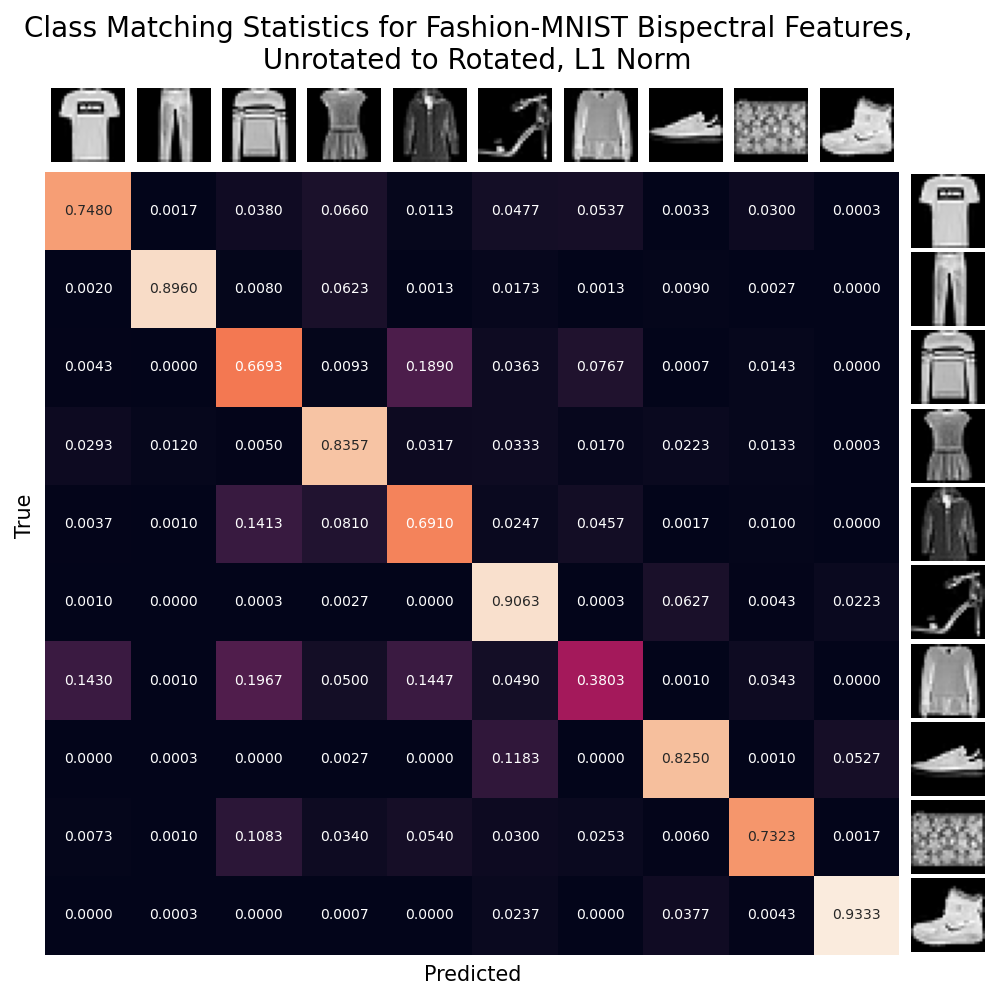}}%
    \quad
    \subfigure[$L_2$ Cost]{
      \includegraphics[width=0.45\linewidth]{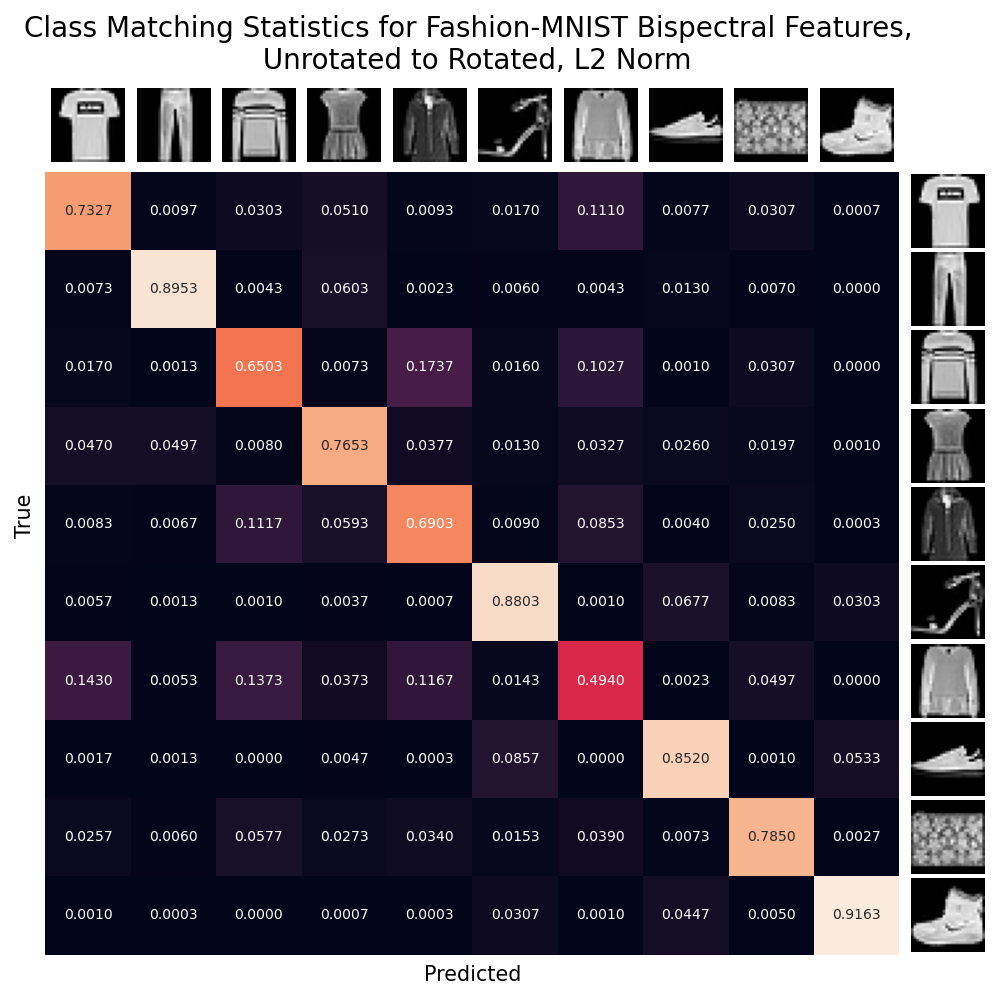}}
      \subfigure[$L_2^2$ Cost]{
      \includegraphics[width=0.45\linewidth]{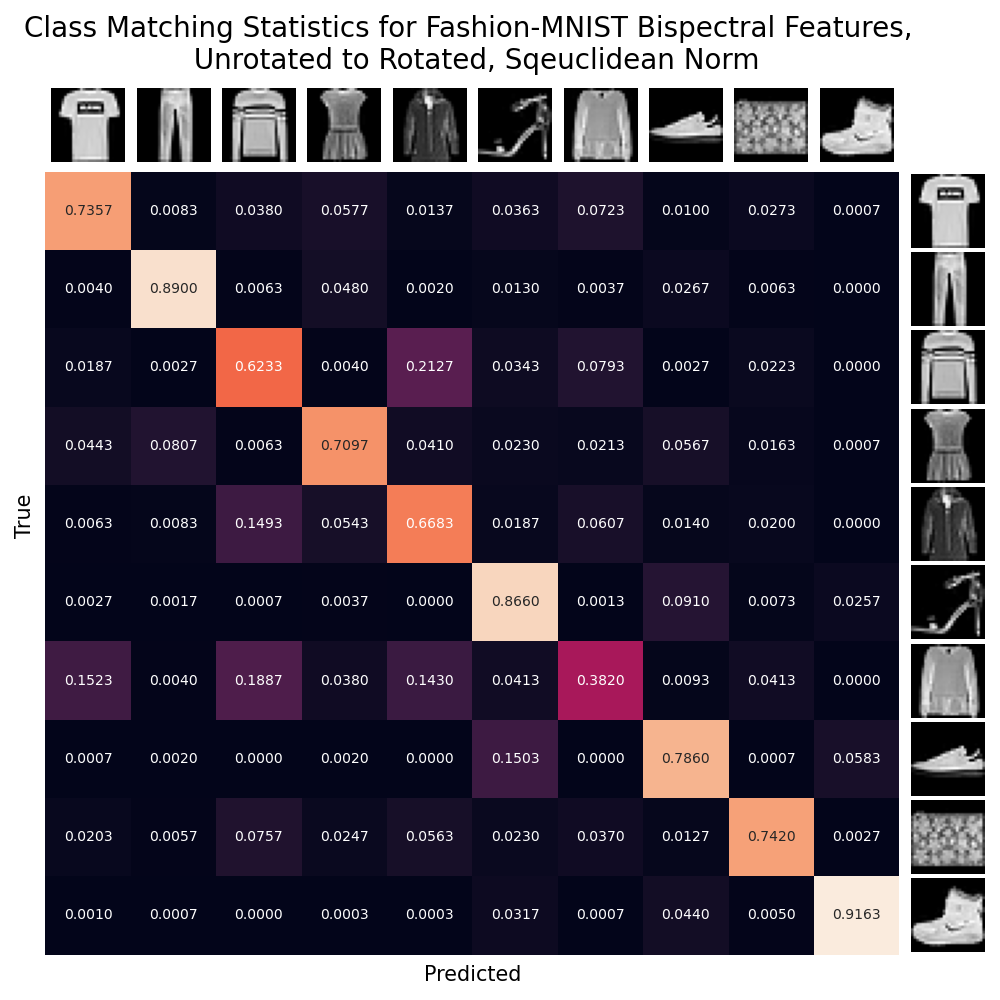}}%
    \quad
    \subfigure[$\cos$ Cost]{
      \includegraphics[width=0.45\linewidth]{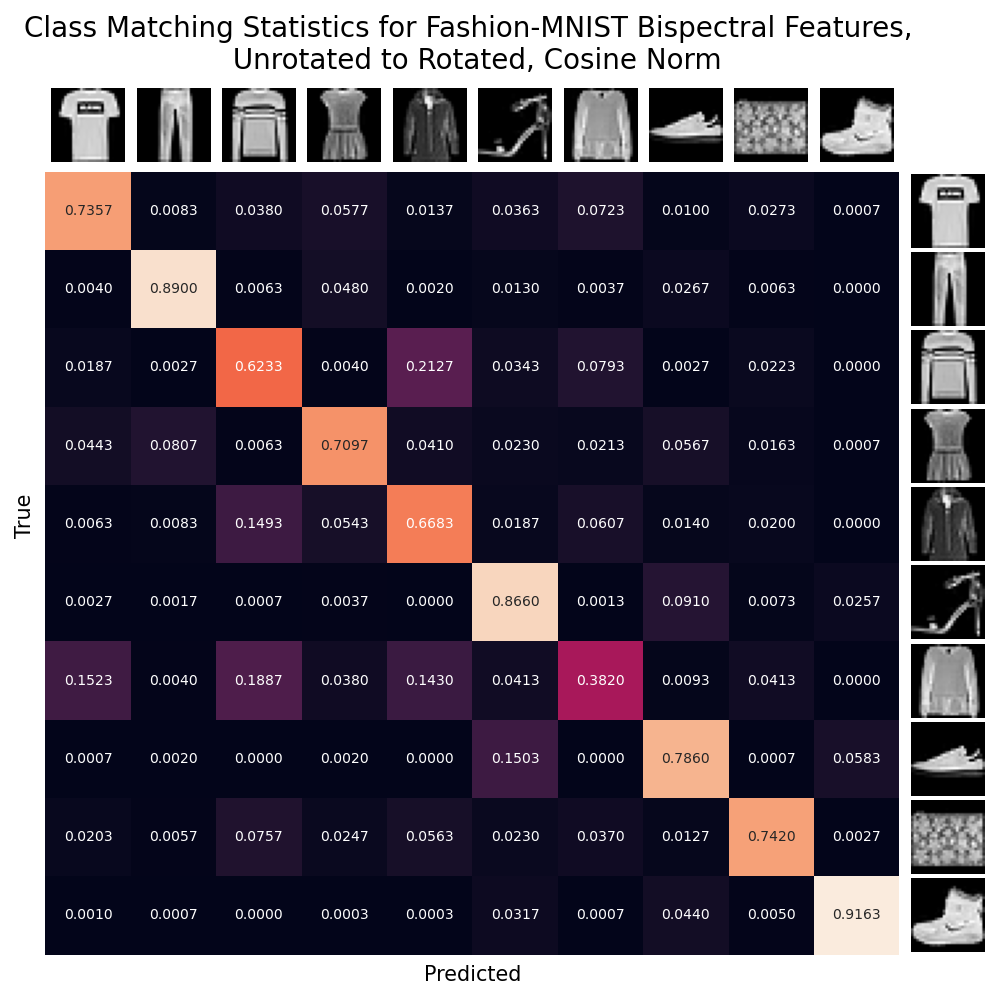}}
    }
\end{figure}

\begin{figure}[htbp]
\floatconts
  {fig:fmnist-raw-real}
  {\caption{Class matching statistics for OT plan from rotated \textsc{fashion-mnist} to unrotated \textsc{fashion-mnist} on raw pixel features.}}
  {%
    \subfigure[$L_1$ Cost]{
      \includegraphics[width=0.45\linewidth]{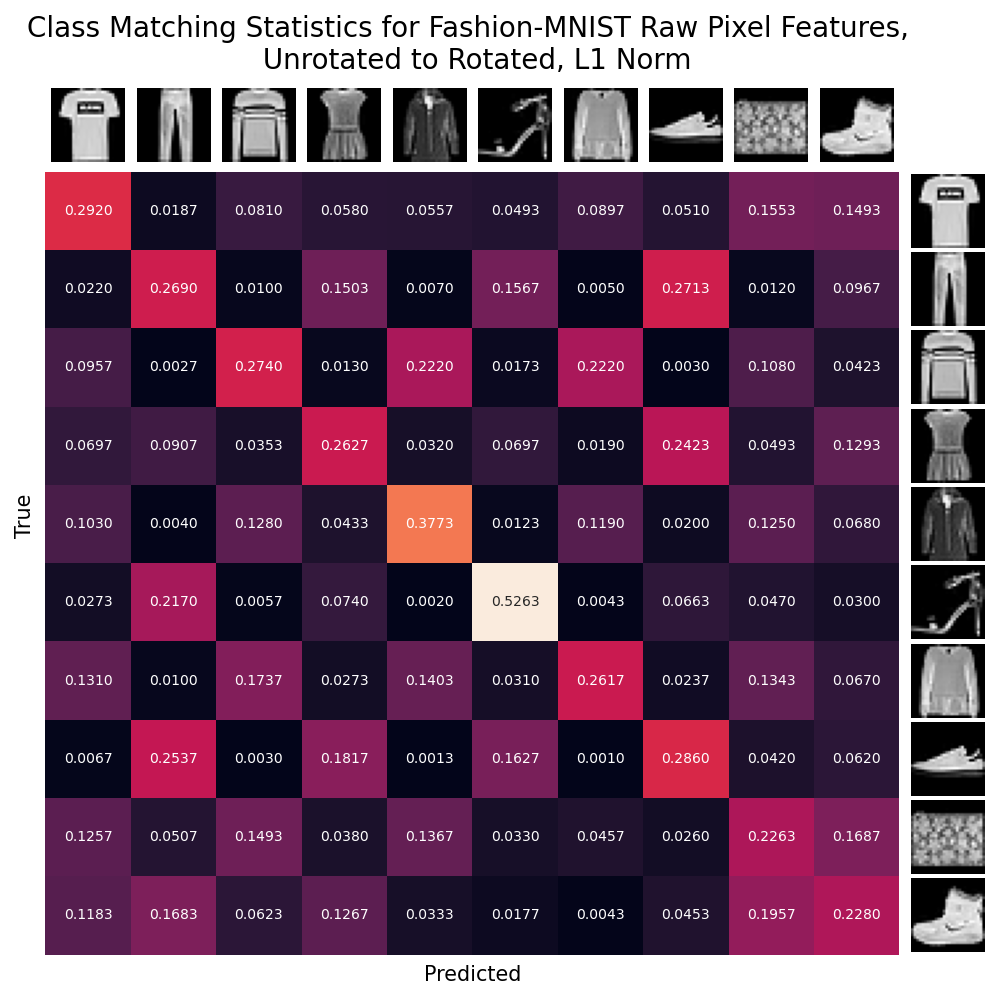}}%
    \quad
    \subfigure[$L_2$ Cost]{
      \includegraphics[width=0.45\linewidth]{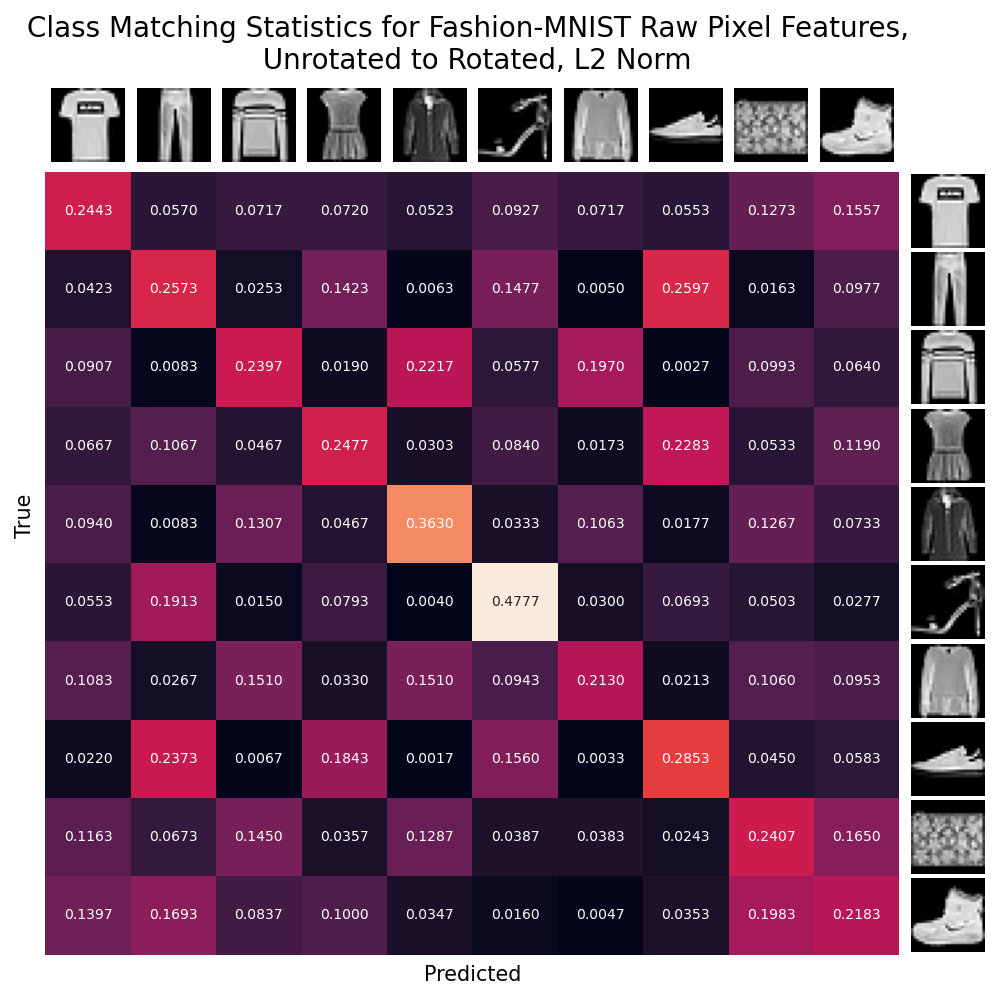}}
      \subfigure[$L_2^2$ Cost]{
      \includegraphics[width=0.45\linewidth]{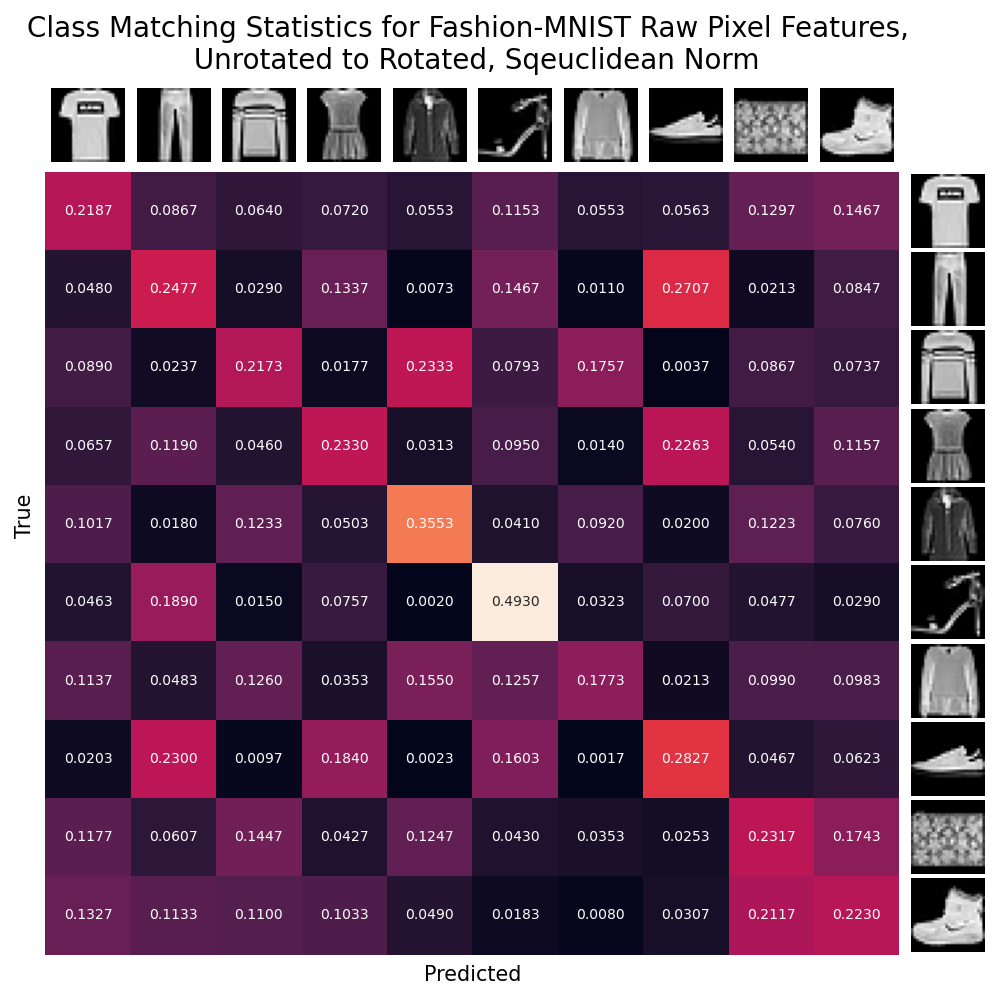}}%
    \quad
    \subfigure[$\cos$ Cost]{
      \includegraphics[width=0.45\linewidth]{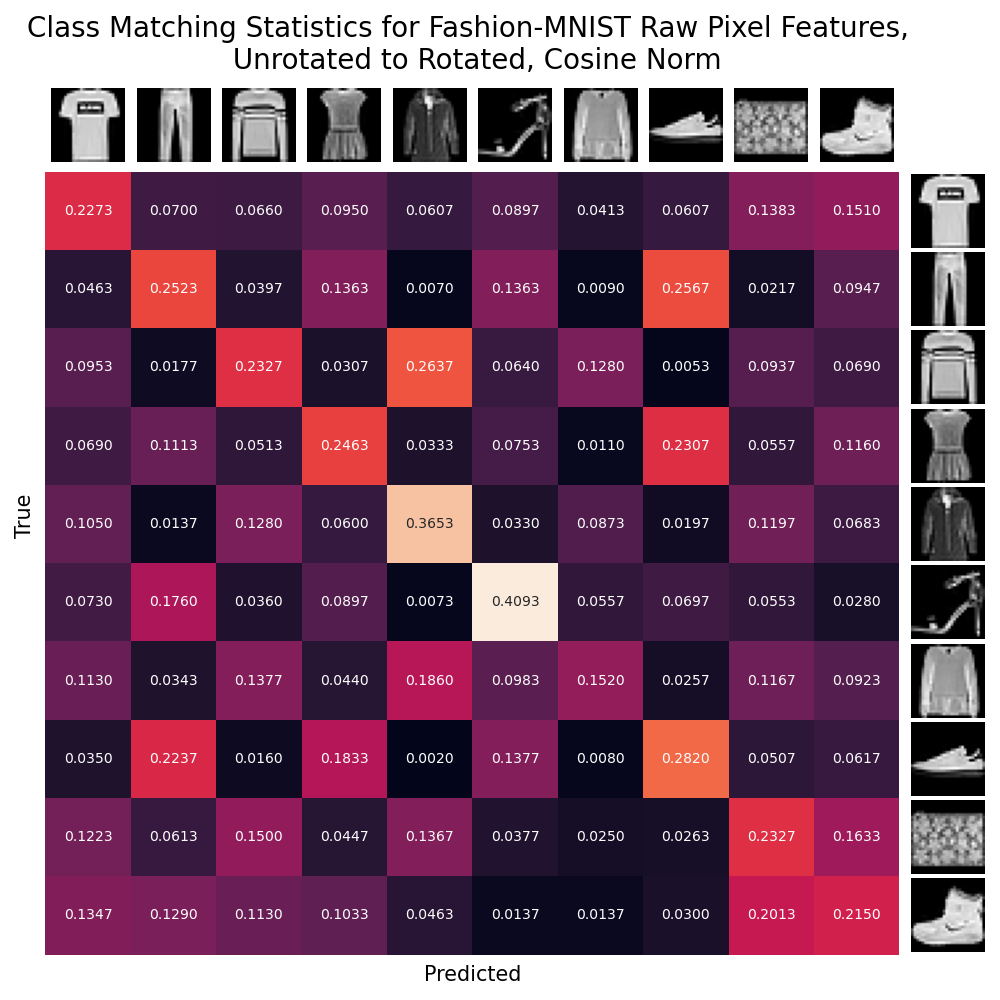}}
      }
\end{figure}

\begin{figure}[htbp]
\floatconts
  {fig:fmnist-bispectral-baseline}
  {\caption{Class matching statistics for OT plan from unrotated \textsc{fashion-mnist} to unrotated \textsc{fashion-mnist} (baseline) on bispectral features.}}
  {%
    \subfigure[$L_1$ Cost]{
      \includegraphics[width=0.45\linewidth]{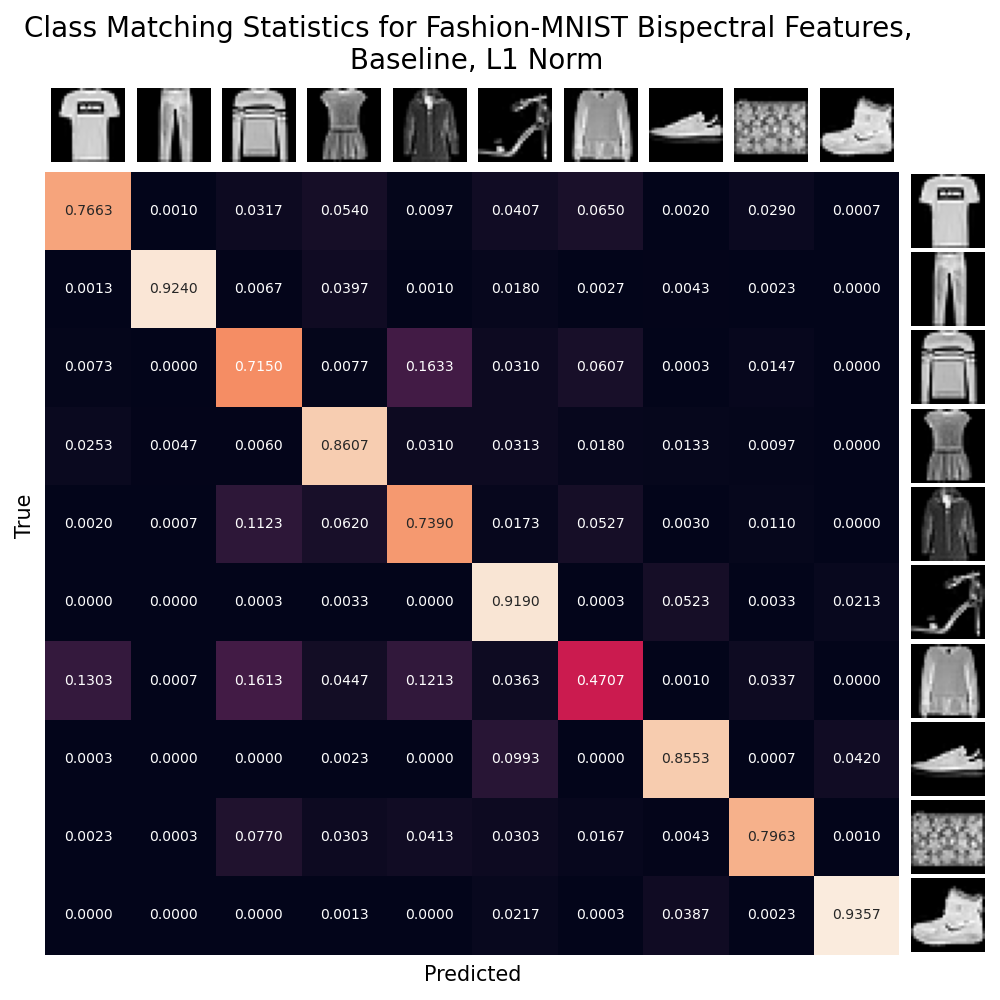}}%
    \quad
    \subfigure[$L_2$ Cost]{
      \includegraphics[width=0.45\linewidth]{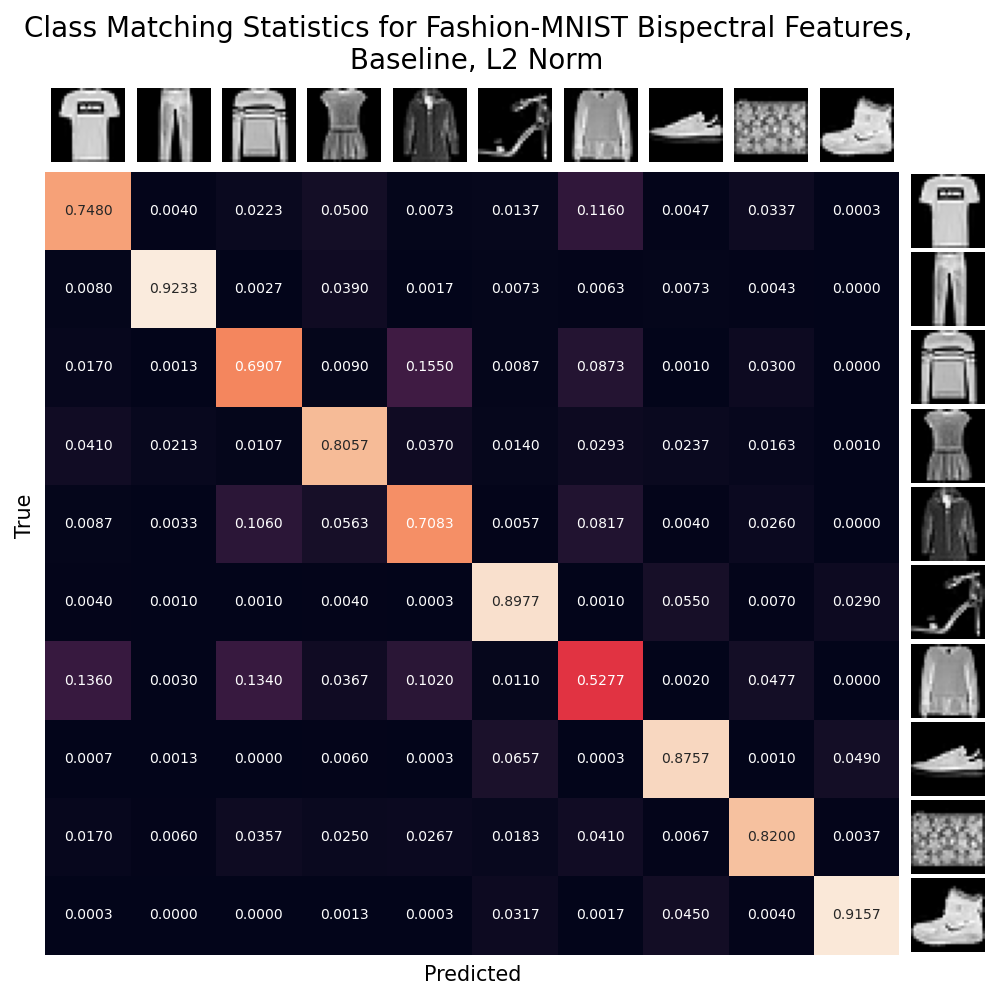}}
      \subfigure[$L_2^2$ Cost]{
      \includegraphics[width=0.45\linewidth]{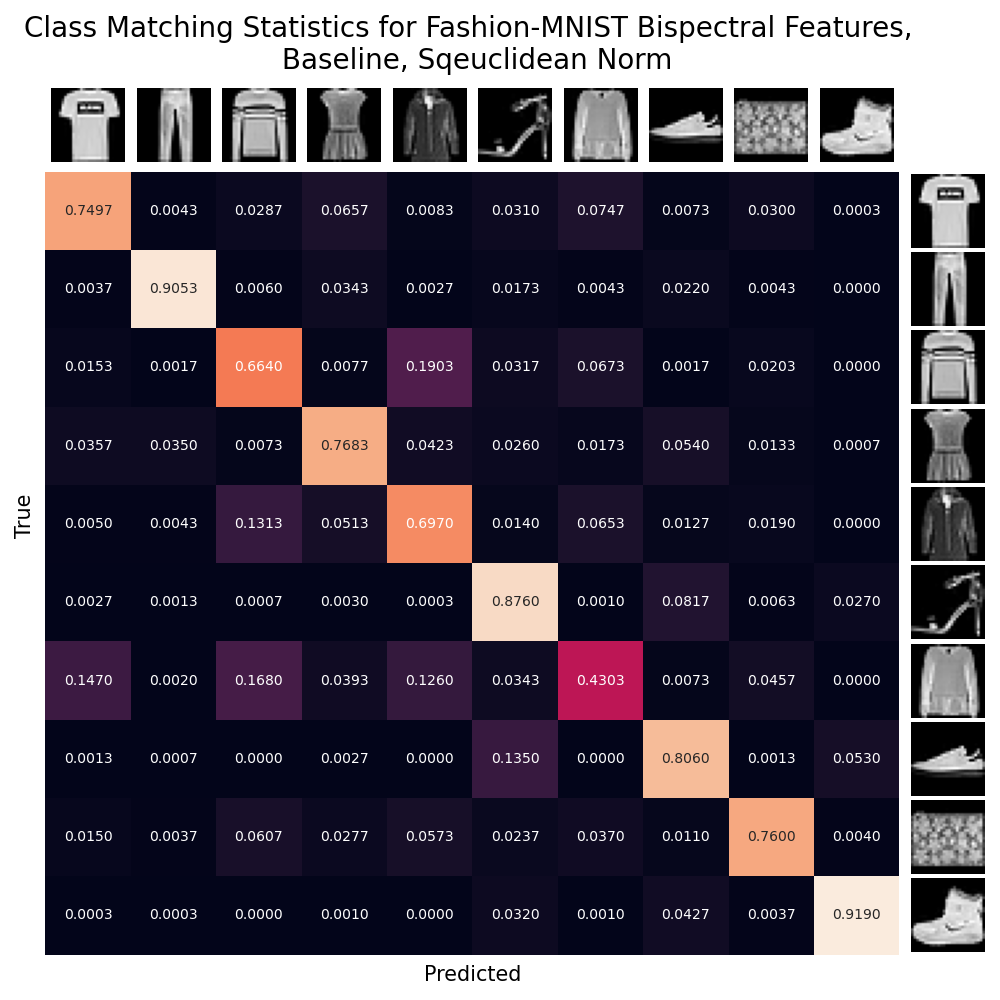}}%
    \quad
    \subfigure[$\cos$ Cost]{
      \includegraphics[width=0.45\linewidth]{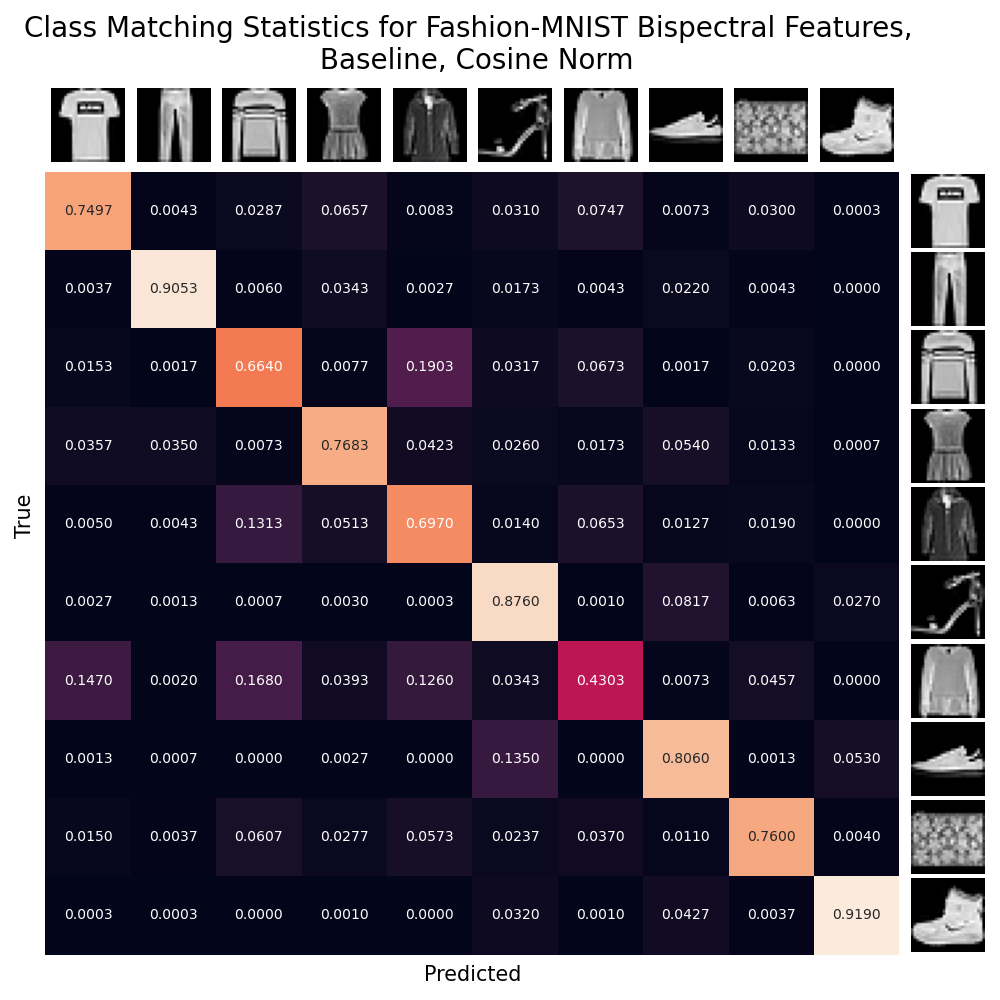}}
    }
\end{figure}

\begin{figure}[htbp]
\floatconts
  {fig:fmnist-raw-baseline}
  {\caption{Class matching statistics for OT plan from unrotated \textsc{fashion-mnist} to unrotated \textsc{fashion-mnist} (baseline) on raw pixel features.}}
  {%
    \subfigure[$L_1$ Cost]{
      \includegraphics[width=0.45\linewidth]{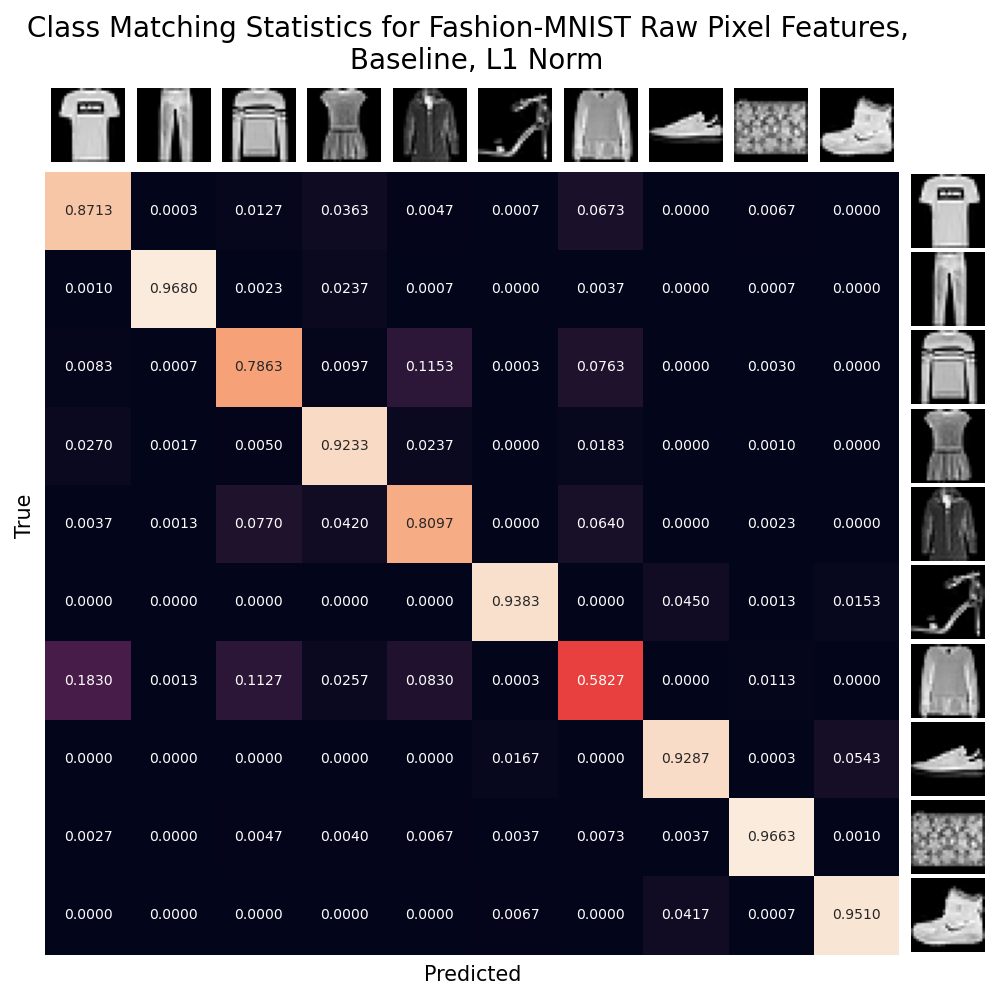}}%
    \quad
    \subfigure[$L_2$ Cost]{
      \includegraphics[width=0.45\linewidth]{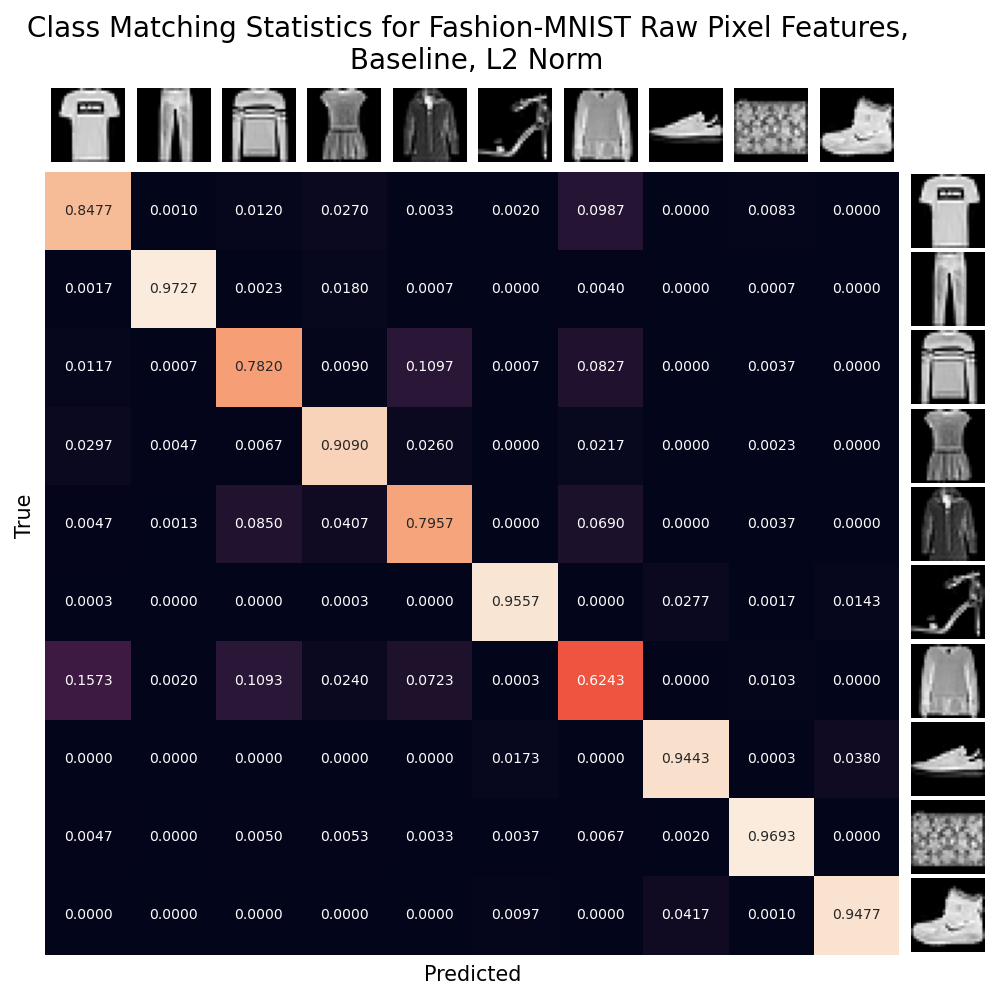}}
      \subfigure[$L_2^2$ Cost]{
      \includegraphics[width=0.45\linewidth]{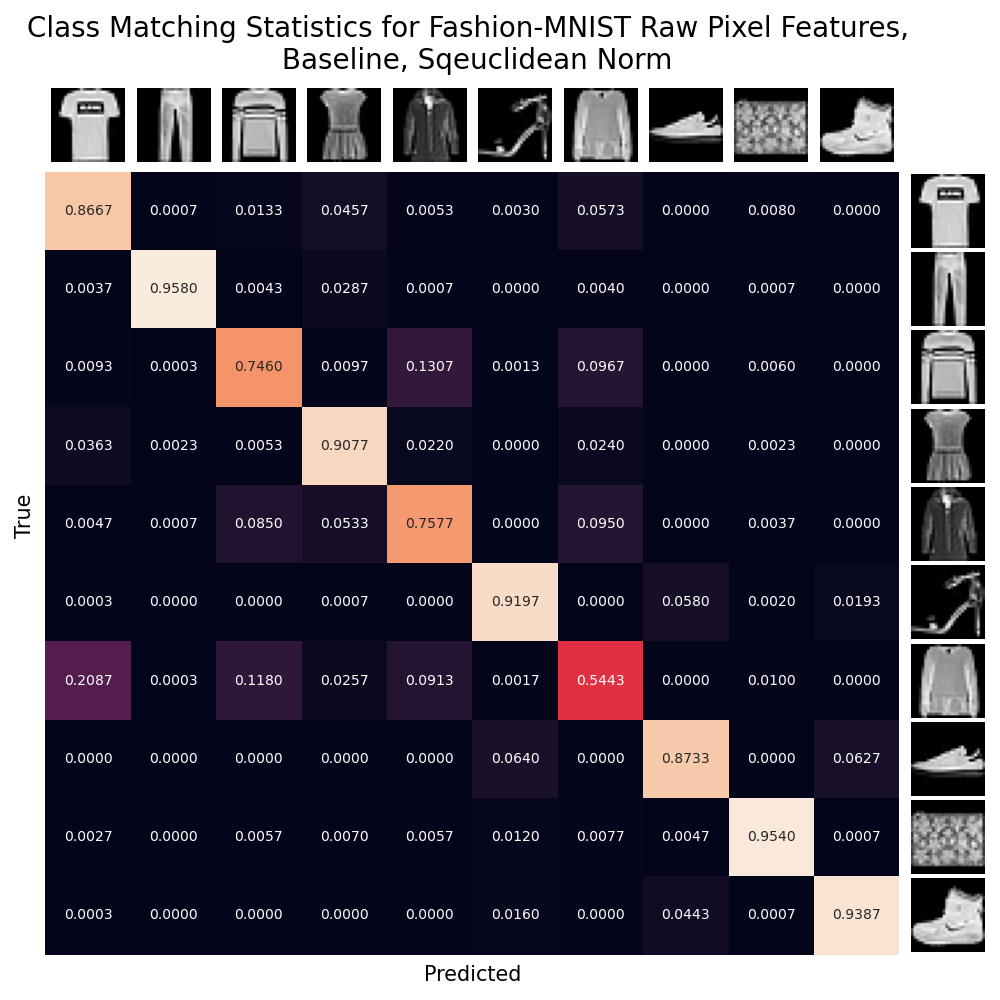}}%
    \quad
    \subfigure[$\cos$ Cost]{
      \includegraphics[width=0.45\linewidth]{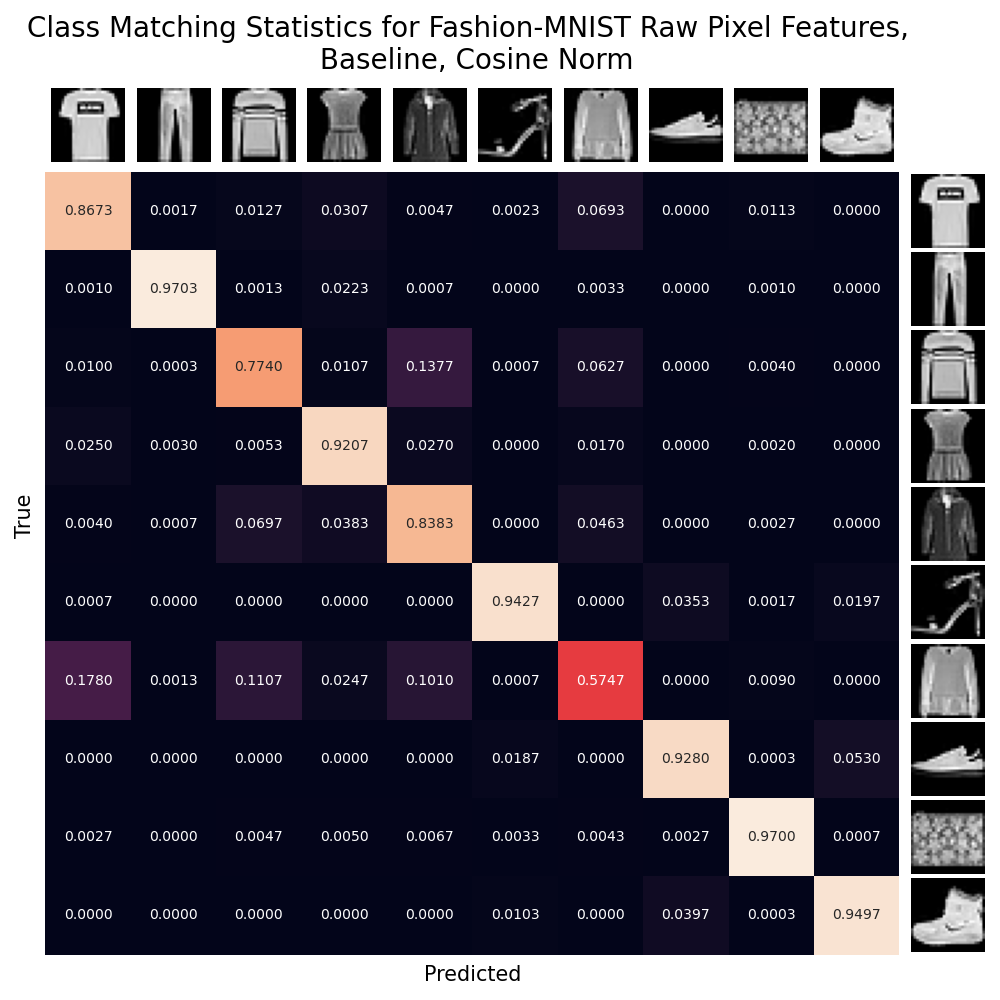}}
    }
\end{figure}

\subsubsection{EMNIST}

We include the confusion matrices for OT on \textsc{fashion-mnist} for the OT matching between rotated and unrotated images in Figures \ref{fig:emnist-bispectral-real} (bispectral features) and \ref{fig:emnist-raw-real} (raw pixel features), and for the baseline experiment (matching unrotated images) in Figures \ref{fig:emnist-bispectral-baseline} (bispectral features) and \ref{fig:emnist-raw-baseline} (raw pixel features). 

\begin{figure}[p]
\floatconts
  {fig:emnist-bispectral-real}
  {\caption{Class matching statistics for OT plan from rotated \textsc{emnist} to unrotated \textsc{emnist} on bispectral features.}}
  {%
    \subfigure[$L_1$ Cost]{
      \includegraphics[width=0.63\linewidth]{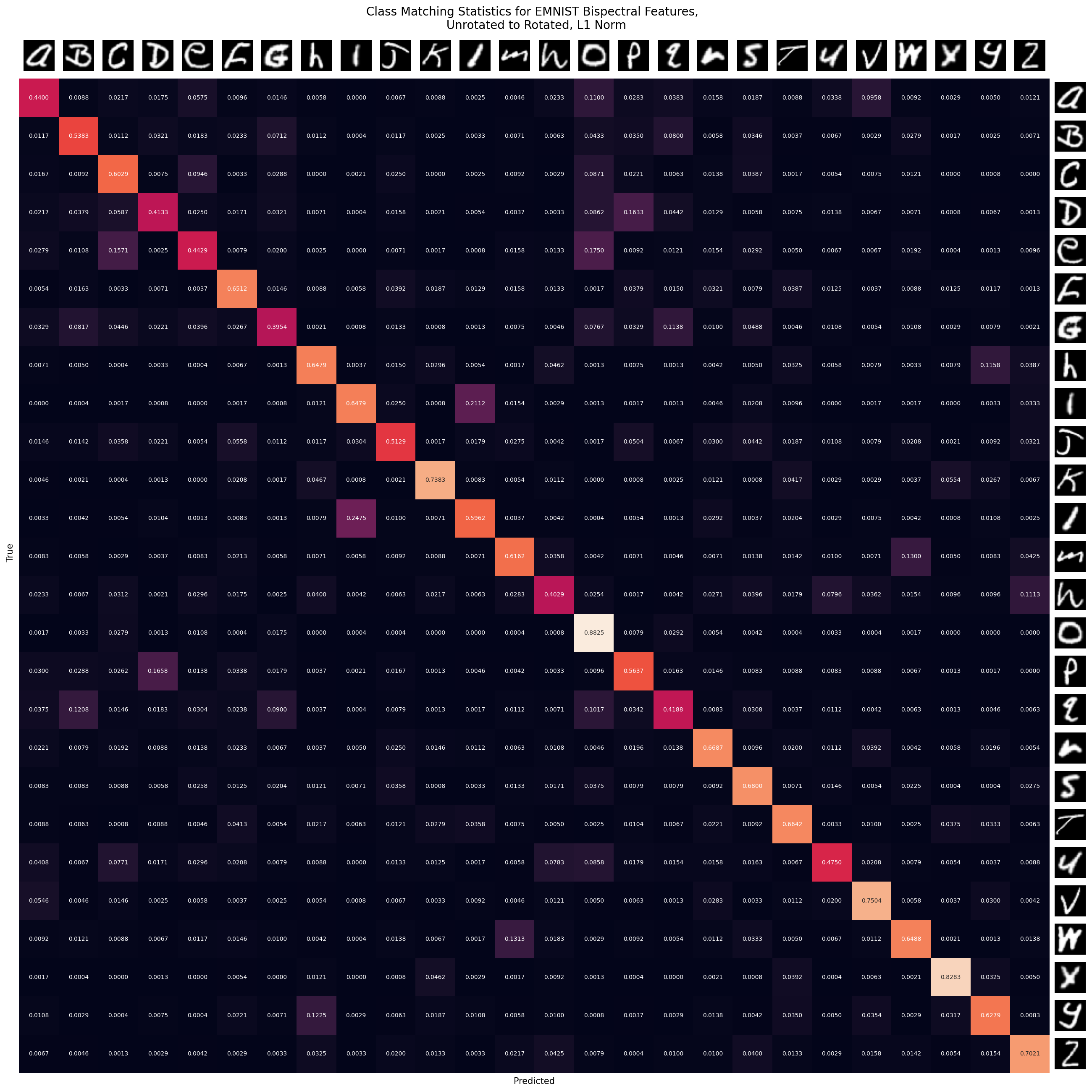}}
    \subfigure[$L_2$ Cost]{
      \includegraphics[width=0.63\linewidth]{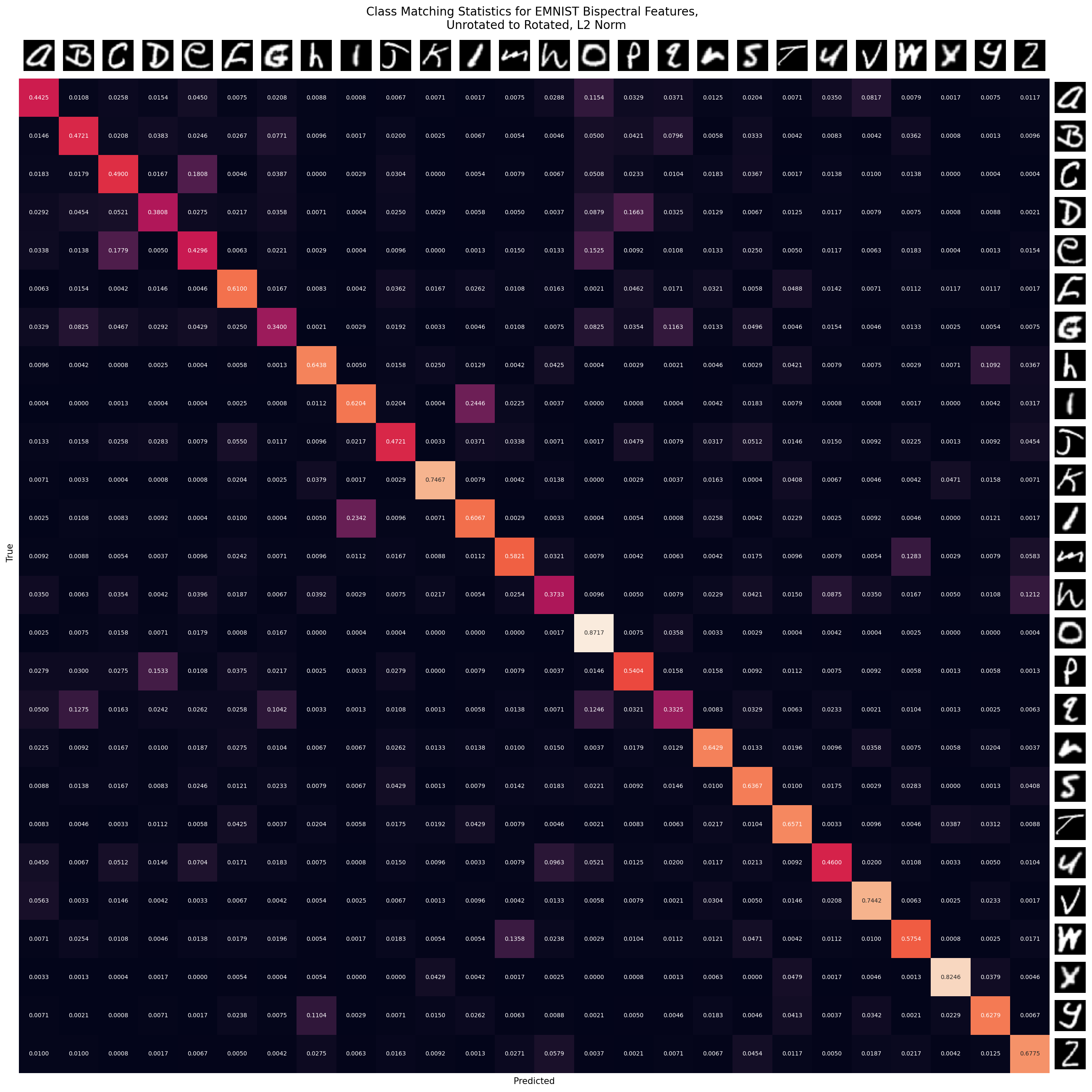}}
  }
\end{figure}

\setcounter{subfigure}{2}
\begin{figure}[p]\ContinuedFloat
\floatconts
  {fig:emnist-bispectral-real}
  {\caption{Class matching statistics for OT plan from rotated \textsc{emnist} to unrotated \textsc{emnist} on bispectral features (continued).}}
  {%
    \subfigure[$L_2^2$ Cost]{
      \includegraphics[width=0.63\linewidth]{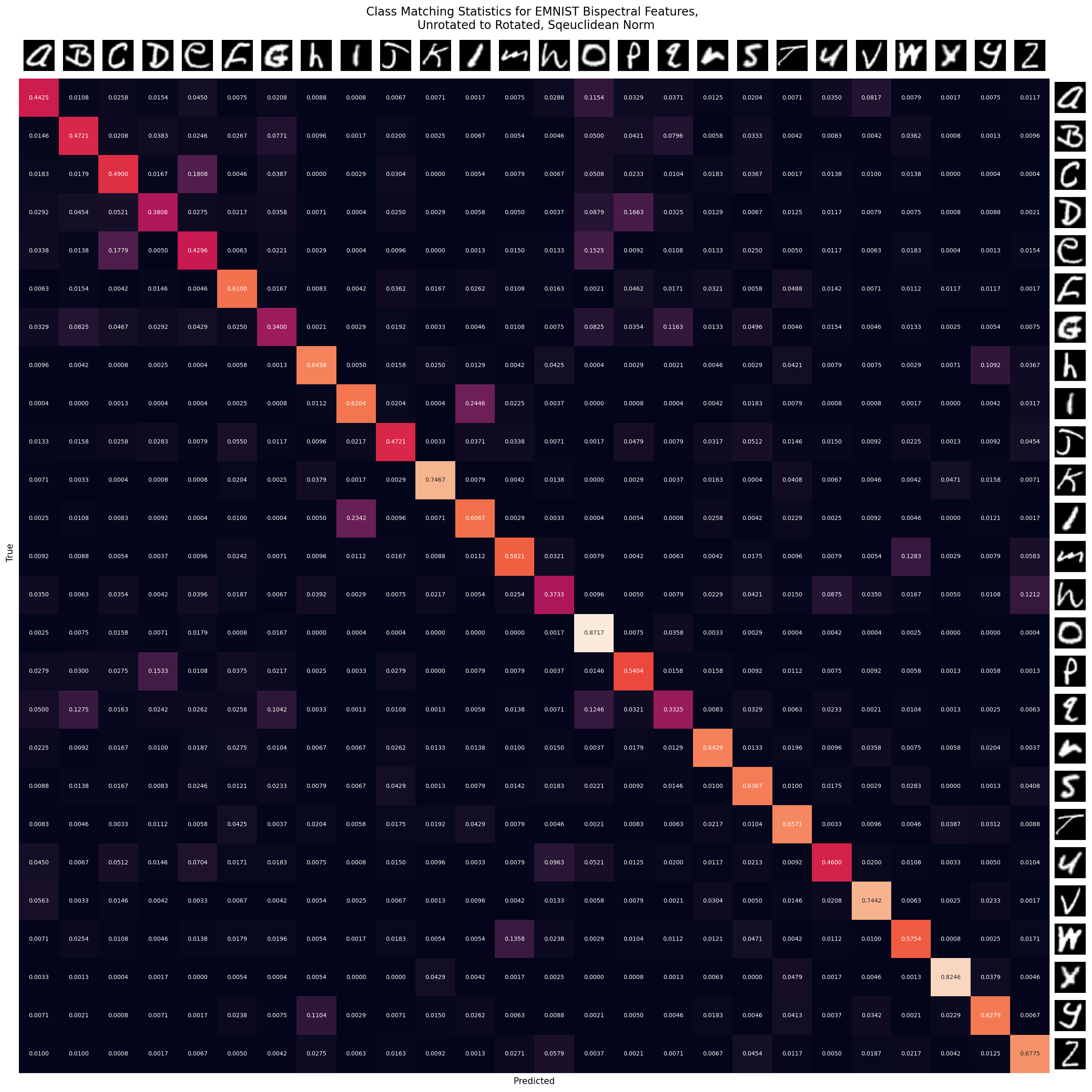}}

    \subfigure[$\cos$ Cost]{
      \includegraphics[width=0.63\linewidth]{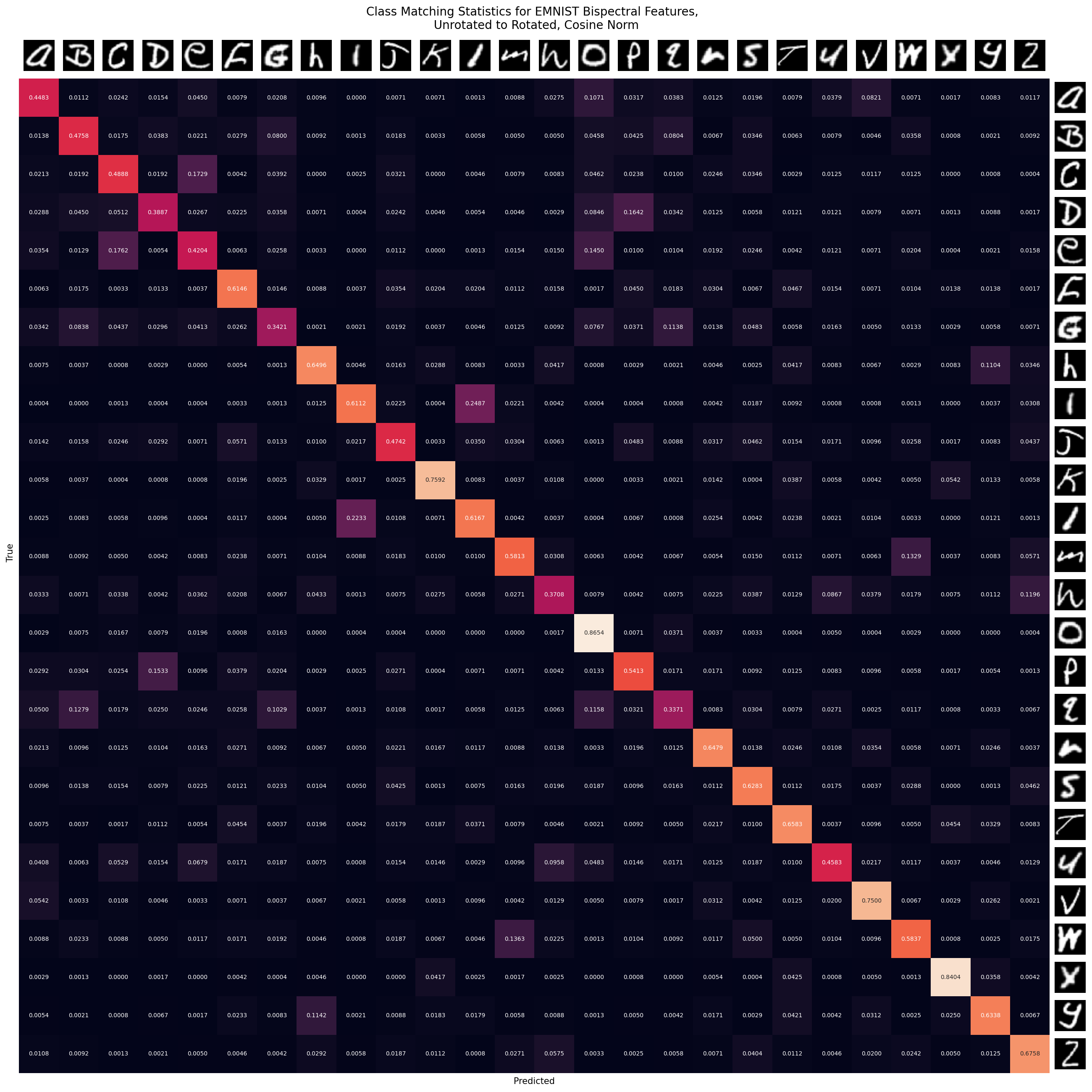}}
  }
\end{figure}

\setcounter{subfigure}{0}
\begin{figure}[p]
\floatconts
  {fig:emnist-raw-real}
  {\caption{Class matching statistics for OT plan from rotated \textsc{emnist} to unrotated \textsc{emnist} on raw pixel features.}}
  {%
    \subfigure[$L_1$ Cost]{
      \includegraphics[width=0.63\linewidth]{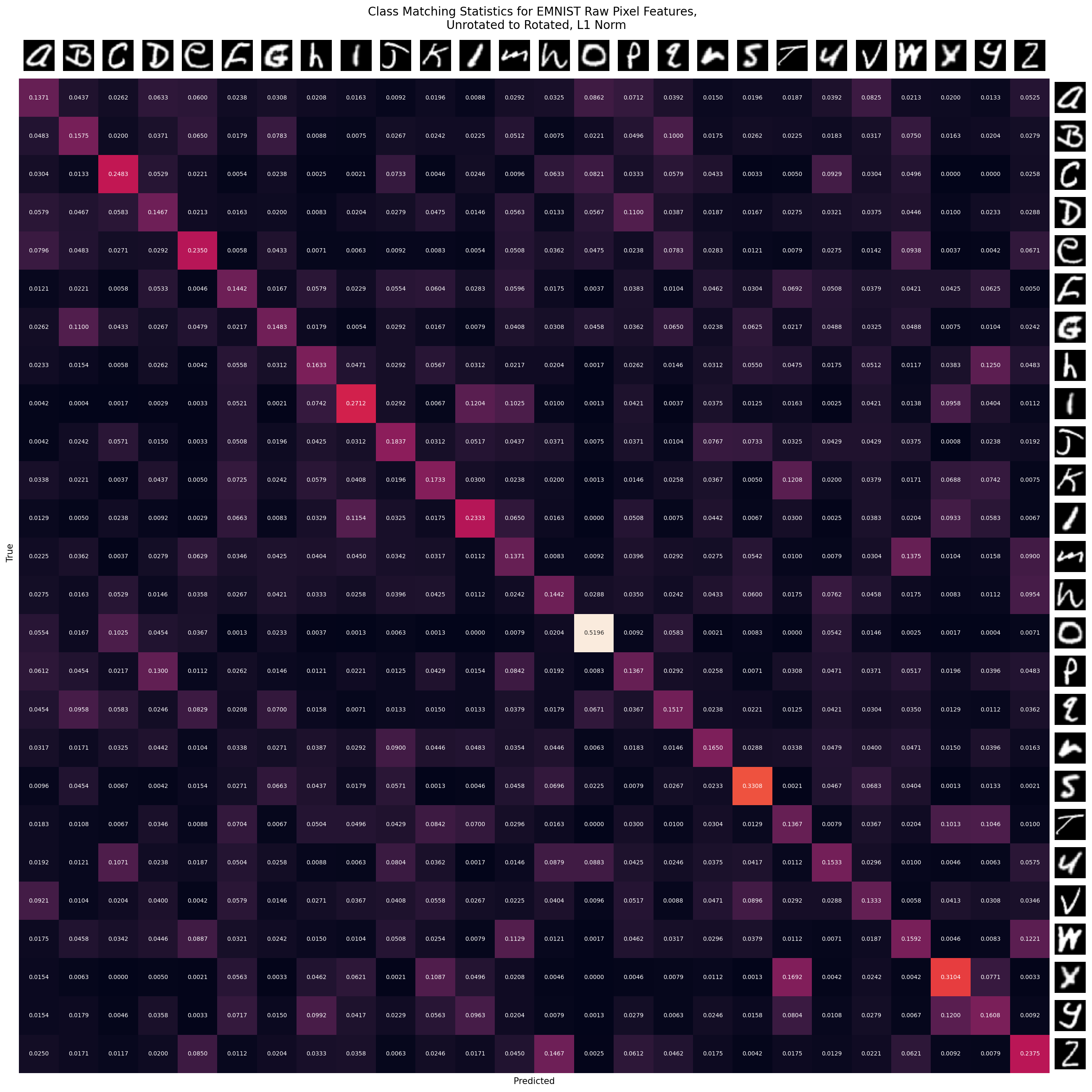}}

    \subfigure[$L_2$ Cost]{
      \includegraphics[width=0.63\linewidth]{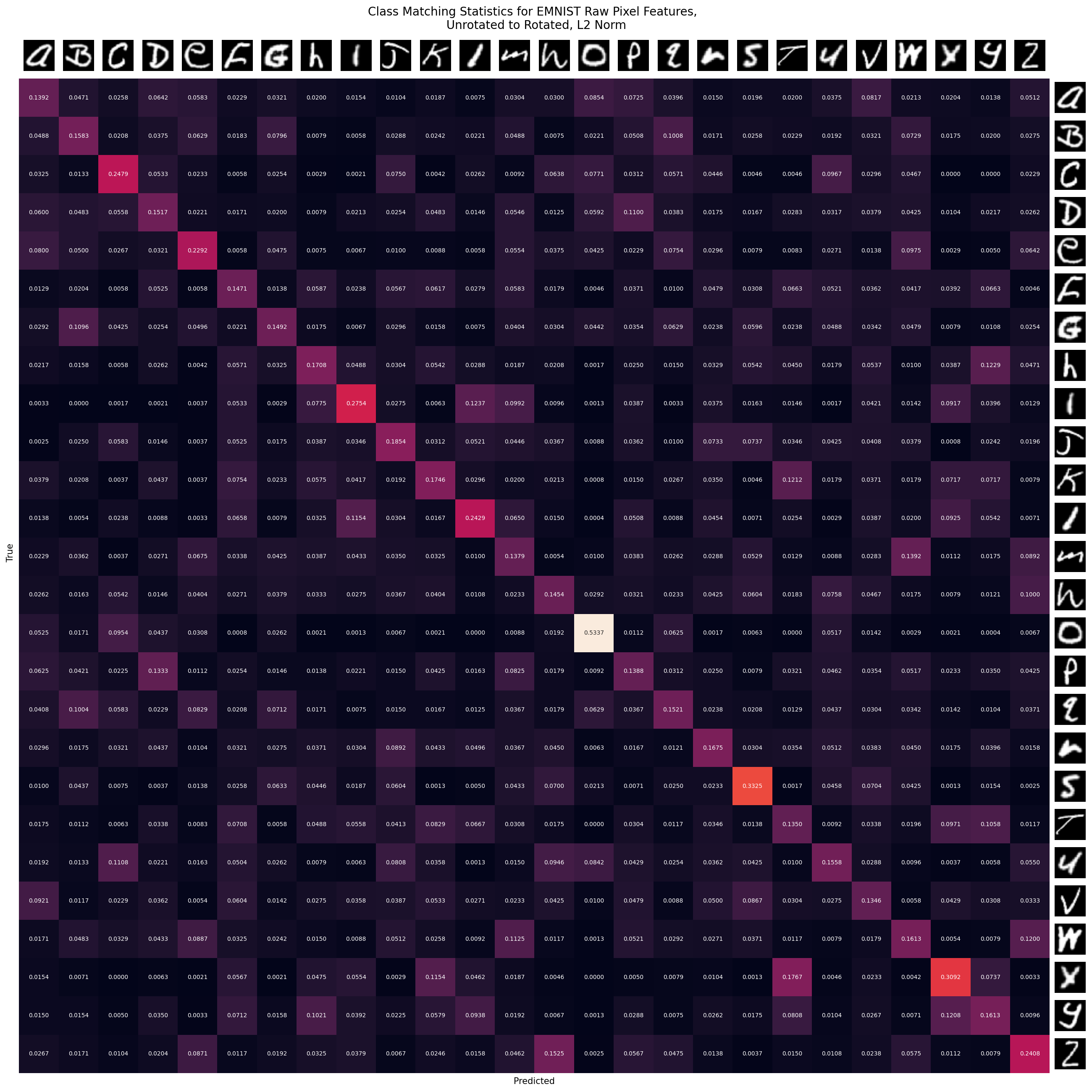}}
  }
\end{figure}

\setcounter{subfigure}{2}
\begin{figure}[p]\ContinuedFloat
\floatconts
  {fig:emnist-raw-real}
  {\caption[]{Class matching statistics for OT plan from rotated \textsc{emnist} to unrotated \textsc{emnist} on raw pixel features (continued).}}
  {%
    \subfigure[$L_2^2$ Cost]{
      \includegraphics[width=0.63\linewidth]{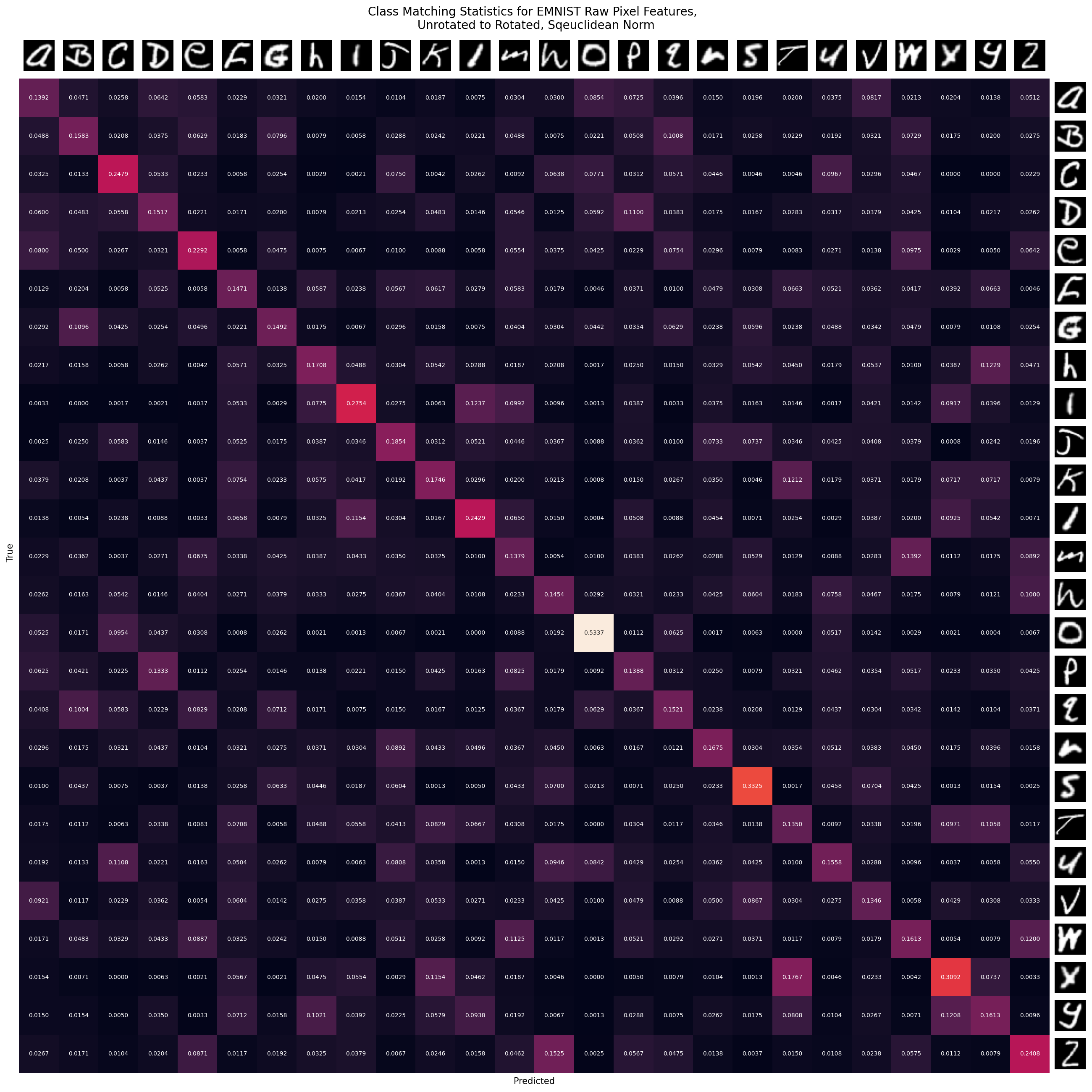}}

    \subfigure[$\cos$ Cost]{
      \includegraphics[width=0.63\linewidth]{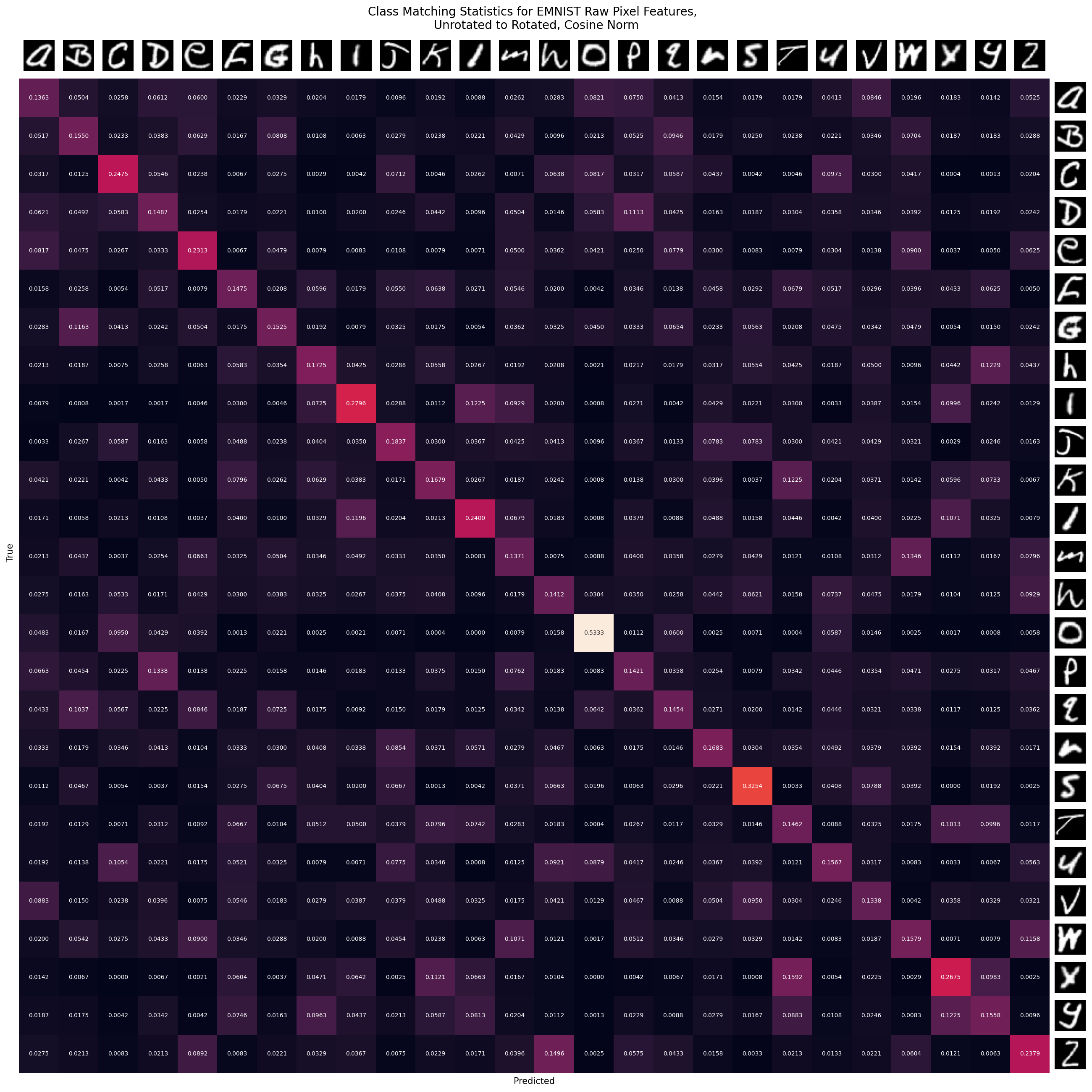}}
  }
\end{figure}

\setcounter{subfigure}{0}
\begin{figure}[p]
\floatconts
  {fig:emnist-bispectral-baseline}
  {\caption{Class matching statistics for OT plan from unrotated \textsc{emnist} to unrotated \textsc{emnist} (baseline) on bispectral features.}}
  {%
    \subfigure[$L_1$ Cost]{
      \includegraphics[width=0.63\linewidth]{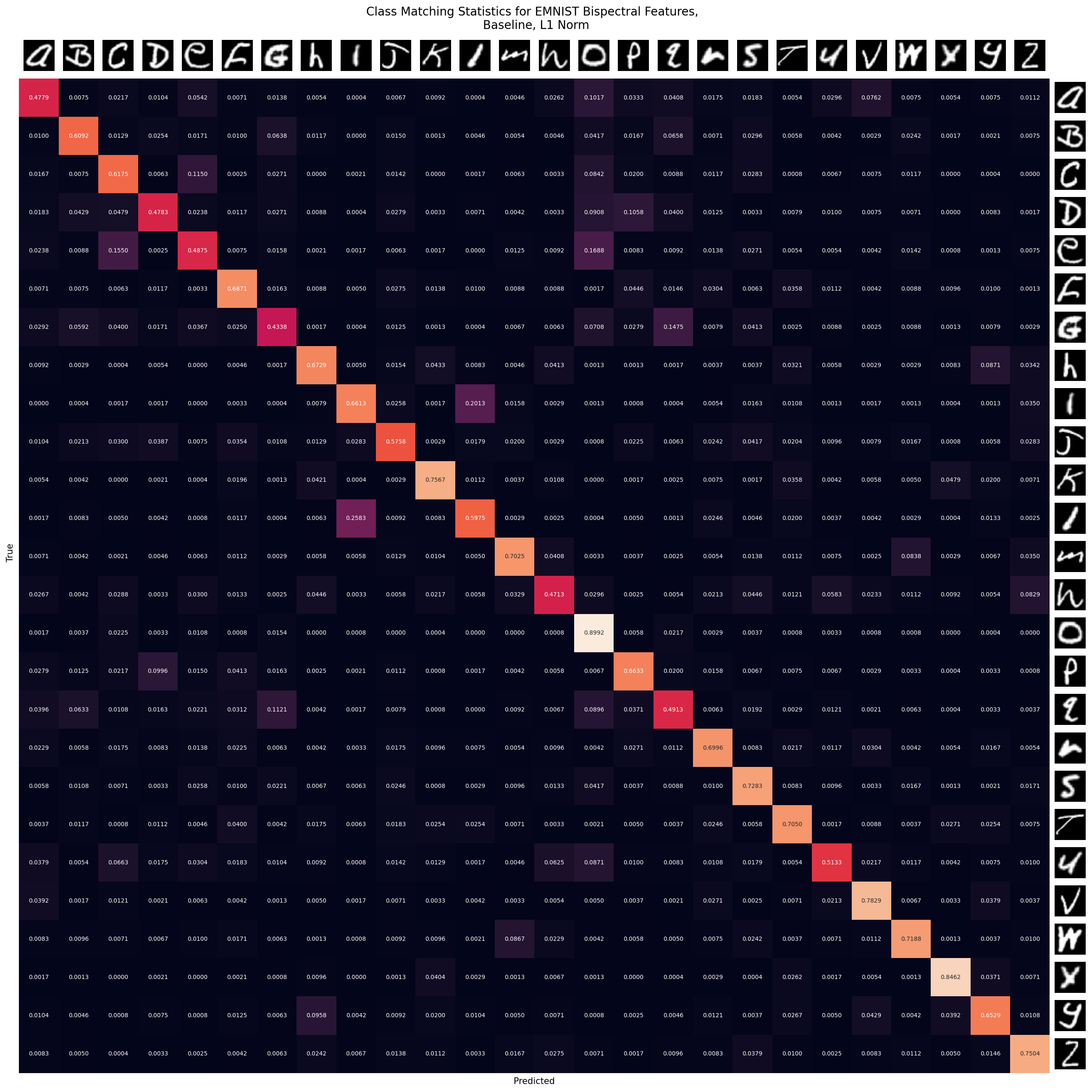}}

    \subfigure[$L_2$ Cost]{
      \includegraphics[width=0.63\linewidth]{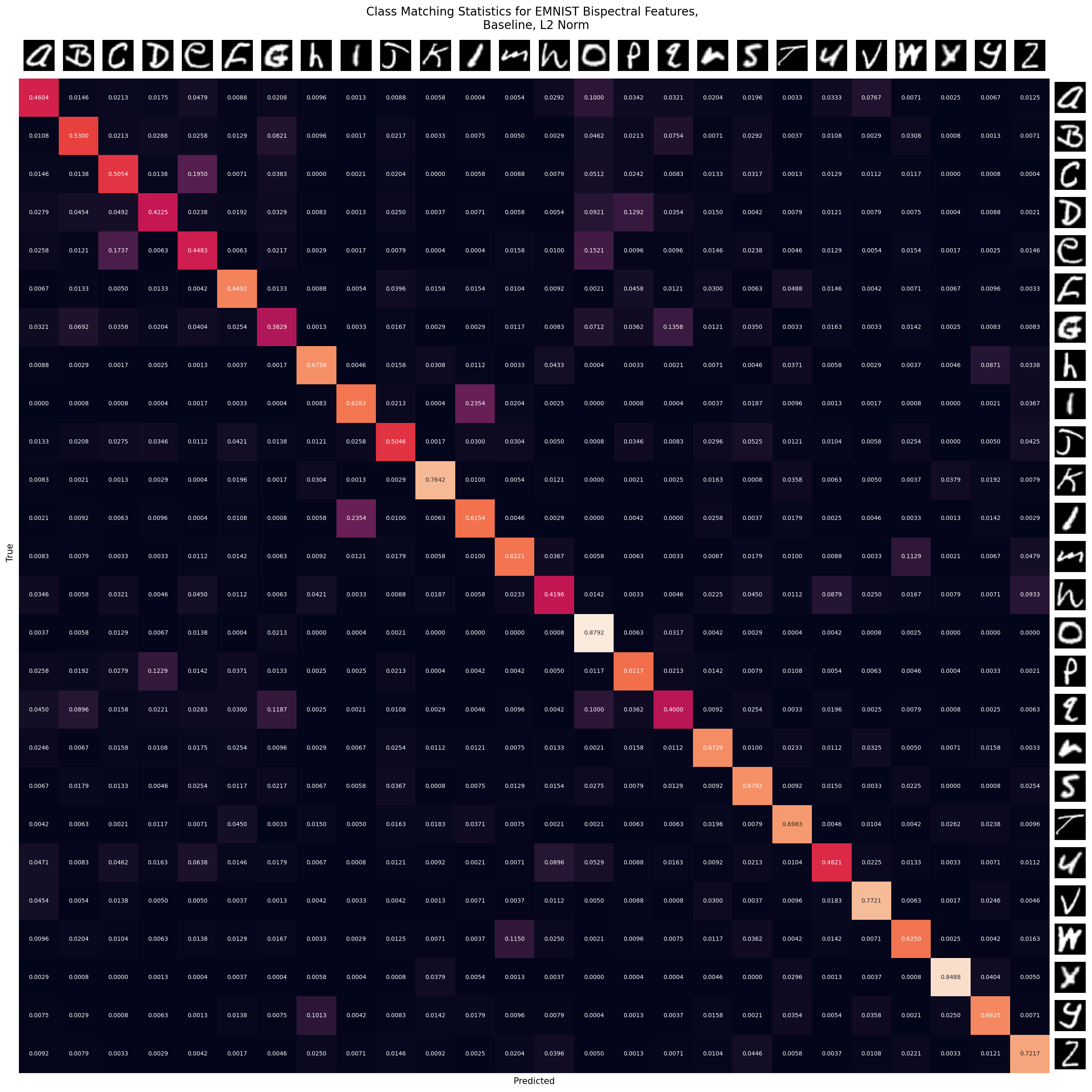}}
  }
\end{figure}

\setcounter{subfigure}{2}
\begin{figure}[p]\ContinuedFloat
\floatconts
  {fig:emnist-bispectral-baseline}
  {\caption[]{Class matching statistics for OT plan from unrotated \textsc{emnist} to unrotated \textsc{emnist} (baseline) on bispectral features (continued).}}
  {%
    \subfigure[$L_2^2$ Cost]{
      \includegraphics[width=0.63\linewidth]{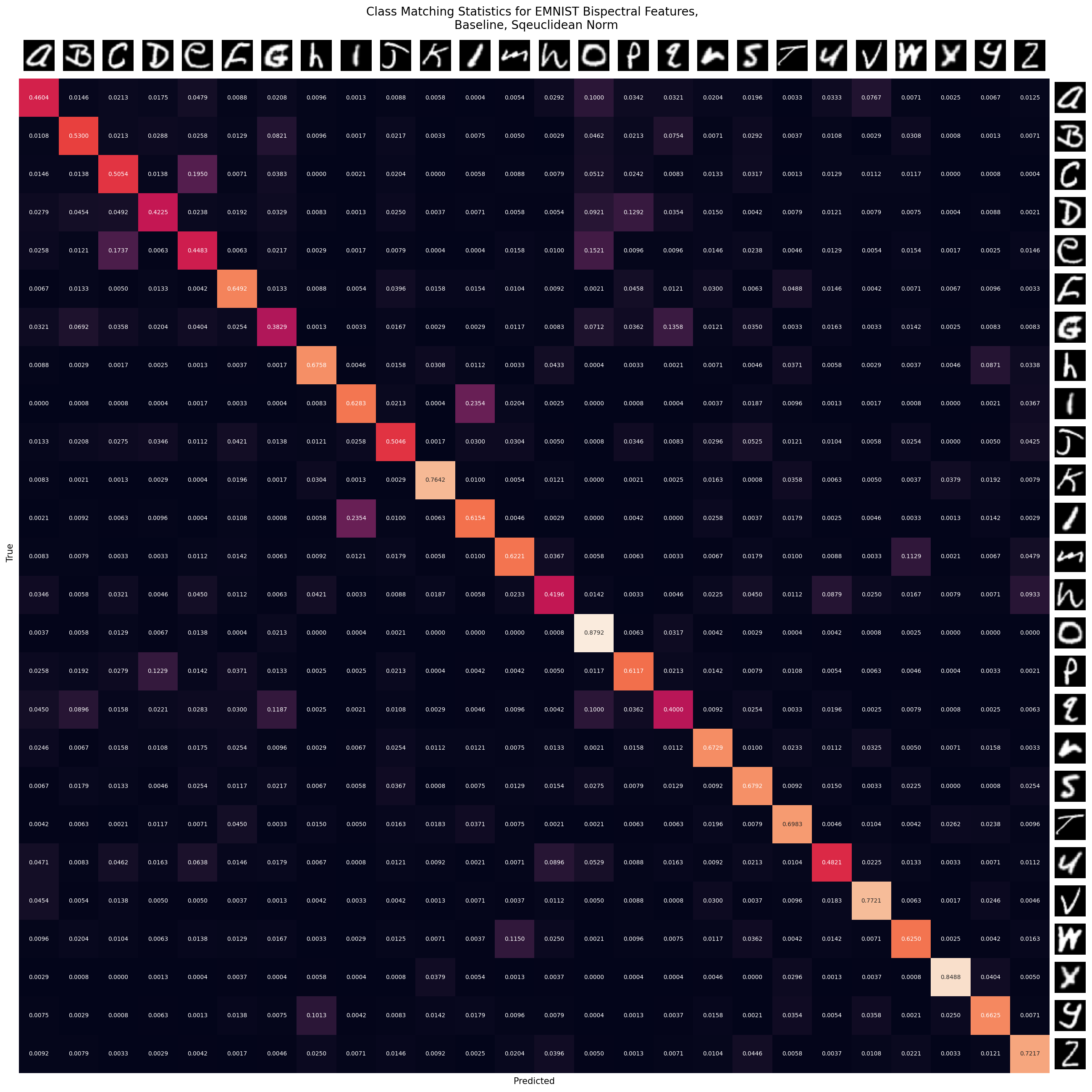}}

    \subfigure[$\cos$ Cost]{
      \includegraphics[width=0.63\linewidth]{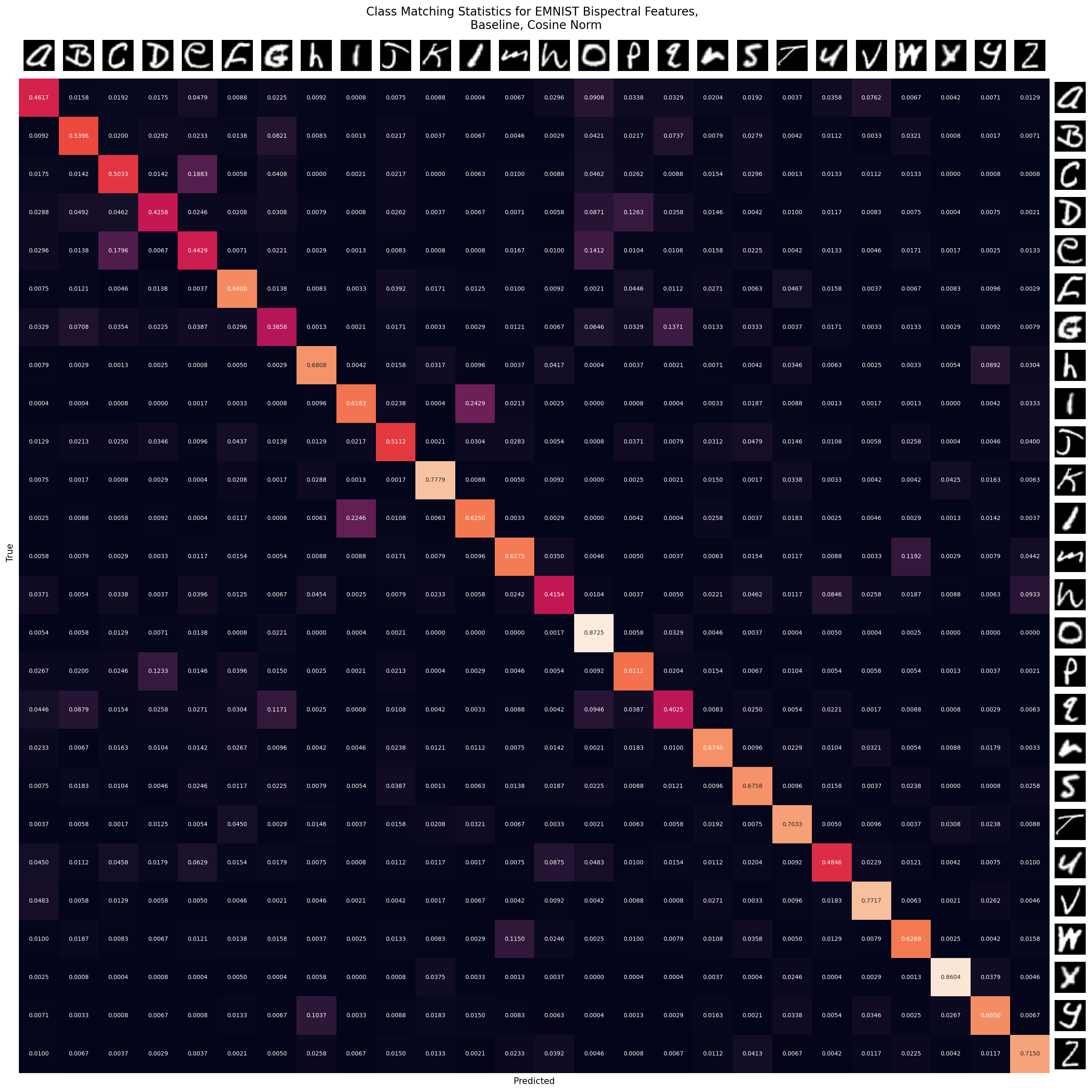}}
  }
\end{figure}

\setcounter{subfigure}{0}
\begin{figure}[p]
\floatconts
  {fig:emnist-raw-baseline}
  {\caption{Class matching statistics for OT plan from unrotated \textsc{emnist} to unrotated \textsc{emnist} (baseline) on raw pixel features.}}
  {%
    \subfigure[$L_1$ Cost]{
      \includegraphics[width=0.63\linewidth]{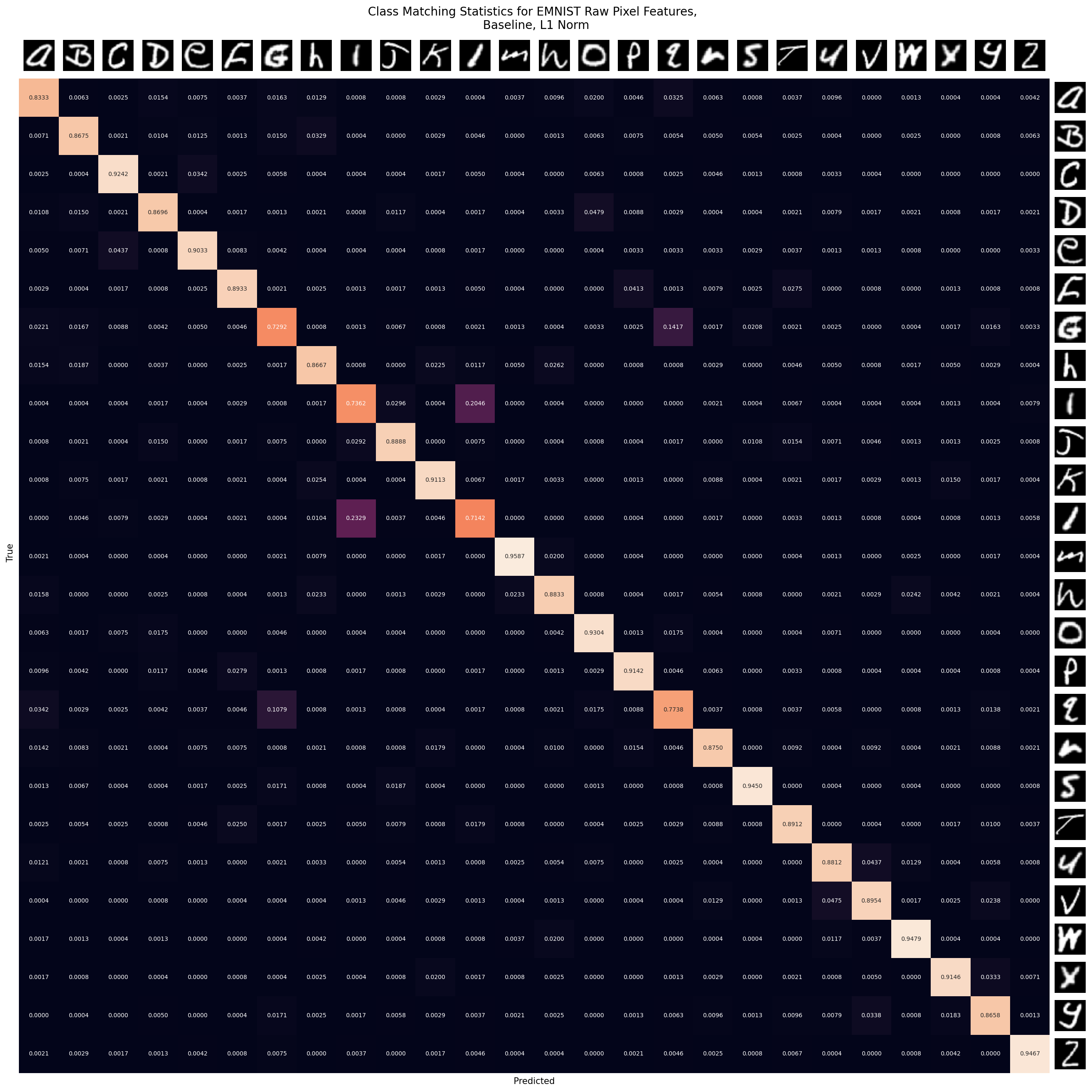}}

    \subfigure[$L_2$ Cost]{
      \includegraphics[width=0.63\linewidth]{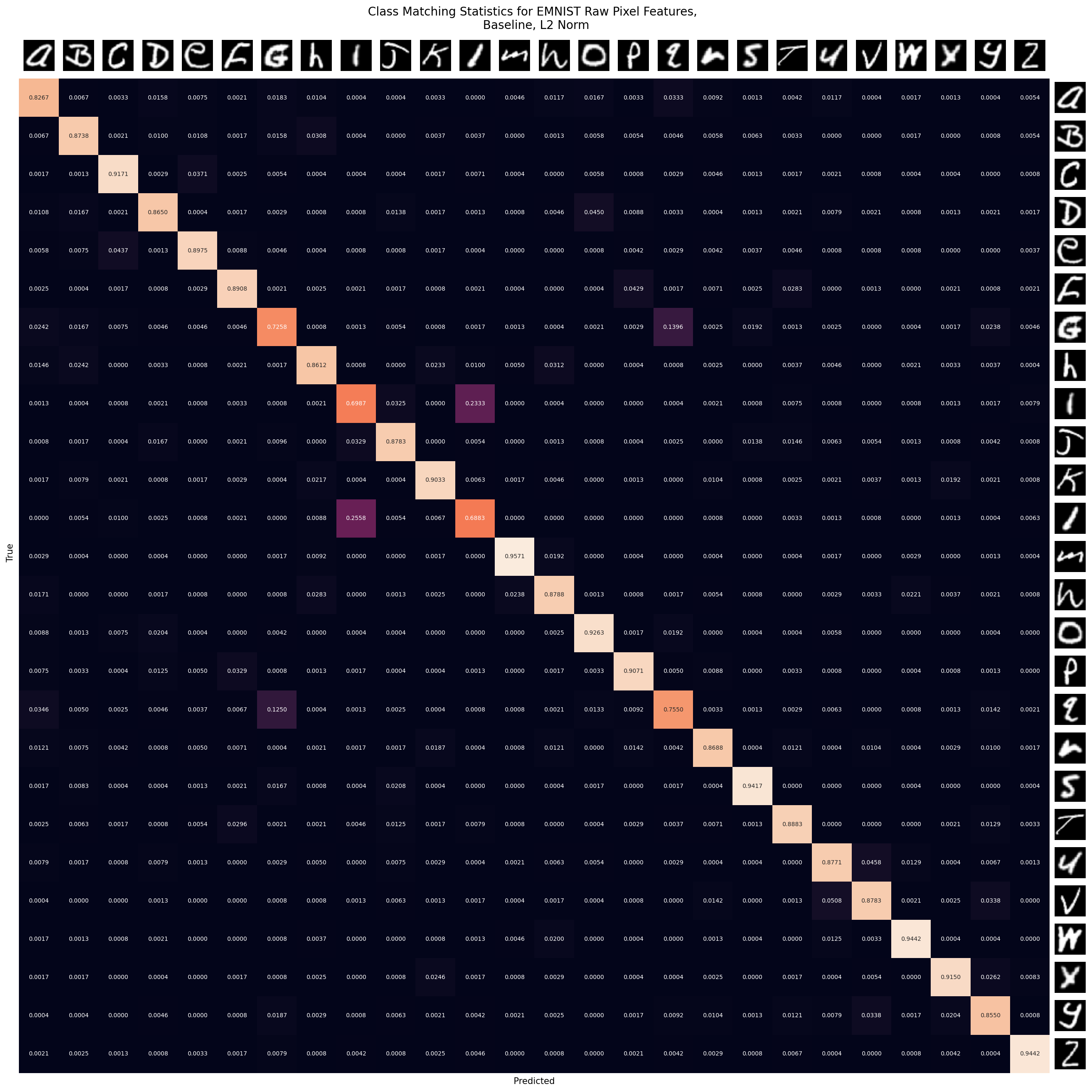}}
  }
\end{figure}

\setcounter{subfigure}{2}
\begin{figure}[p]\ContinuedFloat
\floatconts
  {fig:emnist-raw-baseline}
  {\caption[]{Class matching statistics for OT plan from unrotated \textsc{emnist} to unrotated \textsc{emnist} (baseline) on raw pixel features (continued).}}
  {%
    \subfigure[$L_2^2$ Cost]{
      \includegraphics[width=0.63\linewidth]{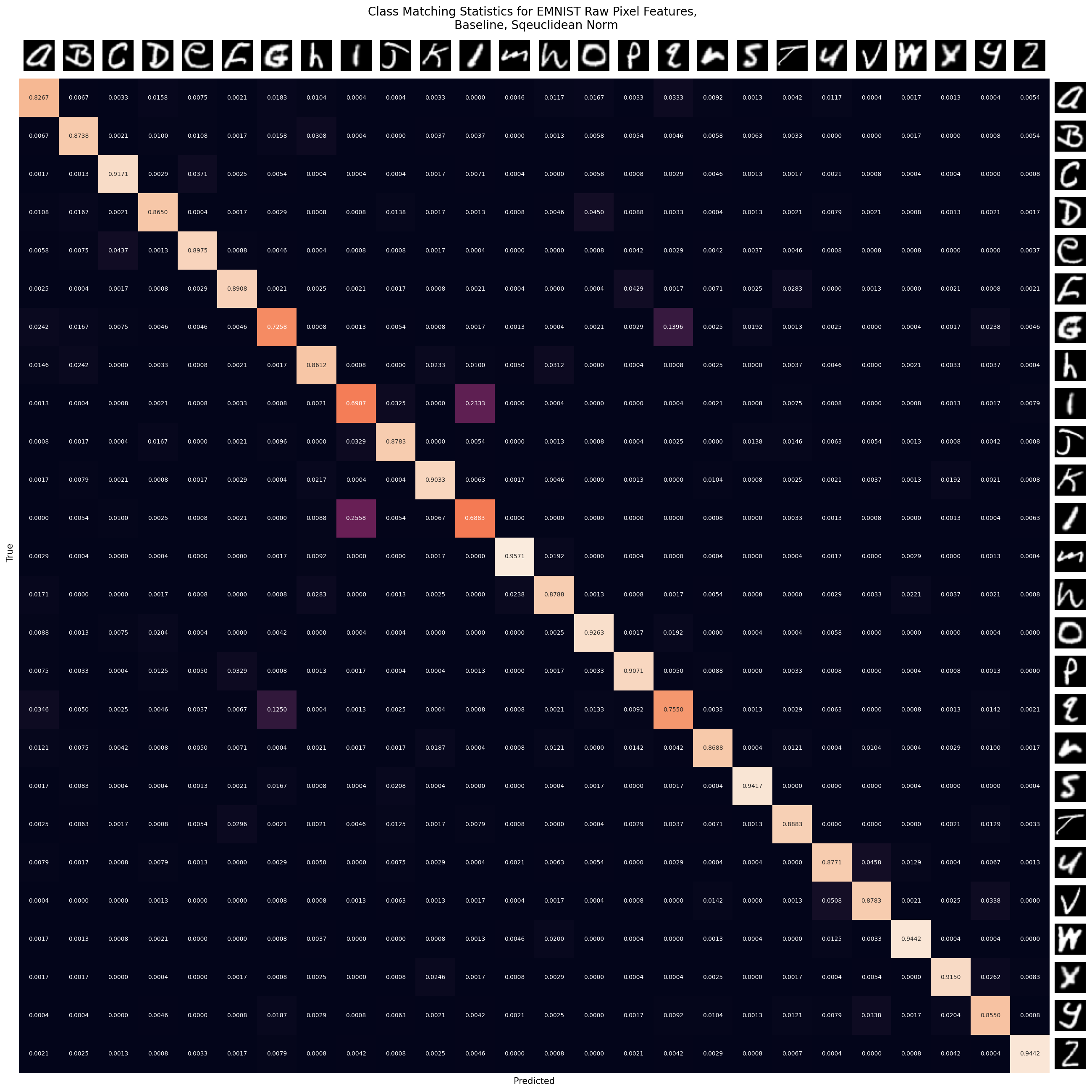}}

    \subfigure[$\cos$ Cost]{
      \includegraphics[width=0.63\linewidth]{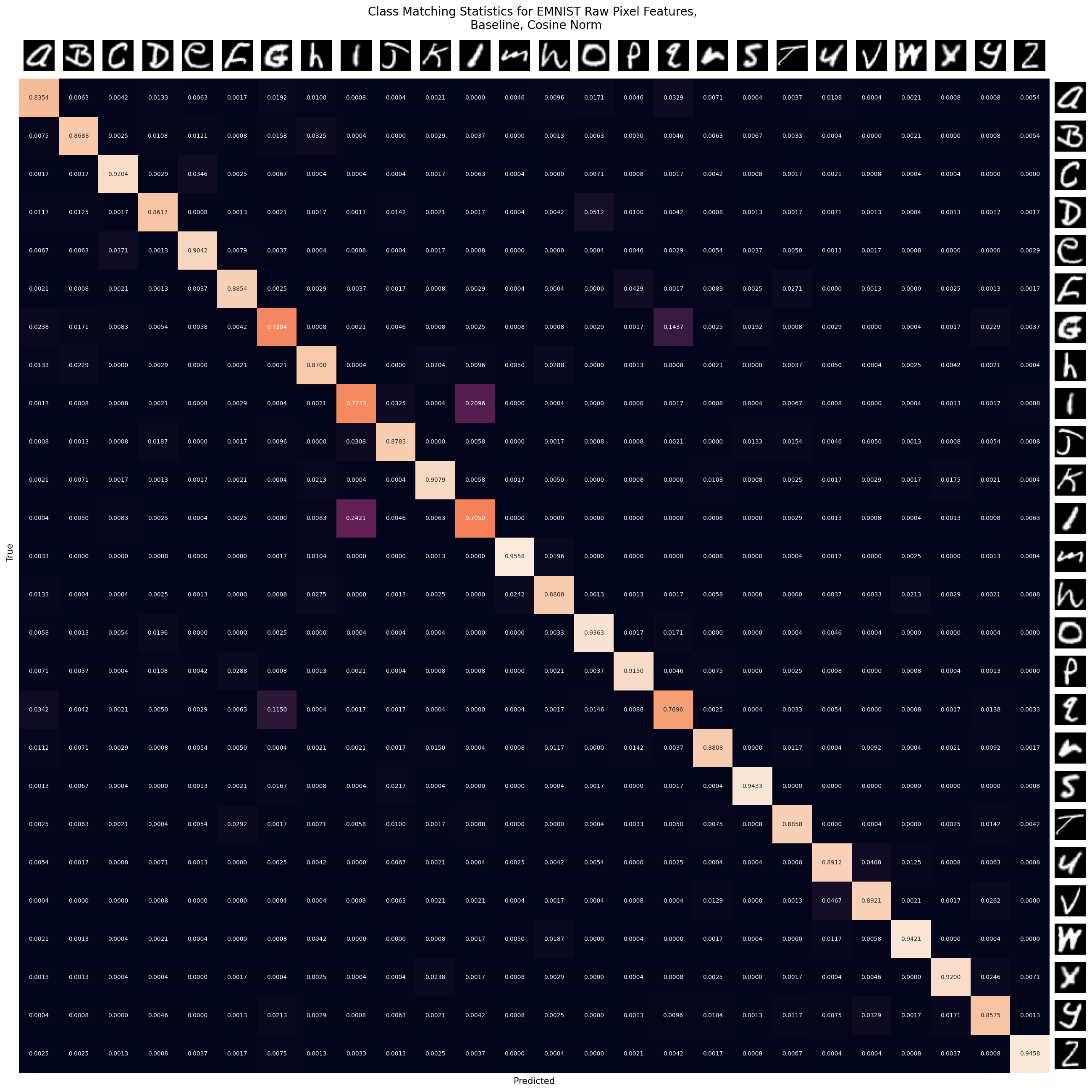}}
  }
\end{figure}

\end{document}